\documentclass[10pt,twocolumn,letterpaper]{article}

\usepackage[pagenumbers]{cvpr}
\usepackage{multirow}
\usepackage{algorithm}

\usepackage{listings}
\usepackage{makecell}

\usepackage{pifont}
\newcommand{\cmark}{\ding{51}}
\newcommand{\xmark}{\ding{55}}

\usepackage{titletoc}
\usepackage{fontawesome}
\usepackage{marvosym}

\usepackage{tikz}
\newcommand*\circled[1]{\tikz[baseline=(char.base)]{\node[shape=circle,fill,inner sep=0.2pt] (char) {\textcolor{white}{#1}};}}

\definecolor{moe_blue}{RGB}{99,113,250}
\definecolor{moe_red}{RGB}{239,99,75}
\definecolor{moe_green}{RGB}{0,180,139}
\definecolor{moe_gray}{RGB}{165,165,165}

\definecolor{nu_barrier}{RGB}{112,128,144}
\definecolor{nu_bicycle}{RGB}{220,20,60}
\definecolor{nu_bus}{RGB}{255,0,0}
\definecolor{nu_car}{RGB}{255,158,0}
\definecolor{nu_cons}{RGB}{233,150,70}
\definecolor{nu_motor}{RGB}{255,61,99}
\definecolor{nu_ped}{RGB}{0,0,230}
\definecolor{nu_cone}{RGB}{47,79,79}
\definecolor{nu_trailer}{RGB}{255,140,0}
\definecolor{nu_truck}{RGB}{255,99,71}
\definecolor{nu_driv}{RGB}{0,207,191}
\definecolor{nu_flat}{RGB}{175,0,75}
\definecolor{nu_sidewalk}{RGB}{75,0,75}
\definecolor{nu_terrain}{RGB}{112,180,60}
\definecolor{nu_manmade}{RGB}{222,184,135}
\definecolor{nu_veg}{RGB}{0,175,0}

\definecolor{cvprblue}{rgb}{0.21,0.49,0.74}
\usepackage[pagebackref=false,breaklinks,colorlinks,citecolor=cvprblue,linkcolor=cvprblue]{hyperref}

\newcommand\blfootnote[1]{%
\begingroup
\renewcommand\thefootnote{}{}\footnote{#1}%
\addtocounter{footnote}{-1}%
\endgroup
}

\usepackage[accsupp]{axessibility}

\title{LiMoE: Mixture of LiDAR Representation Learners from Automotive Scenes}

\author{
    Xiang Xu$^{*,1}$ \quad Lingdong Kong$^{*,2,3}$ \quad Hui Shuai$^4$ \quad Liang Pan$^3$ \quad Ziwei Liu$^5$ \quad Qingshan Liu$^{4,\textrm{\Letter}}$
    \\[0.8ex]
    {\small$^1$Nanjing University of Aeronautics and Astronautics \quad $^2$National University of Singapore \quad $^3$Shanghai AI Laboratory}
    \\
    {\small$^4$Nanjing University of Posts and Telecommunications \quad $^5$S-Lab, Nanyang Technological University}
    \\[0.8ex]
    \faGithub~\textbf{Code \& Checkpoints:} \url{https://github.com/Xiangxu-0103/LiMoE}
    \vspace{-0.7cm}
}

\begin{document}

\maketitle

\blfootnote{$(*)$ Xiang and Lingdong contributed equally to this work.}
\blfootnote{$(\textrm{\Letter})$ Corresponding author: \url{qsliu@njupt.edu.cn}.}

\begin{abstract}
    LiDAR data pretraining offers a promising approach to leveraging large-scale, readily available datasets for enhanced data utilization. However, existing methods predominantly focus on sparse voxel representation, overlooking the complementary attributes provided by other LiDAR representations. In this work, we propose \textsf{\textcolor{moe_red}{Li}\textcolor{moe_green}{MoE}}, a framework that integrates the Mixture of Experts (MoE) paradigm into LiDAR data representation learning to synergistically combine multiple representations, such as range images, sparse voxels, and raw points. Our approach consists of \textbf{three} stages: \textbf{i)} Image-to-LiDAR Pretraining, which transfers prior knowledge from images to point clouds across different representations; \textbf{ii)} Contrastive Mixture Learning (CML), which uses MoE to adaptively activate relevant attributes from each representation and distills these mixed features into a unified 3D network; \textbf{iii)} Semantic Mixture Supervision (SMS), which combines semantic logits from multiple representations to boost downstream segmentation performance. Extensive experiments across eleven large-scale LiDAR datasets demonstrate our effectiveness and superiority. The code has been made publicly accessible.
\end{abstract}

\vspace{-0.4cm}
\section{Introduction}
\label{sec:intro}

LiDAR perception is a cornerstone of modern autonomous driving systems, offering precise 3D spatial understanding crucial for navigation and safety \cite{pendleton2017perception,muhammad2020survey,kong2024robodrive,hao2024is}. However, developing accurate and scalable 3D perception models typically relies on large-scale, human-annotated datasets -- a process that is both costly and labor-intensive \cite{gao2021survey,li2024end,xiao2023survey}. This reliance presents a significant bottleneck in scaling autonomous systems, especially given the vast amount of sensor data generated in real-world driving environments \cite{ma2022vision}.

\begin{figure}[t]
    \centering
    \includegraphics[width=\linewidth]{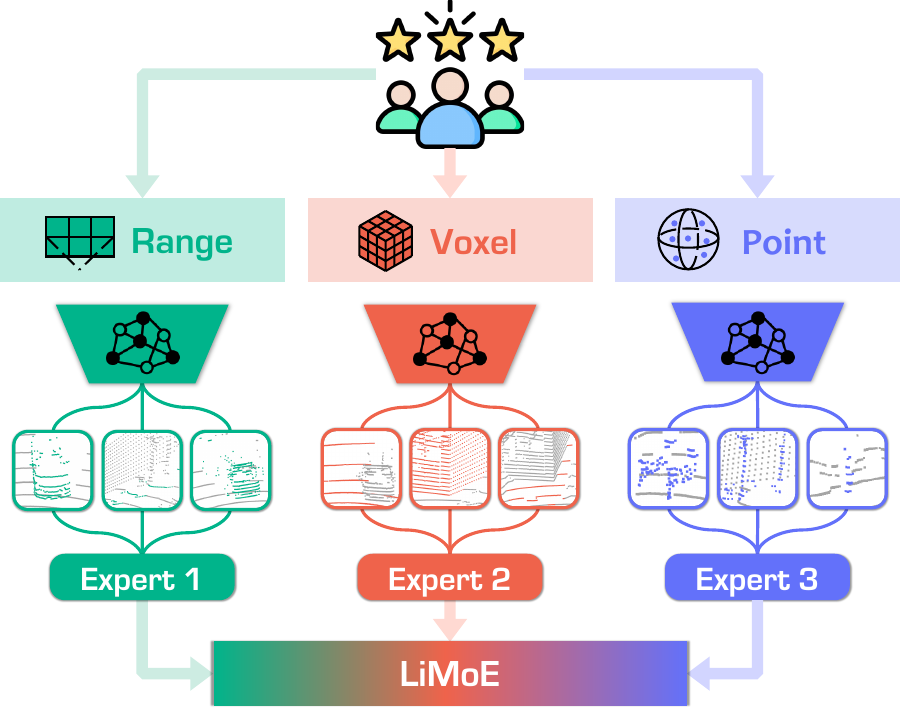}
    \vspace{-0.6cm}
    \caption{Illustration of the proposed mixture of LiDAR representation learning (\textsf{\textcolor{moe_red}{Li}\textcolor{moe_green}{MoE}}) design. We observe unique patterns of each LiDAR representation (\textcolor{moe_green}{range}, \textcolor{moe_red}{voxel}, and \textcolor{moe_blue}{point}) in image-to-LiDAR data pretraining. Our framework aims to integrate distinct attributes from different LiDAR representations into a unified feature space, enabling enhanced 3D scene understanding.}
    \label{fig:teaser}
    \vspace{-0.6cm}
\end{figure}

Recently, data representation learning has emerged as a promising solution to this challenge. By leveraging large-scale, easily accessible datasets, representation learning enables the extraction of meaningful data attributes without heavy reliance on manual annotations \cite{bengio2013survey,zeng2024self}. One line of this research \cite{xie2020pointcontrast,zhang2021depthcontrast} augments point clouds into different views and employs contrastive learning across these views for feature consistency. The other line \cite{liu2021ppkt,sautier2022slidr,liu2023seal,xu2024superflow} transfers knowledge from the image \cite{he2016resnet,dosovitskiy2021vit} to LiDAR space \cite{choy2019minkowski}, facilitating cross-modal learning for large-scale LiDAR scene understanding. However, these methods rely solely on sparse voxels \cite{choy2019minkowski}, failing to fully explore the complementary information existing in the LiDAR scene.

LiDAR data can be converted into various representations, \ie, range view, sparse voxels, and raw points, each offering unique advantages. Range images provide a compact 2D representation that excels at capturing dynamic elements and medium-range objects \cite{milioto2019rangenet++,xu2025frnet,kong2023rethinking}. Raw points preserve fine geometric details, enabling precise modeling of intricate structures \cite{hu2020randla,yan2020pointasnl}. Sparse voxels efficiently represent static objects and sparse regions, making them ideal for large-scale environments \cite{zhu2021cylinder3d,choy2019minkowski,hong2021dsnet}. Harnessing the complementary strengths of these representations holds the key to robust and comprehensive LiDAR-based perception.

In this work, we present \textsf{\textcolor{moe_red}{Li}\textcolor{moe_green}{MoE}}, a novel framework that synergistically integrates three LiDAR representations into a unified representation for feature learning. The framework operates in three stages: \textcolor{moe_green}{\circled{1}} Image-to-LiDAR pretraining, where knowledge from pretrained image backbone is transferred to LiDAR points for each representation \cite{sautier2022slidr,xu2024superflow}, initializing diverse representation-specific features; \textcolor{moe_red}{\circled{2}} Contrastive mixture learning (CML), which fuses pretrained features across representations into a unified representation, leveraging their complementary strengths; and \textcolor{moe_blue}{\circled{3}} Semantic mixture supervision (SMS), which enhances downstream performance by combining multiple semantic logits.

We observe that pretrained models with different LiDAR representations capture distinct data attributes (see~\cref{fig:teaser}). Range images primarily focus on middle laser beams and distances, sparse voxels emphasize upper beams and longer distances, while points capture lower beams and near distances. To effectively integrate these complementary attributes into a unified representation, CML employs a Mixture of Experts (MoE) layer that dynamically activates the relevant data attributes during pretraining. By distilling the combined features into a single representation feature, CML encourages the pretrained network to comprise comprehensive attributes derived from various representations.

In the downstream stage, we observe that different representation models capture distinct object attributes. Range images primarily focus on dynamic objects, sparse voxels emphasize static background structures, while points preserve fine-grained details. To further improve downstream performance, we extend the MoE layer into SMS, which dynamically activates relevant semantic features from various representations. With the supervision of semantic labels, SMS establishes a robust and scalable framework that enhances 3D scene understanding by leveraging complementary semantic features from each LiDAR representation.

To summarize, this work makes contributions as follows:
\begin{itemize}
    \item We propose \textsf{\textcolor{moe_red}{Li}\textcolor{moe_green}{MoE}}, a novel framework that integrates multiple LiDAR representations dynamically through the MoE paradigm. To our knowledge, this is the first work to explore MoE for LiDAR representation learning.
    \item We introduce a three-stage pipeline for enhanced LiDAR scene understanding, comprising knowledge transfer from images to LiDAR, the mixture of data attributes for single-representation pretraining, and integration of semantic attributes for downstream tasks, which offers a robust solution for LiDAR representation learning.
    \item Extensive experiments verify the effectiveness of our approach, achieving large gains over the state-of-the-art method on $11$ datasets, paving the way for developing scalable, robust, and generalizable automotive systems.
\end{itemize}

\section{Related Work}
\label{sec:related_work}

\noindent\textbf{LiDAR Data Representations.}
LiDAR point clouds, being irregular and unstructured, pose significant challenges for scene understanding \cite{gao2021survey,ma2022vision,hong20224dDSNet}. Recent approaches have sought to address this by transforming point clouds into structured representations. Point-based methods \cite{hu2020randla,shuai2021backward,yan2020pointasnl} process point clouds directly, preserving their full structure and enabling the capture of fine-grained details. Range view-based methods \cite{milioto2019rangenet++,kong2023rethinking,xu2025frnet,xu2020squeezesegv3,cortinhal2020salsanext} and bird's eye view-based methods \cite{zhou2020polarNet,zhou2021panoptic,chen2021polarStream} convert point clouds into 2D representations, leveraging image-processing techniques for efficient identification of dynamic or significant objects. Voxel-based methods \cite{choy2019minkowski,zhu2021cylinder3d,hong2021dsnet} discretize point clouds into regular voxel grids and use sparse convolutions \cite{tang2022torchsparse,tang2023torchsparse++,spconv2022} to efficiently process sparse regions within the data, making them suitable for large-scale environments.

\noindent\textbf{Multi-Representation LiDAR Segmentation.}
To capture comprehensive information from various LiDAR representations, recent works have combined multiple representations to explore their complementary strengths \cite{liu2024multi,puy23waffleiron}. AMVNet \cite{liong2020amvnet} and RPVNet \cite{xu2021rpvnet} employ late-fusion strategies, where features from each representation are processed independently and fused at the final stage. GFNet \cite{qiu2022gfNet} introduces a more advanced fusion method by combining features from multiple views at different stages. UniSeg \cite{liu2023uniseg} uses hierarchical fusion to integrate low-level features with high-level semantic information. In contrast, our approach leverages the Mixture of Experts (MoE) framework, which dynamically activates the most relevant features from multiple representations based on task context.

\noindent\textbf{Image-to-LiDAR Data Pretraining.}
Inspired by the success of self-supervised learning in the image domain \cite{chen2020simclr,he2020moco,he2022mae}, recent work has explored 3D data representation learning \cite{xie2020pointcontrast,zhang2021depthcontrast,boulch2023also,sautier2024bevcontrast}, which are limited to single-modality learning for small-scale scenes. The SLidR framework \cite{sautier2022slidr} introduced multi-modal self-supervised learning by transferring 2D image knowledge to 3D LiDAR models using contrastive learning. Subsequent works expanded this framework with class balancing \cite{mahmoud2023st}, hybrid-view distillation \cite{zhang2024hvdistill}, VFM-assisted superpixels \cite{liu2023seal}, spatiotemporal cues \cite{xu2024superflow}, \etc. Our work builds upon this by extending image-to-LiDAR pretraining to multiple LiDAR representations, capturing comprehensive LiDAR attributes.

\noindent\textbf{Mixture of Experts.}
The Mixture of Experts (MoE) framework consists of multiple sub-models that collectively enhance the model's capacity \cite{masoudnia2014mixture,cai2024survey,fedus2022review}. MoE dynamically selects a subset of experts to activate based on input, which increases scalability and flexibility for handling diverse tasks. This approach has been widely successful in Large Language Models (LLMs) \cite{du2022glam,fedus2022switch,lin2024moellava,zhong2024convlora,zhu2024llama}. Recently, MoE has been applied to vision tasks, including image classification \cite{chowdhury2023patch,shazeer2017outrageously,riquelme2021scaling}, object detection \cite{wu2022residual,chen2023adamv,hwang2023tutel}, and segmentation \cite{pavlitskaya2020moeseg,jiang2024m4oe}. However, MoE remains underexplored in 3D perception. In this work, we extend MoE to 3D data representation, enabling dynamic fusion of multiple LiDAR representations for improved scene understanding.

\section{Methodology}
\label{sec:approach}

\begin{figure*}[t]
    \centering
    \includegraphics[width=\linewidth]{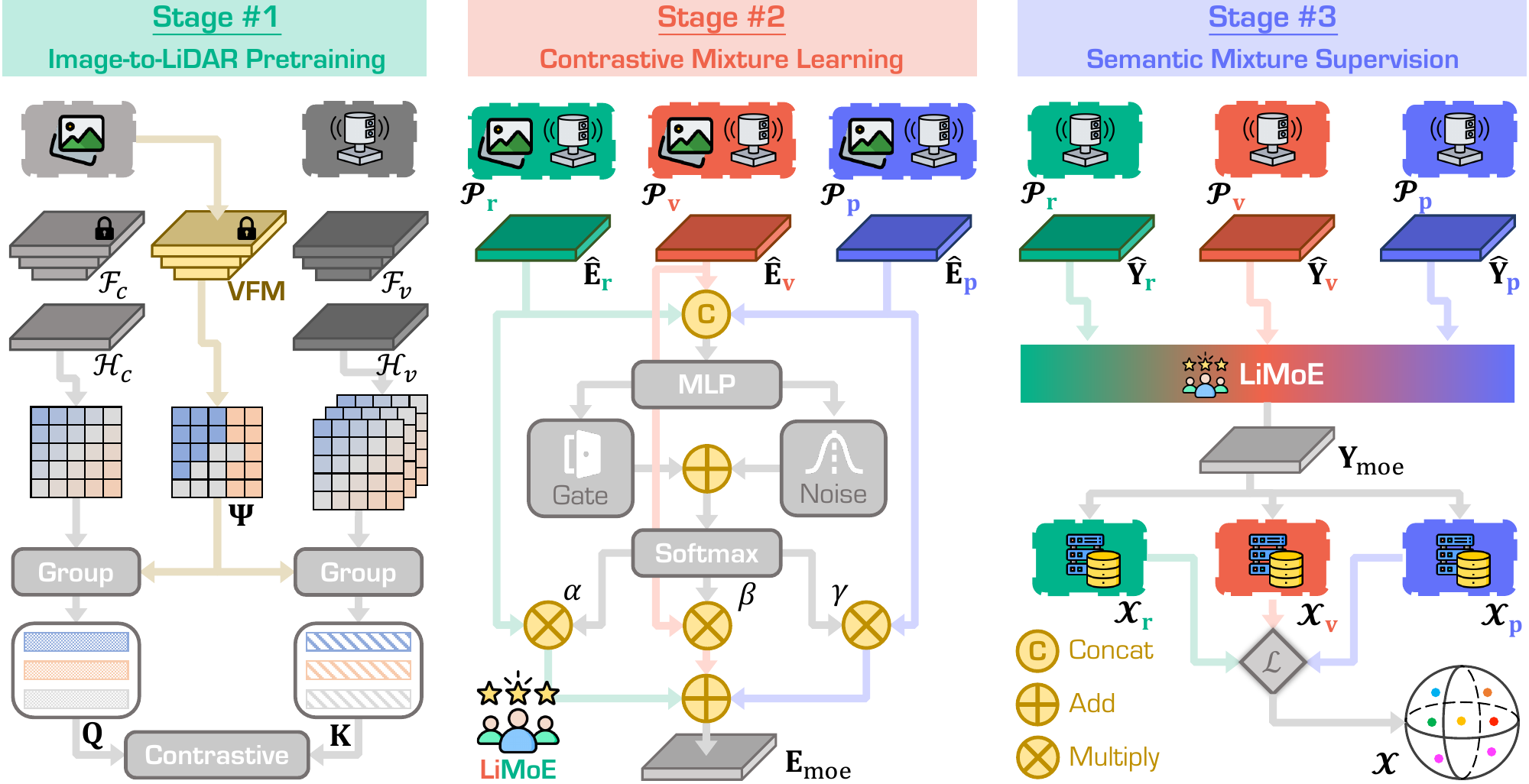}
    \vspace{-0.6cm}
    \caption{Overview of the \textsf{\textcolor{moe_red}{Li}\textcolor{moe_green}{MoE}} framework. Our design consists of \textbf{three} stages: \textbf{\textcolor{moe_green}{(\#1)}} The image-to-LiDAR pretraining transfers knowledge from images to various LiDAR representations (\cf~\cref{sec:stage1}); \textbf{\textcolor{moe_red}{(\#2)}} The contrastive mixture learning (CML) integrates the MoE framework to mix data attributes into a unified representation for pretraining (\cf~\cref{sec:stage2}); and \textbf{\textcolor{moe_blue}{(\#3)}} The semantic mixture supervision (SMS) fuses semantic logits from multiple representations to further enhance performance across different downstream tasks (\cf~\cref{sec:stage3}).}
    \label{fig:framework}
    \vspace{-0.4cm}
\end{figure*}

This work addresses LiDAR-based perception by leveraging multiple data representations to capture complementary information and enhance 3D scene understanding. We first describe the three common LiDAR representations and analyze their strengths (\cref{sec:representation}). Then, we detail the components of the \textsf{\textcolor{moe_red}{Li}\textcolor{moe_green}{MoE}} framework, as illustrated in \cref{fig:framework}. This comprises three stages: \textcolor{moe_green}{\circled{1}} cross-sensor knowledge transfer from images to LiDAR data (\cref{sec:stage1}); \textcolor{moe_red}{\circled{2}} mixture of LiDAR data attributes for pretraining (\cref{sec:stage2}); \textcolor{moe_blue}{\circled{3}} mixture of semantic attributes, supervised by human-annotated labels, for downstream segmentation tasks (\cref{sec:stage3}).

\subsection{LiDAR Representation}
\label{sec:representation}

Let $\mathcal{P} = \{\mathbf{p}_{i} \mid i=1,\ldots,N\}$ denote a LiDAR point cloud consisting of $N$ points, where each point $\mathbf{p}_{i} \in \mathbb{R}^{3+L}$ includes 3D coordinates $(x_{i}, y_{i}, z_{i})$ and $L$-dimensional features (\eg, intensity, elongation). To harness the unstructured and irregular points, various methods convert point clouds into intermediate representations, including range images $\mathcal{P}_{\textcolor{moe_green}{\mathbf{r}}}$, sparse voxels $\mathcal{P}_{\textcolor{moe_red}{\mathbf{v}}}$, and raw points $\mathcal{P}_{\textcolor{moe_blue}{\mathbf{p}}}$.

\noindent\textbf{Range View.}
The range image methods project the point cloud $\mathcal{P}$ onto 2D grids in spherical coordinates. Each point $\mathbf{p}_{i}$ is mapped to a 2D grid $(u_{i}^{r}, v_{i}^{r})$ in the range image as:
\begin{equation}
    \label{eq:p2r}
    \binom{u_{i}^{r}}{v_{i}^{r}} = \binom{\frac{1}{2}\left[1 - \arctan(y_{i}, x_{i}) \pi^{-1}\right]W_{r}}{\left[1 - (\arcsin(z_{i}d_{i}^{-1}) + \phi_{\text{down}})\phi^{-1}\right]H_{r}}~,
\end{equation}
where $d_{i}$ represents the depth of the point; $\phi$ denotes the vertical field-of-view of the sensor; $\phi_{\text{down}}$ is the inclination angles in the downward directions; $H_{r}$ and $W_{r}$ are the height and width of the range image. This projection results in a range image $\mathcal{P}_{\textcolor{moe_green}{\mathbf{r}}} \in \mathbb{R}^{H_{r} \times W_{r} \times (3+L)}$, allowing for efficient processing with image-based techniques. Range images offer a compact 2D representation of the 3D scene, capturing both geometric and intensity-based features \cite{milioto2019rangenet++,kong2023rethinking}. They are particularly effective in tackling dynamic or significant objects in the scene \cite{xu2025frnet,li2024rapid}.

\noindent\textbf{Sparse Voxels.}
The voxel-based methods discretize the point cloud $\mathcal{P}$ into regular voxel grids $\mathcal{P}_{\textcolor{moe_red}{\mathbf{v}}}$, where each voxel represents a small region of 3D space. For each point $\mathbf{p}_{i}$, its position is mapped to the corresponding voxel grid index as $[v_{i}^{x}, v_{i}^{y}, v_{i}^{z}] = [\lfloor x_{i} / s_{x} \rfloor, \lfloor y_{i} / s_{y} \rfloor, \lfloor z_{i} / s_{z} \rfloor]$, where $(s_{x}, s_{y}, s_{z})$ are the voxel sizes along the $x$, $y$ and $z$ dimensions. This discretization results in sparse voxel grids $\mathcal{P}_{\textcolor{moe_red}{\mathbf{v}}} \in \mathbb{R}^{M \times C}$, where $M < N$ is the number of non-empty voxels. To effectively process sparse voxels, sparse convolutions \cite{choy2019minkowski,tang2022torchsparse,tang2023torchsparse++,spconv2022} are employed, significantly reducing the computational complexity compared to regular voxel grids. Sparse voxels are particularly well-suited for representing large, sparsely populated areas, but can lose detail in dense regions due to quantization effects \cite{choy2019minkowski,zhu2021cylinder3d}.

\noindent\textbf{Raw Points.} The point-based methods process the point cloud (\ie, $\mathcal{P}_{\textcolor{moe_blue}{\mathbf{p}}} = \mathcal{P}$) without converting to other representations. These methods typically involve three key steps: 1) Sampling a set of central points (centroids) from the point cloud; 2) Neighbor search, where points in the vicinity of each centroid are identified based on spatial proximity; 3) Feature aggregation, where MLPs are used to combine the features of neighboring points and propagate them to the centroids. While raw points preserve the fine-grained structure of the scene, they are computationally expensive due to the need for point-wise operations \cite{hu2020randla,yan2020pointasnl}.

\subsection{Image-to-LiDAR Pretraining}
\label{sec:stage1}

Image-to-LiDAR pretraining aims to transfer knowledge from images to LiDAR data, facilitating the learning of effective 3D representations even in the absence of extensive LiDAR labels. Previous works \cite{sautier2022slidr,mahmoud2023st,liu2023seal,xu2024superflow} have largely focused on sparse voxel representations for this task, as they offer efficient processing for large-scale LiDAR data. 

Given a set of $V$ synchronized images $\mathcal{I} = \{\mathbf{I}_{i} \mid i=1,\ldots,V\}$ and their corresponding LiDAR point clouds $\mathcal{P}$, where each image $\mathbf{I} \in \mathbb{R}^{H \times W \times 3}$ has a spatial resolution with height $H$ and width $W$. Each LiDAR point $\mathbf{p}_{i}$ can be projected onto the corresponding image plane $(u_{i}, v_{i})$ as:
\begin{equation}
    \label{eq:calibration}
    [u_{i}, v_{i}]^{\text{T}} = (1/{z_{i}}) \times \Gamma_{K} \times \Gamma_{l \to c} \times [x_{i}, y_{i}, z_{i}]^{\text{T}}~,
\end{equation}
where $\Gamma_{l \to c}$ is the transformation matrix from the LiDAR to the camera coordinate system, and $\Gamma_{K}$ is the camera intrinsic matrix. This pretraining process involves two key steps, as shown on the left side (Phase \#1) of \cref{fig:framework}.

\noindent\textbf{Superpixel \& Superpoint Generation.} To establish correlations between images and the point cloud, prior works use the unsupervised SLIC algorithm \cite{achanta2012slic} or vision foundation models (VFMs) \cite{kirillov2023sam,zou2023seem,zou2023xcoder,zhang2023openSeeD} to generate a set of $S$ superpixels for each image, denoted as $\mathbf{\Psi} = \{\mathbf{\psi}_{i} \mid i=1,\ldots,S\}$. The corresponding superpoint set $\mathbf{\Omega} = \{\mathbf{\omega}_{i} \mid i=1,\ldots,S\}$ is then derived by projecting these superpixels onto the point cloud using the transformation in \cref{eq:calibration}.

\noindent\textbf{Contrastive Objective.} To transfer the knowledge from images to LiDAR, both image and sparse voxel data are passed through their respective backbones: the 2D pretrained backbone $\mathcal{F}_{c}$ for images and the 3D voxel backbone $\mathcal{F}_{\textcolor{moe_red}{\mathbf{v}}}$ for sparse voxels. This generates the respective image and voxel features. These features are then processed by linear projection heads $\mathcal{H}_{c}$ and $\mathcal{H}_{\textcolor{moe_red}{\mathbf{v}}}$, which align the feature spaces and produce $D$-dimensional feature embeddings. The image and voxel features are then grouped based on superpixels $\mathbf{\Psi}$ and superpoints $\mathbf{\Omega}$, resulting in superpixel embeddings $\mathbf{Q} \in \mathbb{R}^{S \times D}$ and superpoint embeddings $\mathbf{K} \in \mathbb{R}^{S \times D}$. Finally, a contrastive loss is applied to ensure that each superpoint embedding is closely correlated with its corresponding superpixel embedding:
\begin{equation}
    \label{eq:contrastive}
    \mathcal{L}_{\text{con}} = \frac{1}{S} \sum_{i=1}^{S} \log \frac{e^{\langle \mathbf{k}_{i}, \mathbf{q}_{i} \rangle / \tau}}{\sum_{j \neq i} e^{\langle \mathbf{k}_{i}, \mathbf{q}_{j} \rangle / \tau}}~,
\end{equation}
where $\langle \cdot, \cdot \rangle$ denotes the dot product between the superpixel and superpoint embeddings, and $\tau > 0$ is a temperature.

However, sparse voxel-based pretraining is limited in its ability to fully exploit the detailed geometry and appearance characteristics of 3D scenes. In fact, LiDAR point clouds can be represented in various forms, each emphasizing different attributes, such as laser beams, distance, and static/dynamic objects, within the scene \cite{xu2021rpvnet}. To this end, we propose a novel pretraining paradigm that integrates multiple representations of LiDAR point clouds. This approach enables a more comprehensive understanding of the scenes by capturing both geometric and detailed information across diverse representations, ultimately constructing a richer, more detailed representation of 3D environments.

\subsection{CML: Contrastive Mixture Learning}
\label{sec:stage2}
We extend the image-to-LiDAR pretraining method by independently training three separate 3D networks for each representation: range images, sparse voxels, and raw points. These networks serve as the foundation for our CML approach. As shown in the middle of \cref{fig:framework}, each network processes the point cloud in its respective representation form:
\begin{align}
    \label{eq:pretrain_forward}
    \mathbf{E}_{\textcolor{moe_green}{\mathbf{r}}} &= \mathcal{H}_{\textcolor{moe_green}{\mathbf{r}}}(\mathcal{F}_{\textcolor{moe_green}{\mathbf{r}}}(\mathcal{P}_{\textcolor{moe_green}{\mathbf{r}}})) \in \mathbb{R}^{H_{r} \times W_{r} \times D}~, \nonumber \\
    \mathbf{E}_{\textcolor{moe_red}{\mathbf{v}}} &= \mathcal{H}_{\textcolor{moe_red}{\mathbf{v}}}(\mathcal{F}_{\textcolor{moe_red}{\mathbf{v}}}(\mathcal{P}_{\textcolor{moe_red}{\mathbf{v}}})) \in \mathbb{R}^{M \times D}~, \\
    \mathbf{E}_{\textcolor{moe_blue}{\mathbf{p}}} &= \mathcal{H}_{\textcolor{moe_blue}{\mathbf{p}}}(\mathcal{F}_{\textcolor{moe_blue}{\mathbf{p}}}(\mathcal{P}_{\textcolor{moe_blue}{\mathbf{p}}})) \in \mathbb{R}^{N \times D}~. \nonumber
\end{align}
Following the success of the Mixture of Experts (MoE) approach \cite{zhong2024convlora,lin2024moellava}, we introduce an MoE layer that dynamically selects and combines relevant features from different representations for each point.

\begin{figure*}[t]
    \centering
    \begin{subfigure}[h]{0.48\textwidth}
        \centering
        \includegraphics[width=\linewidth]{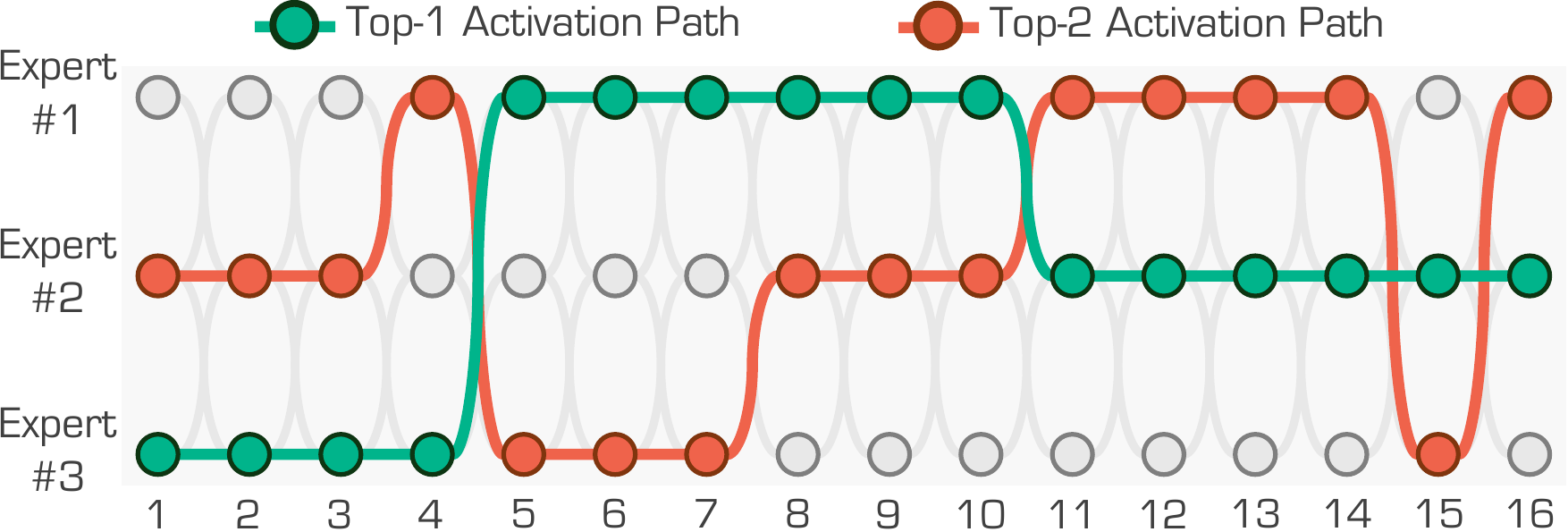}
        \caption{Expert Activation Paths (Beam Number)}
        \label{fig:path_beam}
    \end{subfigure}~~~~~
    \begin{subfigure}[h]{0.48\textwidth}
        \centering
        \includegraphics[width=\linewidth]{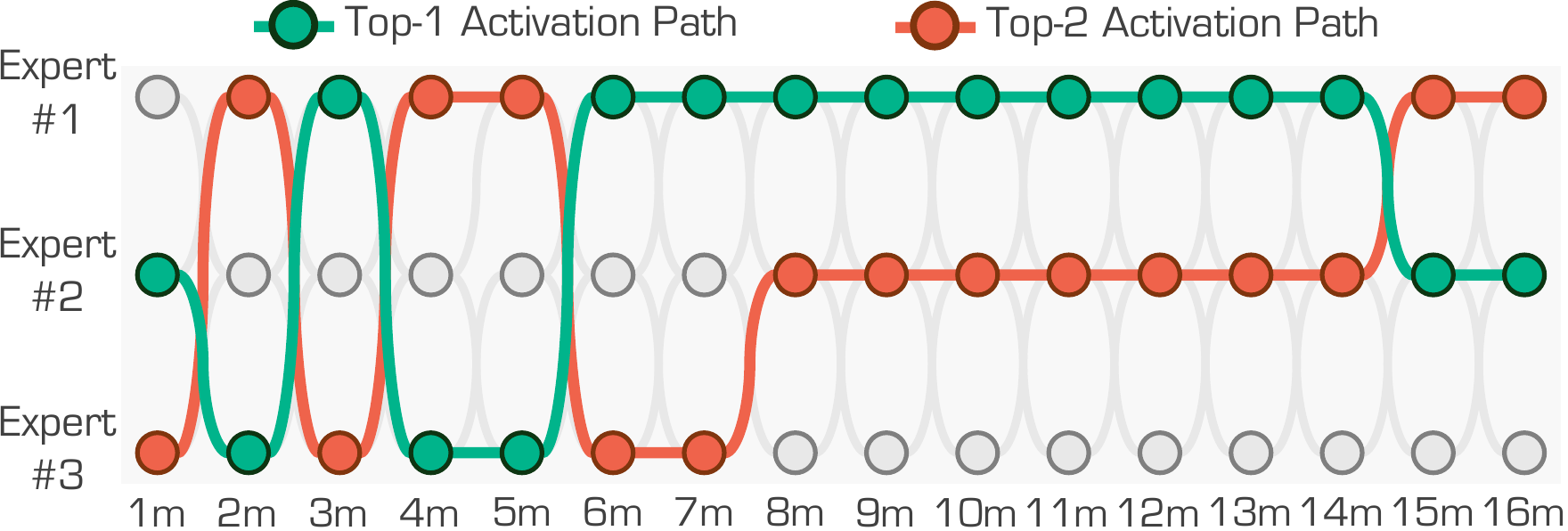}
        \caption{Expert Activation Paths (Distances to Ego)}
        \label{fig:path_distance}
    \end{subfigure}
    \vspace{-0.2cm}
    \caption{Visual interpretations of the expert activation paths in CML. The experts are \textcolor{moe_green}{\texttt{\#1}} range view, \textcolor{moe_red}{\texttt{\#2}} voxel, and \textcolor{moe_blue}{\texttt{\#3}} point, respectively.}
    \label{fig:ablation_activation}
    \vspace{-0.4cm}
\end{figure*}

\noindent\textbf{Feature Alignment.} The features generated from each representation, as in \cref{eq:pretrain_forward}, vary significantly due to the different representation attributes. To align these features, we project the range image and sparse voxel features into the point cloud space. This results in aligned feature sets $\widehat{\mathbf{E}}_{\textcolor{moe_green}{\mathbf{r}}}$, $\widehat{\mathbf{E}}_{\textcolor{moe_red}{\mathbf{v}}}$, and $\widehat{\mathbf{E}}_{\textcolor{moe_blue}{\mathbf{p}}}$, all with the form $\mathbb{R}^{N \times D}$, ensuring consistent feature space across these three representations.

\noindent\textbf{MoE Layer.} This layer takes $\widehat{\mathbf{E}}_{\textcolor{moe_green}{\mathbf{r}}}$, $\widehat{\mathbf{E}}_{\textcolor{moe_red}{\mathbf{v}}}$, and $\widehat{\mathbf{E}}_{\textcolor{moe_blue}{\mathbf{p}}}$ as the input. To combine the features from these three representations, we first concatenate them and then apply an MLP layer to reduce the channel dimension. The MoE layer consists of two key components: a gate module, which dynamically selects the activated representation for each point, and a noise module, which introduces perturbations to prevent overfitting. This process can be formulated as follows:
\begin{align}
    \label{eq:moe}
    \mathbf{G} &= \mathbf{E} \cdot \mathbf{Z}_{g} + \chi \cdot \sigma(\mathbf{E} \cdot \mathbf{Z}_{n})~,\\ \mathbf{E} &= \text{MLP}([\widehat{\mathbf{E}}_{\textcolor{moe_green}{\mathbf{r}}}, \widehat{\mathbf{E}}_{\textcolor{moe_red}{\mathbf{v}}}, \widehat{\mathbf{E}}_{\textcolor{moe_blue}{\mathbf{p}}}])~.
\end{align}
Here, $\mathbf{Z}_{g} \in \mathbb{R}^{D \times 3}$ and $\mathbf{Z}_{n} \in \mathbb{R}^{D \times 3}$ are trainable weights for the gate and noise modules, respectively. The variable $\chi$ represents a random noise distribution, which is applied in the noise module to introduce variability into the feature selection process. The function $\sigma$ denotes the Softplus activation \cite{nair2010softplus}, ensuring that the perturbations remain positive and smooth. The operation $[\cdot,\cdot]$ denotes feature concatenation. A softmax function is then applied to the gate values $\mathbf{G}$, yielding selection scores $\widehat{\mathbf{G}} \in \mathbb{R}^{N \times 3}$. These scores are then split into $\alpha$, $\beta$, and $\gamma$ for each representation, respectively, indicating the importance of each representation for each point. The final output features are obtained by weighting and summing the features from each representation: $\mathbf{E}_{\text{moe}} = \alpha \widehat{\mathbf{E}}_{\textcolor{moe_green}{\mathbf{r}}} + \beta \widehat{\mathbf{E}}_{\textcolor{moe_red}{\mathbf{v}}} + \gamma \widehat{\mathbf{E}}_{\textcolor{moe_blue}{\mathbf{p}}}$. These features capture the dynamic contributions of each representation, enabling the model to adaptively prioritize the most relevant features for each point in the LiDAR point cloud. This selective feature fusion enables the network to leverage the strengths of each representation dependent on the context, thereby improving its ability to capture detailed scene attributes.

\noindent\textbf{Training Objective.} In this stage, the goal is to distill the MoE features from multiple representations into a single LiDAR representation (range, voxel, or point). To accomplish this, we use a 3D student network to extract corresponding student features $\mathbf{E}_{l}^{s} = \mathcal{H}_{l}^{s}(\mathcal{F}_{l}^{s}(\mathcal{P}_{l}))$ ($l \in \{\textcolor{moe_green}{\mathbf{r}},\textcolor{moe_red}{\mathbf{v}},\textcolor{moe_blue}{\mathbf{p}}\}$). Then, the MoE features and student features are grouped based on superpoints to generate the superpoint embeddings $\mathbf{K}^{\text{moe}}$ and $\mathbf{K}^{l}$. A contrastive loss is then applied between them as:
\begin{equation}
    \label{eq:moe_loss}
    \mathcal{L}_{\text{CML}} = \frac{1}{S} \sum_{i=1}^{S} \log \frac{e^{\langle \mathbf{k}_{i}^{l}, \mathbf{k}_{i}^{\text{moe}} \rangle / \tau}}{\sum_{j \neq i} e^{\langle \mathbf{k}_{i}^{l}, \mathbf{k}_{j}^{\text{moe}} \rangle / \tau}}~.
\end{equation}
The contrastive loss encourages the student features to align with the MoE features, allowing the single 3D student network to effectively capture informative structures from all the different representations. As depicted in \cref{fig:ablation_activation}, CML encourages the MoE framework to focus on data attributes from laser beams and distances.

\subsection{SMS: Semantic Mixture Supervision}
\label{sec:stage3}

To further improve downstream semantic segmentation performance, we extend the MoE layer into the downstream tasks, as shown in the right side of \cref{fig:framework}. This integration enables the model to dynamically select and prioritize the most relevant object attributes from each representation, tailoring feature fusion to the specific context of the task. Building upon the pretrained representations in \cref{sec:stage2}, each backbone independently processes the point cloud to generate semantic logits for each representation:
\begin{align}
    \label{eq:downstream_forward}
    \mathbf{Y}_{\textcolor{moe_green}{\mathbf{r}}} &= \widetilde{\mathcal{H}}_{\textcolor{moe_green}{\mathbf{r}}}^{s}(\mathcal{F}_{\textcolor{moe_green}{\mathbf{r}}}^{s}(\mathcal{P}_{\textcolor{moe_green}{\mathbf{r}}})) \in \mathbb{R}^{H_{r} \times W_{r} \times C}~, \nonumber \\
    \mathbf{Y}_{\textcolor{moe_red}{\mathbf{v}}} &= \widetilde{\mathcal{H}}_{\textcolor{moe_red}{\mathbf{v}}}^{s}(\mathcal{F}_{\textcolor{moe_red}{\mathbf{v}}}^{s}(\mathcal{P}_{\textcolor{moe_red}{\mathbf{v}}})) \in \mathbb{R}^{M \times C}~, \\
    \mathbf{Y}_{\textcolor{moe_blue}{\mathbf{p}}} &= \widetilde{\mathcal{H}}_{\textcolor{moe_blue}{\mathbf{p}}}^{s}(\mathcal{F}_{\textcolor{moe_blue}{\mathbf{p}}}^{s}(\mathcal{P}_{\textcolor{moe_blue}{\mathbf{p}}})) \in \mathbb{R}^{N \times C}~, \nonumber
\end{align}
where $C$ is the number of semantic classes. $\widetilde{\mathcal{H}}_{\textcolor{moe_green}{\mathbf{r}}}^{s}$, $\widetilde{\mathcal{H}}_{\textcolor{moe_red}{\mathbf{v}}}^{s}$, and $\widetilde{\mathcal{H}}_{\textcolor{moe_blue}{\mathbf{p}}}^{s}$ are linear heads that project backbone features into semantic logits for each representation. To align logits across representations, we project them into the point cloud space, yielding $\widehat{\mathbf{Y}}_{\textcolor{moe_green}{\mathbf{r}}}$,  $\widehat{\mathbf{Y}}_{\textcolor{moe_red}{\mathbf{v}}}$, and $\widehat{\mathbf{Y}}_{\textcolor{moe_blue}{\mathbf{p}}}$, each with a form of $\mathbb{R}^{N \times C}$.

The MoE layer consists of a gating module that dynamically selects the activated representations for each point, and a noise module that introduces perturbations to the features during training to mitigate overfitting. Different from \cref{eq:moe}, the noise gate is only active during training. The whole process can be formulated as:
\begin{equation}
    \label{eq:moe_downstream}
    \mathbf{G}_{d} = \mathbf{Y} \cdot \mathbf{Z}_{g} + \zeta \cdot \chi \cdot \sigma(\mathbf{Y} \cdot \mathbf{Z}_{n})~,
\end{equation}
where $\zeta = 1$ for training and $\zeta = 0$ for inference, and $\mathbf{Y} = \text{MLP}([\widehat{\mathbf{Y}}_{\textcolor{moe_green}{\mathbf{r}}}, \widehat{\mathbf{Y}}_{\textcolor{moe_red}{\mathbf{v}}}, \widehat{\mathbf{Y}}_{\textcolor{moe_blue}{\mathbf{p}}}])$. After applying the softmax function to the gate value $\mathbf{G}_{d}$, we obtain the coefficients $\alpha_{d}$, $\beta_{d}$, and $\gamma_{d}$, which indicate the relative importance of each representation for every point. The final semantic logits are obtained by a weighted sum of the logits from each representation: $\mathbf{Y}_{\text{moe}} = \alpha_{d} \widehat{\mathbf{Y}}_{\textcolor{moe_green}{\mathbf{r}}} + \beta_{d} \widehat{\mathbf{Y}}_{\textcolor{moe_red}{\mathbf{v}}} + \gamma_{d} \widehat{\mathbf{Y}}_{\textcolor{moe_blue}{\mathbf{p}}}$.

\begin{table*}[t]
    \centering
    \caption{\textbf{Comparisons of state-of-the-art LiDAR pretraining methods} pretrained on \textit{nuScenes} \cite{caesar2020nuScenes} and fine-tuned on \textit{nuScenes} \cite{fong2022panoptic-nuScenes}, \textit{SemanticKITTI} \cite{behley2019semanticKITTI}, and \textit{Waymo Open} \cite{sun2020waymoOpen} datasets, respectively, with specified data portions. \textbf{LP} denotes linear probing with frozen backbones. All scores are given in percentage (\%). The \textbf{best} and \underline{2nd best} scores under each group are highlighted in \textbf{bold} and \underline{underline}.}
    \vspace{-0.25cm}
    \label{tab:benchmark}
    \resizebox{\linewidth}{!}{
    \begin{tabular}{r|r|c|c|r|p{30pt}<{\centering}p{30pt}<{\centering}p{30pt}<{\centering}p{30pt}<{\centering}p{30pt}<{\centering}p{30pt}<{\centering}|p{30pt}<{\centering}|p{30pt}<{\centering}}
    \toprule
    \multirow{2}{*}{\textbf{Method}} & \multirow{2}{*}{\textbf{Venue}} & ~\textbf{Backbone}~ & ~\textbf{Backbone}~ & \multirow{2}{*}{\textbf{Expert}} & \multicolumn{6}{c|}{\textbf{nuScenes}} & \textbf{KITTI} & \textbf{Waymo}
    \\
    & & \textbf{(2D)} & \textbf{(3D)} & & \textbf{LP} & \textbf{1\%} & \textbf{5\%} & \textbf{10\%} & \textbf{25\%} & \textbf{Full} & \textbf{1\%} & \textbf{1\%}
    \\\midrule\midrule
    \cellcolor{moe_gray!18}\textcolor{gray}{Random} & \cellcolor{moe_gray!18}\textcolor{gray}{-} & \cellcolor{moe_gray!18}\textcolor{gray}{-} & \cellcolor{moe_gray!18}\textcolor{gray}{-} & \cellcolor{moe_gray!18}\textcolor{gray}{-} & \cellcolor{moe_gray!18}\textcolor{gray}{$8.10$} & \cellcolor{moe_gray!18}\textcolor{gray}{$30.30$} & \cellcolor{moe_gray!18}\textcolor{gray}{$47.84$} & \cellcolor{moe_gray!18}\textcolor{gray}{$56.15$} & \cellcolor{moe_gray!18}\textcolor{gray}{$65.48$} & \cellcolor{moe_gray!18}\textcolor{gray}{$74.66$} & \cellcolor{moe_gray!18}\textcolor{gray}{$39.50$} & \cellcolor{moe_gray!18}\textcolor{gray}{$39.41$}
    \\\midrule
    SLidR \cite{sautier2022slidr} & CVPR'22 & \multirow{5}{*}{\makecell{ResNet-50 \\ \cite{he2016resnet}}} & \multirow{5}{*}{\makecell{MinkUNet-34 \\ \cite{choy2019minkowski}}} & Single~\textcolor{moe_red}{$\circ$} & $38.80$ & $38.30$ & $52.49$ & $59.84$ & $66.91$ & $74.79$ & $44.60$ & $47.12$
    \\
    TriCC \cite{pang2023tricc} & CVPR'23 & & & Single~\textcolor{moe_red}{$\circ$} & $38.00$ & $41.20$ & $54.10$ & $60.40$ & $67.60$ & $75.60$ & $45.90$ & $-$
    \\
    Seal \cite{liu2023seal} & NeurIPS'23 & & & Single~\textcolor{moe_red}{$\circ$} & \underline{$44.95$} & \underline{$45.84$} & $55.64$ & \underline{$62.97$} & $68.41$ & $75.60$ & $46.63$ & $\mathbf{49.34}$
    \\
    CSC \cite{chen2024csc} & CVPR'24 & & & Single~\textcolor{moe_red}{$\circ$} & $\mathbf{46.00}$ & $\mathbf{47.00}$ & $\mathbf{57.00}$ & $\mathbf{63.30}$ & \underline{$68.60$} & \underline{$75.70$} & \underline{$47.20$} & $-$
    \\
    HVDistill \cite{zhang2024hvdistill} & IJCV'24 & & & Single~\textcolor{moe_red}{$\circ$} & $39.50$ & $42.70$ & \underline{$56.60$} & $62.90$ & $\mathbf{69.30}$ & $\mathbf{76.60}$ & $\mathbf{49.70}$ & $-$
    \\\midrule
    SLidR \cite{sautier2022slidr} & CVPR'22 & \multirow{5}{*}{\makecell{ViT-S \\ \cite{oquab2023dinov2}}} & \multirow{5}{*}{\makecell{MinkUNet-34 \\ \cite{choy2019minkowski}}} & Single~\textcolor{moe_red}{$\circ$} & $44.70$ & $41.16$ & $53.65$ & $61.47$ & $66.71$ & $74.20$ & $44.67$ & $47.57$
    \\
    + \textsf{\textcolor{moe_red}{Li}\textcolor{moe_green}{MoE}} & \textbf{Ours} & & & Multi~\textcolor{moe_green}{$\bullet$} & \cellcolor{moe_green!12}$45.80$ & \cellcolor{moe_green!12}$46.82$ & \cellcolor{moe_green!12}$57.54$ & \cellcolor{moe_green!12}$63.85$ & \cellcolor{moe_green!12}$68.61$ & \cellcolor{moe_green!12}$75.64$ & \cellcolor{moe_green!12}$46.81$ & \cellcolor{moe_green!12}$48.81$
    \\
    Seal \cite{liu2023seal} & NeurIPS'23 & & & Single~\textcolor{moe_red}{$\circ$} & $45.16$ & $44.27$ & $55.13$ & $62.46$ & $67.64$ & $75.58$ & $46.51$ & $48.67$
    \\
    SuperFlow \cite{xu2024superflow} & ECCV'24 & & & Single~\textcolor{moe_red}{$\circ$} & \underline{$46.44$} & \underline{$47.81$} & \underline{$59.44$} & \underline{$64.47$} & \underline{$69.20$} & \underline{$76.54$} & \underline{$47.97$} & \underline{$49.94$}
    \\
    + \textsf{\textcolor{moe_red}{Li}\textcolor{moe_green}{MoE}} & \textbf{Ours} & & & Multi~\textcolor{moe_green}{$\bullet$} & \cellcolor{moe_green!12}$\mathbf{48.20}$ & \cellcolor{moe_green!12}$\mathbf{49.60}$ & \cellcolor{moe_green!12}$\mathbf{60.54}$ & \cellcolor{moe_green!12}$\mathbf{65.65}$ & \cellcolor{moe_green!12}$\mathbf{71.39}$ & \cellcolor{moe_green!12}$\mathbf{77.27}$ & \cellcolor{moe_green!12}$\mathbf{49.53}$ & \cellcolor{moe_green!12}$\mathbf{51.42}$
    \\\midrule
    SLidR \cite{sautier2022slidr} & CVPR'22 & \multirow{5}{*}{\makecell{ViT-B \\ \cite{oquab2023dinov2}}} & \multirow{5}{*}{\makecell{MinkUNet-34 \\ \cite{choy2019minkowski}}} & Single~\textcolor{moe_red}{$\circ$} & $45.35$ & $41.64$ & $55.83$ & $62.68$ & $67.61$ & $74.98$ & $45.50$ & $48.32$
    \\
    + \textsf{\textcolor{moe_red}{Li}\textcolor{moe_green}{MoE}} & \textbf{Ours} & & & Multi~\textcolor{moe_green}{$\bullet$} & \cellcolor{moe_green!12}$46.56$ & \cellcolor{moe_green!12}$46.89$ & \cellcolor{moe_green!12}$58.09$ & \cellcolor{moe_green!12}$63.87$ & \cellcolor{moe_green!12}$69.02$ & \cellcolor{moe_green!12}$75.87$ & \cellcolor{moe_green!12}$47.96$ & \cellcolor{moe_green!12}$49.50$
    \\
    Seal \cite{liu2023seal} & NeurIPS'23 & & & Single~\textcolor{moe_red}{$\circ$} & $46.59$ & $45.98$ & $57.15$ & $62.79$ & $68.18$ & $75.41$ & $47.24$ & $48.91$
    \\
    SuperFlow \cite{xu2024superflow} & ECCV'24 & & & Single~\textcolor{moe_red}{$\circ$} & \underline{$47.66$} & \underline{$48.09$} & \underline{$59.66$} & \underline{$64.52$} & \underline{$69.79$} & \underline{$76.57$} & \underline{$48.40$} & \underline{$50.20$}
    \\
    + \textsf{\textcolor{moe_red}{Li}\textcolor{moe_green}{MoE}} & \textbf{Ours} & & & Multi~\textcolor{moe_green}{$\bullet$} & \cellcolor{moe_green!12}$\mathbf{49.07}$ & \cellcolor{moe_green!12}$\mathbf{50.23}$ & \cellcolor{moe_green!12}$\mathbf{61.51}$ & \cellcolor{moe_green!12}$\mathbf{66.17}$ & \cellcolor{moe_green!12}$\mathbf{71.56}$ & \cellcolor{moe_green!12}$\mathbf{77.81}$ & \cellcolor{moe_green!12}$\mathbf{50.30}$ & \cellcolor{moe_green!12}$\mathbf{51.77}$
    \\\midrule
    SLidR \cite{sautier2022slidr} & CVPR'22 & \multirow{5}{*}{\makecell{ViT-L \\ \cite{oquab2023dinov2}}} & \multirow{5}{*}{\makecell{MinkUNet-34 \\ \cite{choy2019minkowski}}} & Single~\textcolor{moe_red}{$\circ$} & $45.70$ & $42.77$ & $57.45$ & $63.20$ & $68.13$ & $75.51$ & $47.01$ & $48.60$
    \\
    + \textsf{\textcolor{moe_red}{Li}\textcolor{moe_green}{MoE}} & \textbf{Ours} & & & Multi~\textcolor{moe_green}{$\bullet$} & \cellcolor{moe_green!12}$47.43$ & \cellcolor{moe_green!12}$46.92$ & \cellcolor{moe_green!12}$58.41$ & \cellcolor{moe_green!12}$64.54$ & \cellcolor{moe_green!12}$69.69$ & \cellcolor{moe_green!12}$76.32$ & \cellcolor{moe_green!12}$48.25$ & \cellcolor{moe_green!12}$50.23$
    \\
    Seal \cite{liu2023seal} & NeurIPS'23 & & & Single~\textcolor{moe_red}{$\circ$} & $46.81$ & $46.27$ & $58.14$ & $63.27$ & $68.67$ & $75.66$ & $47.55$ & $50.02$
    \\
    SuperFlow \cite{xu2024superflow} & ECCV'24 & & & Single~\textcolor{moe_red}{$\circ$} & \underline{$48.01$} & \underline{$49.95$} & \underline{$60.72$} & \underline{$65.09$} & \underline{$70.01$} & \underline{$77.19$} & \underline{$49.07$} & \underline{$50.67$}
    \\
    + \textsf{\textcolor{moe_red}{Li}\textcolor{moe_green}{MoE}} & \textbf{Ours} & & & Multi~\textcolor{moe_green}{$\bullet$} & \cellcolor{moe_green!12}$\mathbf{49.35}$ & \cellcolor{moe_green!12}$\mathbf{51.41}$ & \cellcolor{moe_green!12}$\mathbf{62.07}$ & \cellcolor{moe_green!12}$\mathbf{66.64}$ & \cellcolor{moe_green!12}$\mathbf{71.59}$ & \cellcolor{moe_green!12}$\mathbf{77.85}$ & \cellcolor{moe_green!12}$\mathbf{50.69}$ & \cellcolor{moe_green!12}$\mathbf{51.93}$
    \\\bottomrule
\end{tabular}}
\vspace{-0.2cm}
\end{table*}

\begin{table*}[t]
    \centering
    \caption{\textbf{Domain generalization study of different LiDAR pretraining methods} pretrained on the \textit{nuScenes} \cite{caesar2020nuScenes} dataset and fine-tuned on a collection of seven different LiDAR semantic segmentation datasets \cite{unal2022scribbleKITTI,jiang2021rellis3D,pan2020semanticPOSS,xiao2023semanticSTF,xiao2022synLiDAR,klokov2023daps3D,saltori2020synth4D}, respectively, with specific data portions. All scores are given in percentage (\%). The \textbf{best} and \underline{2nd best} scores from each metric are highlighted in \textbf{bold} and \underline{underline}.}
    \vspace{-0.25cm}
    \label{tab:generalization}
    \resizebox{\linewidth}{!}{
    \begin{tabular}{r|r|cc|cc|cc|cc|cc|cc|cc}
    \toprule
    \multirow{2}{*}{\textbf{Method}} & \multirow{2}{*}{\textbf{Venue}} & \multicolumn{2}{c|}{\textbf{ScriKITTI}} & \multicolumn{2}{c|}{\textbf{Rellis-3D}} & \multicolumn{2}{c|}{\textbf{SemPOSS}} & \multicolumn{2}{c|}{\textbf{SemSTF}} & \multicolumn{2}{c|}{\textbf{SynLiDAR}} & \multicolumn{2}{c|}{\textbf{DAPS-3D}} & \multicolumn{2}{c}{\textbf{Synth4D}}
    \\
    & & \textbf{1\%} & \textbf{10\%} & \textbf{1\%} & \textbf{10\%} & \textbf{Half} & \textbf{Full} & \textbf{Half} & \textbf{Full} & \textbf{1\%} & \textbf{10\%} & \textbf{Half} & \textbf{Full} & \textbf{1\%} & \textbf{10\%}
    \\\midrule\midrule
    \cellcolor{moe_gray!18}\textcolor{gray}{Random} & \cellcolor{moe_gray!18}\textcolor{gray}{-} & \cellcolor{moe_gray!18}\textcolor{gray}{$23.81$} & \cellcolor{moe_gray!18}\textcolor{gray}{$47.60$} & \cellcolor{moe_gray!18}\textcolor{gray}{$38.46$} & \cellcolor{moe_gray!18}\textcolor{gray}{$53.60$} & \cellcolor{moe_gray!18}\textcolor{gray}{$46.26$} & \cellcolor{moe_gray!18}\textcolor{gray}{$54.12$} & \cellcolor{moe_gray!18}\textcolor{gray}{$48.03$} & \cellcolor{moe_gray!18}\textcolor{gray}{$48.15$} & \cellcolor{moe_gray!18}\textcolor{gray}{$19.89$} & \cellcolor{moe_gray!18}\textcolor{gray}{$44.74$} & \cellcolor{moe_gray!18}\textcolor{gray}{$74.32$} & \cellcolor{moe_gray!18}\textcolor{gray}{$79.38$} & \cellcolor{moe_gray!18}\textcolor{gray}{$20.22$} & \cellcolor{moe_gray!18}\textcolor{gray}{$66.87$}
    \\\midrule
    PPKT \cite{liu2021ppkt} & arXiv'21 & $36.50$ & $51.67$ & $49.71$ & $54.33$ & $50.18$ & $56.00$ & $50.92$ & $54.69$ & $37.57$ & $46.48$ & $78.90$ & $84.00$ & $61.10$ & $62.41$
    \\
    SLidR \cite{sautier2022slidr} & CVPR'22 & $39.60$ & $50.45$ & $49.75$ & $54.57$ & $51.56$ & $55.36$ & $52.01$ & $54.35$ & $42.05$ & $47.84$ & $81.00$ & $85.40$ & $63.10$ & $62.67$
    \\
    + \textsf{\textcolor{moe_red}{Li}\textcolor{moe_green}{MoE}} & \textbf{Ours} & \cellcolor{moe_green!12}$41.48$ & \cellcolor{moe_green!12}$53.41$ & \cellcolor{moe_green!12}$51.28$ & \cellcolor{moe_green!12}$55.21$ & \cellcolor{moe_green!12}$53.14$ & \cellcolor{moe_green!12}$56.42$ & \cellcolor{moe_green!12}$53.16$ & \cellcolor{moe_green!12}$55.51$ & \cellcolor{moe_green!12}$43.72$ & \cellcolor{moe_green!12}$49.57$ & \cellcolor{moe_green!12}$81.70$ & \cellcolor{moe_green!12}$85.76$ & \cellcolor{moe_green!12}$64.69$ & \cellcolor{moe_green!12}$66.79$
    \\
    Seal \cite{liu2023seal} & NeurIPS'23 & $40.64$ & $52.77$ & $51.09$ & $55.03$ & $53.26$ & $56.89$ & $53.46$ & $55.36$ & $43.58$ & $49.26$ & $81.88$ & $85.90$ & $64.50$ & $66.96$
    \\
    SuperFlow \cite{xu2024superflow} & ECCV'24 & \underline{$42.70$} & \underline{$54.00$} & \underline{$52.83$} & \underline{$55.71$} & \underline{$54.41$} & \underline{$57.33$} & \underline{$54.72$} & \underline{$56.57$} & \underline{$44.85$} & \underline{$51.38$} & \underline{$82.43$} & \underline{$86.21$} & \underline{$65.31$} & \underline{$69.43$}
    \\
    + \textsf{\textcolor{moe_red}{Li}\textcolor{moe_green}{MoE}} & \textbf{Ours} & \cellcolor{moe_green!12}$\mathbf{43.95}$ & \cellcolor{moe_green!12}$\mathbf{55.96}$ & \cellcolor{moe_green!12}$\mathbf{53.74}$ & \cellcolor{moe_green!12}$\mathbf{56.67}$ & \cellcolor{moe_green!12}$\mathbf{55.42}$ & \cellcolor{moe_green!12}$\mathbf{57.83}$ & \cellcolor{moe_green!12}$\mathbf{55.60}$ & \cellcolor{moe_green!12}$\mathbf{57.31}$ & \cellcolor{moe_green!12}$\mathbf{45.79}$ & \cellcolor{moe_green!12}$\mathbf{52.27}$ & \cellcolor{moe_green!12}$\mathbf{83.24}$ & \cellcolor{moe_green!12}$\mathbf{86.68}$ & \cellcolor{moe_green!12}$\mathbf{66.54}$ & \cellcolor{moe_green!12}$\mathbf{71.07}$
    \\\bottomrule
\end{tabular}}
\vspace{-0.2cm}
\end{table*}

\noindent\textbf{Training Objective.}
Let $\mathcal{X} \in \mathbb{R}^{N}$ denote the semantic labels for the point cloud $\mathcal{P}$. For each representation, these labels can be projected as follows: $\mathcal{X}_{\textcolor{moe_green}{\mathbf{r}}} \in \mathbb{R}^{H_{r} \times W_{r}}$ for the range image, $\mathcal{X}_{\textcolor{moe_red}{\mathbf{v}}} \in \mathbb{R}^{M}$ for the sparse voxels, and $\mathcal{X}_{\textcolor{moe_blue}{\mathbf{p}}} \in \mathbb{R}^{N}$ for raw points. Each representation is fine-tuned independently with a supervised loss that aligns the predicted logits with the corresponding projected semantic labels. In addition, we apply an MoE-based supervised loss at the output to further refine the model by selecting the most relevant features from each representation. The overall loss is:
\begin{equation}
    \mathcal{L}_{\text{SMS}} = \mathcal{L}_{\text{d}}(\mathcal{X}, \mathbf{Y}_{\text{moe}}) ~~+ \sum_{l \in \{\textcolor{moe_green}{\mathbf{r}},\textcolor{moe_red}{\mathbf{v}},\textcolor{moe_blue}{\mathbf{p}}\}} \mathcal{L}_{l}(\mathcal{X}_{l}, \mathbf{Y}_{l})~,
\end{equation}
where each $\mathcal{L}_{l}$ (for $l \in \{d,\textcolor{moe_green}{\mathbf{r}},\textcolor{moe_red}{\mathbf{v}},\textcolor{moe_blue}{\mathbf{p}}\}$) is a weighted combination of Cross-Entropy, Lovasz-Softmax \cite{berman2018lovasz}, and Boundary loss \cite{razani2021lite}, contributing to a comprehensive optimization of the segmentation performance across all representations.

\section{Experiments}
\label{sec:experiments}

\begin{figure*}[t]
    \centering
    \includegraphics[width=\linewidth]{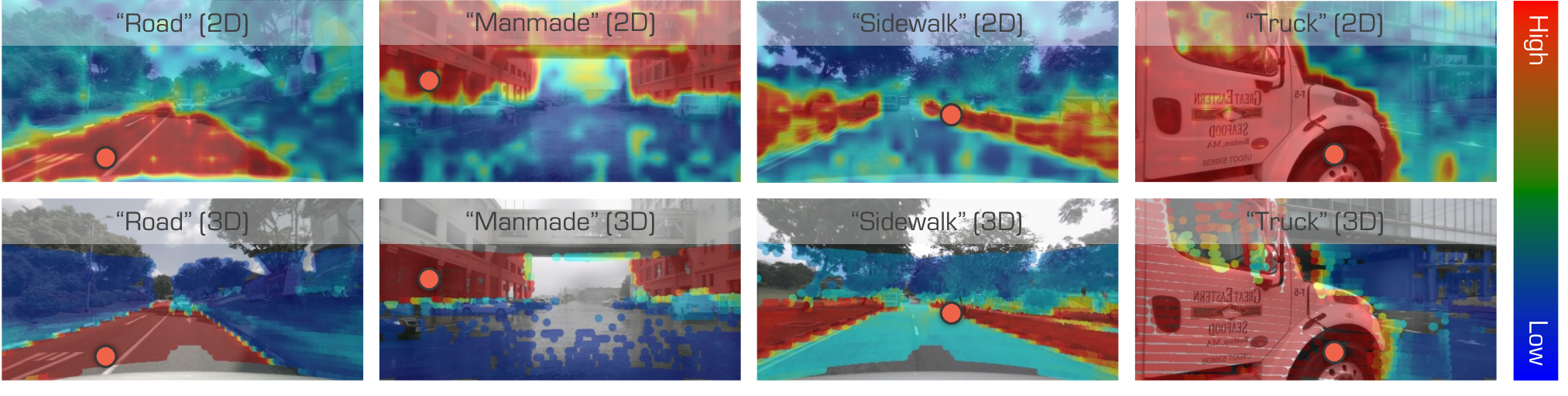}
    \vspace{-0.65cm}
    \caption{\textbf{Cosine similarity between learned features} of a query point (denoted as the \textcolor{moe_red}{\textbf{red}} dot) and: (1) the features of the image of the same scene (the first row); and (2) the features of the LiDAR points projected onto the image (the second row). Best viewed in colors.}
    \label{fig:similarity}
    \vspace{-0.15cm}
\end{figure*}

\begin{table*}[t]
    \centering
    \caption{\textbf{Out-of-distribution robustness assessment} of LiDAR pretraining methods under corruptions and sensor failures in the \textit{nuScenes-C} dataset from the \textit{Robo3D} benchmark \cite{kong2023robo3d}. \textbf{Full} denotes fine-tuning with full labels. \textbf{LP} denotes linear probing with frozen backbones. All mCE, mRR, and mIoU scores are given in percentage (\%). The \textbf{best} and \underline{2nd best} scores are highlighted in \textbf{bold} and \underline{underline}.}
    \vspace{-0.2cm}
    \label{tab:robo3d}
    \resizebox{\linewidth}{!}{
    \begin{tabular}{c|r|r|p{32pt}<{\centering}|p{32pt}<{\centering}|p{32pt}<{\centering}p{32pt}<{\centering}p{32pt}<{\centering}p{32pt}<{\centering}p{32pt}<{\centering}p{32pt}<{\centering}p{32pt}<{\centering}p{32pt}<{\centering}|c}
    \toprule
    \textbf{\#} & \textbf{Method} & \textbf{Venue} & \textbf{mCE}$\downarrow$ & \textbf{mRR}$\uparrow$ & ~\textbf{Fog}$\uparrow$~ & \textbf{Rain}$\uparrow$ & \textbf{Snow}$\uparrow$ & \textbf{Blur}$\uparrow$ & \textbf{Beam}$\uparrow$ & \textbf{Cross}$\uparrow$ & \textbf{Echo}$\uparrow$ & \textbf{Sensor}$\uparrow$ & \textbf{Average}$\uparrow$
    \\\midrule\midrule
    \cellcolor{moe_gray!18}\textbf{\textcolor{gray}{Full}} & \cellcolor{moe_gray!18}\textcolor{gray}{Random} & \cellcolor{moe_gray!18}\textcolor{gray}{-} & \cellcolor{moe_gray!18}\textcolor{gray}{$112.20$} & \cellcolor{moe_gray!18}\textcolor{gray}{$72.57$} & \cellcolor{moe_gray!18}\textcolor{gray}{$62.96$} & \cellcolor{moe_gray!18}\textcolor{gray}{$70.65$} & \cellcolor{moe_gray!18}\textcolor{gray}{$55.48$} & \cellcolor{moe_gray!18}\textcolor{gray}{$51.71$} & \cellcolor{moe_gray!18}\textcolor{gray}{$62.01$} & \cellcolor{moe_gray!18}\textcolor{gray}{$31.56$} & \cellcolor{moe_gray!18}\textcolor{gray}{$59.64$} & \cellcolor{moe_gray!18}\textcolor{gray}{$39.41$} & \cellcolor{moe_gray!18}\textcolor{gray}{$54.18$}
    \\\midrule
    \multirow{6}{*}{\textbf{Full}} & PPKT \cite{liu2021ppkt} & arXiv'21 & $105.64$ & $75.87$ & $64.01$ & $72.18$ & $59.08$ & $57.17$ & $63.88$ & $36.34$ & $60.59$ & $39.57$ & $56.60$
    \\
    & SLidR \cite{sautier2022slidr} & CVPR'22 & $106.08$ & $75.99$ & $65.41$ & $72.31$ & $56.01$ & $56.07$ & $62.87$ & $41.94$ & $61.16$ & $38.90$ & $56.83$
    \\
    & + \textsf{\textcolor{moe_red}{Li}\textcolor{moe_green}{MoE}} & \textbf{Ours} & \cellcolor{moe_green!12}$101.74$ & \cellcolor{moe_green!12}$77.77$ & \cellcolor{moe_green!12}$67.92$ & \cellcolor{moe_green!12}$73.25$ & \cellcolor{moe_green!12}$57.02$ & \cellcolor{moe_green!12}$56.30$ & \cellcolor{moe_green!12}$64.72$ & \cellcolor{moe_green!12}$44.81$ & \cellcolor{moe_green!12}\underline{$61.23$} & \cellcolor{moe_green!12}$45.37$ & \cellcolor{moe_green!12}$58.83$
    \\
    & Seal \cite{liu2023seal} & NeurIPS'23 & $92.63$ & $83.08$ & $\mathbf{72.66}$ & $74.31$ & $\mathbf{66.22}$ & $\mathbf{66.14}$ & $65.96$ & $57.44$ & $59.87$ & $39.85$ & $62.81$
    \\
    & SuperFlow \cite{xu2024superflow} & ECCV'24 & \underline{$91.67$} & \underline{$83.17$} & $70.32$ & \underline{$75.77$} & $65.41$ & $61.05$ & \underline{$68.09$} & \underline{$60.02$} & $58.36$ & \underline{$50.41$} & \underline{$63.68$}
    \\
    & + \textsf{\textcolor{moe_red}{Li}\textcolor{moe_green}{MoE}} & \textbf{Ours} & \cellcolor{moe_green!12}$\mathbf{88.43}$ & \cellcolor{moe_green!12}$\mathbf{83.28}$ & \cellcolor{moe_green!12}\underline{$71.10$} & \cellcolor{moe_green!12}$\mathbf{75.92}$ & \cellcolor{moe_green!12}\underline{$65.66$} & \cellcolor{moe_green!12}\underline{$63.86$} & \cellcolor{moe_green!12}$\mathbf{68.52}$ & \cellcolor{moe_green!12}$\mathbf{60.78}$ & \cellcolor{moe_green!12}$\mathbf{61.91}$ & \cellcolor{moe_green!12}$\mathbf{50.66}$ & \cellcolor{moe_green!12}$\mathbf{64.80}$
    \\\midrule
    \multirow{6}{*}{\textbf{LP}} & PPKT \cite{liu2021ppkt} & arXiv'21 & $183.44$ & \underline{$78.15$} & $30.65$ & $35.42$ & $28.12$ & $29.21$ & $32.82$ & $19.52$ & $28.01$ & $20.71$ & $28.06$
    \\
    & SLidR \cite{sautier2022slidr} & CVPR'22 & $179.38$ & $77.18$ & $34.88$ & $38.09$ & $32.64$ & $26.44$ & $33.73$ & $20.81$ & $31.54$ & $21.44$ & $29.95$
    \\
    & + \textsf{\textcolor{moe_red}{Li}\textcolor{moe_green}{MoE}} & \textbf{Ours} & \cellcolor{moe_green!12}$163.75$ & \cellcolor{moe_green!12}$75.49$ & \cellcolor{moe_green!12}$37.29$ & \cellcolor{moe_green!12}$43.41$ & \cellcolor{moe_green!12}$36.04$ & \cellcolor{moe_green!12}$38.33$ & \cellcolor{moe_green!12}$40.66$ & \cellcolor{moe_green!12}$22.46$ & \cellcolor{moe_green!12}$37.61$ & \cellcolor{moe_green!12}$25.38$ & \cellcolor{moe_green!12}$35.15$
    \\
    & Seal \cite{liu2023seal} & NeurIPS'23 & $166.18$ & $75.38$ & $37.33$ & $42.77$ & $29.93$ & $37.73$ & $40.32$ & $20.31$ & $37.73$ & $24.94$ & $33.88$
    \\
    & SuperFlow \cite{xu2024superflow} & ECCV'24 & \underline{$161.78$} & $75.52$ & \underline{$37.59$} & \underline{$43.42$} & \underline{$37.60$} & \underline{$39.57$} & \underline{$41.40$} & \underline{$23.64$} & \underline{$38.03$} & \underline{$26.69$} & \underline{$35.99$}
    \\
    & + \textsf{\textcolor{moe_red}{Li}\textcolor{moe_green}{MoE}} & \textbf{Ours} & \cellcolor{moe_green!12}$\mathbf{155.77}$ & \cellcolor{moe_green!12}$\mathbf{78.23}$ & \cellcolor{moe_green!12}$\mathbf{40.35}$ & \cellcolor{moe_green!12}$\mathbf{45.28}$ & \cellcolor{moe_green!12}$\mathbf{39.14}$ & \cellcolor{moe_green!12}$\mathbf{42.10}$ & \cellcolor{moe_green!12}$\mathbf{44.21}$ & \cellcolor{moe_green!12}$\mathbf{27.33}$ & \cellcolor{moe_green!12}$\mathbf{39.20}$ & \cellcolor{moe_green!12}$\mathbf{29.49}$ & \cellcolor{moe_green!12}$\mathbf{38.39}$
    \\\bottomrule
\end{tabular}}
\vspace{-0.4cm}
\end{table*}

\subsection{Configurations}

\noindent\textbf{Datasets.}
Following the practical settings \cite{liu2023seal,xu2024superflow}, we use the \textit{nuScenes} \cite{caesar2020nuScenes} for pretraining and evaluate it across 11 LiDAR semantic segmentation datasets, including \textit{nuScenes} \cite{fong2022panoptic-nuScenes}, \textit{SemanticKITTI} \cite{behley2019semanticKITTI}, \textit{Waymo Open} \cite{sun2020waymoOpen}, \textit{ScribbleKITTI} \cite{unal2022scribbleKITTI}, \textit{RELLIS-3D} \cite{jiang2021rellis3D}, \textit{SemanticPOSS} \cite{pan2020semanticPOSS}, \textit{SemanticSTF} \cite{xiao2023semanticSTF}, \textit{SynLiDAR} \cite{xiao2022synLiDAR}, \textit{DAPS-3D} \cite{klokov2023daps3D}, \textit{Synth4D} \cite{saltori2020synth4D}, and \textit{Robo3D} \cite{kong2023robo3d}. More details are in the Appendix.

\noindent\textbf{Implementation Details.}
All experiments are conducted using the MMDetection3D \cite{mmdet3d} codebase. For each LiDAR representation, we select FRNet \cite{xu2025frnet} (\textcolor{moe_green}{range}), MinkUNet \cite{choy2019minkowski} (\textcolor{moe_red}{voxel}), and SPVCNN \cite{tang2020spvcnn} (\textcolor{moe_blue}{point}) as the backbones. For image-to-LiDAR pretraining, we adopt the SLidR \cite{sautier2022slidr} and SuperFlow \cite{xu2024superflow} paradigms. The learning rate for pretraining is set to 0.01, while the learning rate for the second CML stage is 0.001. In the third stage, we use a separate learning rate scheme: 0.001 for the backbone and 0.01 for other parts. We use the AdamW optimizer \cite{loshchilov2018adamw} and a OneCycle scheduler \cite{onecycle} for all stages. In line with standard protocol, we report the mean Intersection-over-Union (mIoU) for segmentation, as well as mean Corruption Error (mCE) and mean Resilience Rate (mRR) for robustness.

\begin{figure*}[t]
    \centering
    \begin{subfigure}[h]{0.49\textwidth}
        \centering
        \includegraphics[width=\linewidth]{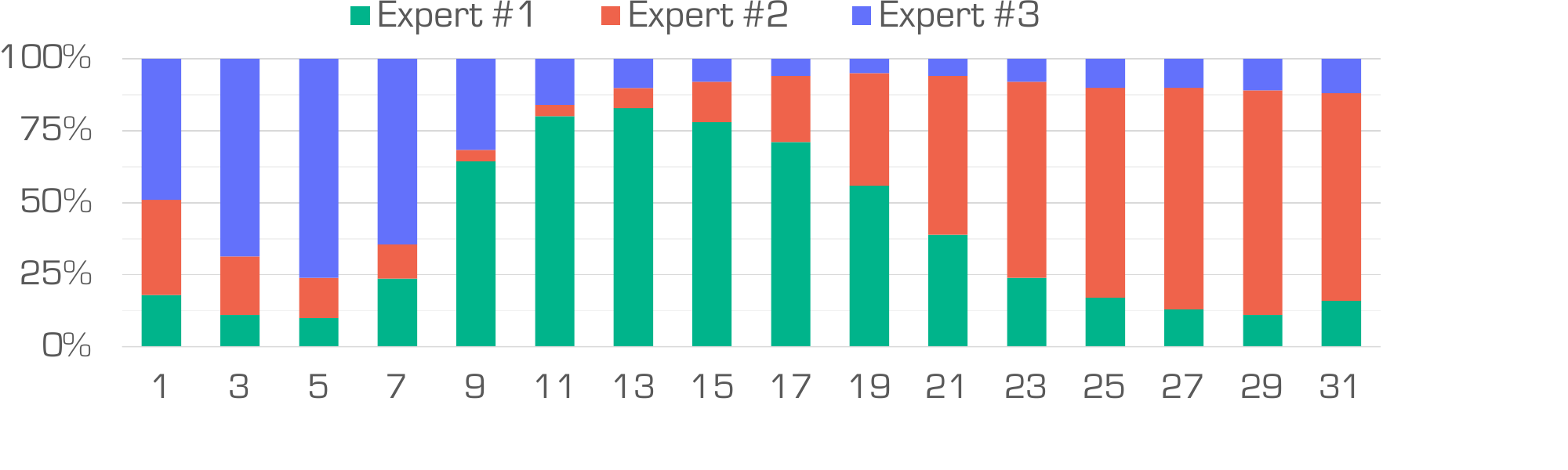}
        \caption{Expert Loadings (Beam Number)}
        \label{fig:distribution_beam}
    \end{subfigure}~~
    \begin{subfigure}[h]{0.49\textwidth}
        \centering
        \includegraphics[width=\linewidth]{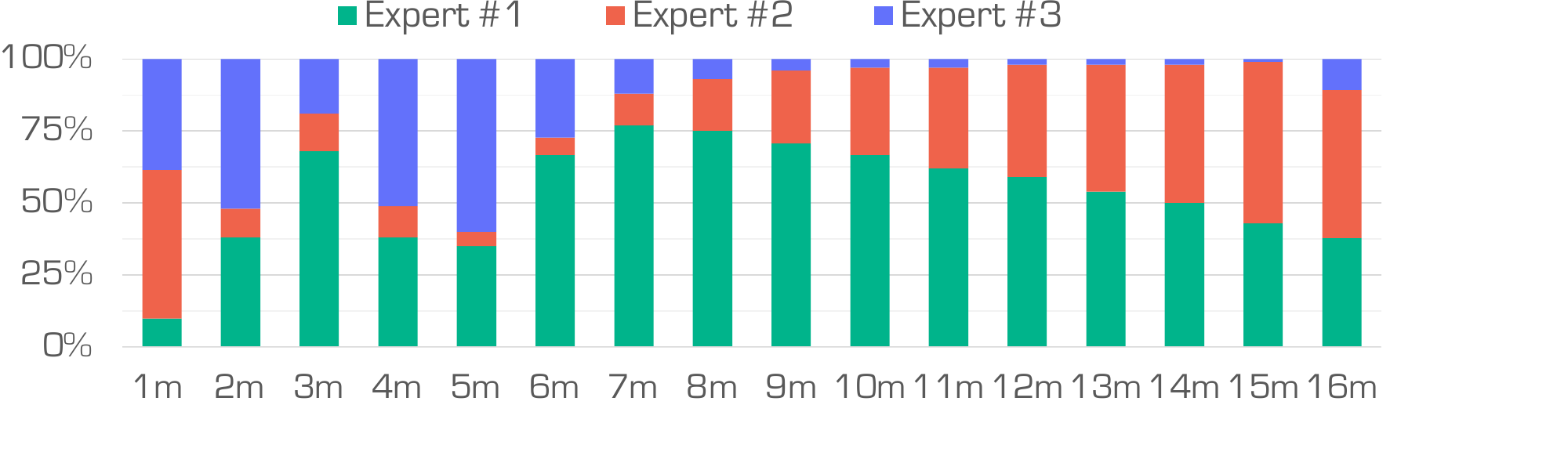}
        \caption{Expert Loadings (Distances to Ego)}
        \label{fig:distribution_distance}
    \end{subfigure}
    \vspace{-0.3cm}
    \caption{\textbf{Ablation study on distributions of expert loadings} in CML. The distributions are based on (a) LiDAR beam numbers and (b) distances. The three experts are \textcolor{moe_green}{\texttt{\#1}} range view, \textcolor{moe_red}{\texttt{\#2}} voxel, and \textcolor{moe_blue}{\texttt{\#3}} point, respectively.}
    \label{fig:ablation_distribution}
    \vspace{-0.3cm}
\end{figure*}

\subsection{Comparative Study}
\label{sec:comparative}

\noindent\textbf{Comparisons to State of the Arts.}
In \cref{tab:benchmark}, we compare \textsf{\textcolor{moe_red}{Li}\textcolor{moe_green}{MoE}} with state-of-the-art pretraining methods using both Linear Probing (LP) and fine-tuning on various portions of the \textit{nuScenes} \cite{fong2022panoptic-nuScenes} dataset. In the LP setting, where the backbone is frozen and only the task-specific head is trained, our approach outperforms single-representation methods \cite{sautier2022slidr,xu2024superflow}, achieving mIoU gains from $1.10\%$ to $1.76\%$. In the fine-tuning setting, where both the backbone and task head are updated, we observe even greater improvements, with mIoU gains ranging from $0.66\%$ to $5.66\%$. These results highlight the advantages of combining multiple representations, leading to enhanced performance.

\noindent\textbf{Cross-Dataset Knowledge Transfer.}
To evaluate the scalability of \textsf{\textcolor{moe_red}{Li}\textcolor{moe_green}{MoE}}, we conduct a comprehensive study across nine diverse 3D semantic segmentation datasets. Notably, except for the SemanticKITTI \cite{behley2019semanticKITTI} and ScribbleKITTI \cite{unal2022scribbleKITTI} datasets, the remaining seven datasets cannot be effectively converted into range images due to the unknown FoV parameters of their LiDAR sensors. As a result, for these datasets, we combine only voxel and point representations in the SMS stage, since range image representations are not feasible. As shown in \cref{tab:benchmark} and \cref{tab:generalization}, our method consistently outperforms single-representation methods across all datasets, demonstrating superior adaptability and robustness for a wide range of 3D segmentation tasks.

\noindent\textbf{Robustness Assessments.}
The robustness of a 3D perception model against unprecedented conditions is crucial for its development in real-world applications. In \cref{tab:robo3d}, we compare \textsf{\textcolor{moe_red}{Li}\textcolor{moe_green}{MoE}} with previous pretraining methods using the \textit{nuScenes-C} dataset from the Robo3D benchmark \cite{kong2023robo3d}. The results show that our method excels in handling various types of corruption, demonstrating its robustness and resilience under challenging real-world conditions.

\noindent\textbf{Qualitative Results.}
\cref{fig:similarity} illustrates the similarity between a query point and the pretrained 2D image backbone, as well as other LiDAR points during the CML stage. The visual results confirm that combining multiple representations enhances the semantic features during pretraining, resulting in more effective feature learning.

\begin{table}[t]
    \centering
    \caption{\textbf{Ablation study on experts} in CML. Only sparse voxel representation is employed for the downstream task. All scores are given in percentage (\%). The \textbf{best} scores are highlighted in \textbf{bold}.}
    \vspace{-0.2cm}
    \label{tab:ablation_pretrain}
    \resizebox{\linewidth}{!}{
    \begin{tabular}{c|p{28pt}<{\centering}p{28pt}<{\centering}p{28pt}<{\centering}|cc|c}
        \toprule
        \multirow{2}{*}{\textbf{Pretrain}} & \multicolumn{3}{c|}{\textbf{Experts}} & \multicolumn{2}{c|}{\textbf{nuScenes}} & \textbf{KITTI}
        \\
        & \textbf{\textcolor{moe_red}{Voxel}} & \textbf{\textcolor{moe_blue}{Point}} & \textbf{\textcolor{moe_green}{Range}} & \textbf{LP} & \textbf{1\%} & \textbf{1\%}
        \\\midrule\midrule
        \cellcolor{moe_gray!18}\textcolor{gray}{Random} & \cellcolor{moe_gray!18}\textcolor{gray}{\cmark} & \cellcolor{moe_gray!18}\textcolor{gray}{\xmark} & \cellcolor{moe_gray!18}\textcolor{gray}{\xmark} & \cellcolor{moe_gray!18}\textcolor{gray}{$8.10$} & \cellcolor{moe_gray!18}\textcolor{gray}{$30.30$} & \cellcolor{moe_gray!18}\textcolor{gray}{$39.50$}
        \\\midrule
        SLidR~\cite{sautier2022slidr} & \textcolor{moe_red}{\cmark} & \textcolor{moe_blue}{\xmark} & \textcolor{moe_green}{\xmark} & $45.35$ & $41.64$ & $45.50$
        \\\midrule
        \multirow{4}{*}{\textsf{\textcolor{moe_red}{Li}\textcolor{moe_green}{MoE}}} & \textcolor{moe_red}{\cmark} & \textcolor{moe_blue}{\cmark} & \textcolor{moe_green}{\xmark} & $45.73$ & $43.28$ & $46.54$
        \\
        & \textcolor{moe_red}{\cmark} & \textcolor{moe_blue}{\xmark} & \textcolor{moe_green}{\cmark} & $45.89$ & $43.81$ & $46.60$
        \\
        & \textcolor{moe_red}{\xmark} & \textcolor{moe_blue}{\cmark} & \textcolor{moe_green}{\cmark} & $45.56$ & $42.56$ & $46.21$
        \\
        & \cellcolor{moe_green!12}\textcolor{moe_red}{\cmark} & \cellcolor{moe_green!12}\textcolor{moe_blue}{\cmark} & \cellcolor{moe_green!12}\textcolor{moe_green}{\cmark} & \cellcolor{moe_green!12}$\mathbf{46.02}$ & \cellcolor{moe_green!12}$\mathbf{44.85}$ & \cellcolor{moe_green!12}$\mathbf{46.92}$
        \\\bottomrule
    \end{tabular}}
    \vspace{-0.4cm}
\end{table}

\begin{table}[t]
    \centering
    \caption{\textbf{Ablation study on experts} in SMS. All scores are given in percentage (\%). The \textbf{best} scores are highlighted in \textbf{bold}.}
    \vspace{-0.2cm}
    \label{tab:ablation_downstream}
    \resizebox{\linewidth}{!}{
    \begin{tabular}{c|p{31pt}<{\centering}p{31pt}<{\centering}p{31pt}<{\centering}|cc|c}
        \toprule
        \multirow{2}{*}{\textbf{Pretrain}} & \multicolumn{3}{c|}{\textbf{Experts}} & \multicolumn{2}{c|}{\textbf{nuScenes}} & \textbf{KITTI}
        \\
        & \textbf{\textcolor{moe_red}{Voxel}} & \textbf{\textcolor{moe_blue}{Point}} & \textbf{\textcolor{moe_green}{Range}} & \textbf{LP} & \textbf{1\%} & \textbf{1\%}
        \\\midrule\midrule
        \cellcolor{moe_gray!18} & \cellcolor{moe_gray!18}\textcolor{gray}{\cmark} & \cellcolor{moe_gray!18}\textcolor{gray}{\xmark} & \cellcolor{moe_gray!18}\textcolor{gray}{\xmark} & \cellcolor{moe_gray!18}\textcolor{gray}{$8.10$} & \cellcolor{moe_gray!18}\textcolor{gray}{$30.30$} & \cellcolor{moe_gray!18}\textcolor{gray}{$39.50$}
        \\
        \cellcolor{moe_gray!18}\textcolor{gray}{Random} & \cellcolor{moe_gray!18}\textcolor{gray}{\xmark} & \cellcolor{moe_gray!18}\textcolor{gray}{\cmark} & \cellcolor{moe_gray!18}\textcolor{gray}{\xmark} & \cellcolor{moe_gray!18}\textcolor{gray}{$8.80$} & \cellcolor{moe_gray!18}\textcolor{gray}{$34.79$} & \cellcolor{moe_gray!18}\textcolor{gray}{$38.74$}
        \\
        \cellcolor{moe_gray!18} & \cellcolor{moe_gray!18}\textcolor{gray}{\xmark} & \cellcolor{moe_gray!18}\textcolor{gray}{\xmark} & \cellcolor{moe_gray!18}\textcolor{gray}{\cmark} & \cellcolor{moe_gray!18}\textcolor{gray}{$11.85$} & \cellcolor{moe_gray!18}\textcolor{gray}{$29.56$} & \cellcolor{moe_gray!18}\textcolor{gray}{$39.20$}
        \\\midrule
        \multirow{8}{*}{\textsf{\textcolor{moe_red}{Li}\textcolor{moe_green}{MoE}}} & \textcolor{moe_red}{\cmark} & \textcolor{moe_blue}{\xmark} & \textcolor{moe_green}{\xmark} & $46.02$ & $44.85$ & $46.92$
        \\
        & \textcolor{moe_red}{\xmark} & \textcolor{moe_blue}{\cmark} & \textcolor{moe_green}{\xmark} & $45.97$ & $44.79$ & $47.06$
        \\
        & \textcolor{moe_red}{\xmark} & \textcolor{moe_blue}{\xmark} & \textcolor{moe_green}{\cmark} & $45.09$ & $40.88$ & $44.34$
        \\\cmidrule{2-7}
        & \textcolor{moe_red}{\cmark} & \textcolor{moe_blue}{\cmark} & \textcolor{moe_green}{\xmark} & $46.37$ & $45.74$ & $47.51$
        \\
        & \textcolor{moe_red}{\cmark} & \textcolor{moe_blue}{\xmark} & \textcolor{moe_green}{\cmark} & $46.28$ & $45.70$ & $47.35$
        \\
        & \textcolor{moe_red}{\xmark} & \textcolor{moe_blue}{\cmark} & \textcolor{moe_green}{\cmark} & $46.09$ & $45.32$ & $47.49$
        \\\cmidrule{2-7}
        & \cellcolor{moe_green!12}\textcolor{moe_red}{\cmark} & \cellcolor{moe_green!12}\textcolor{moe_blue}{\cmark} & \cellcolor{moe_green!12}\textcolor{moe_green}{\cmark} & \cellcolor{moe_green!12}$\mathbf{46.56}$ & \cellcolor{moe_green!12}$\mathbf{46.89}$ & \cellcolor{moe_green!12}$\mathbf{47.96}$
        \\\bottomrule
    \end{tabular}}
    \vspace{-0.4cm}
\end{table}

\subsection{Ablation Study}
\label{sec:ablation}

\begin{figure}
    \centering
    \includegraphics[width=\linewidth]{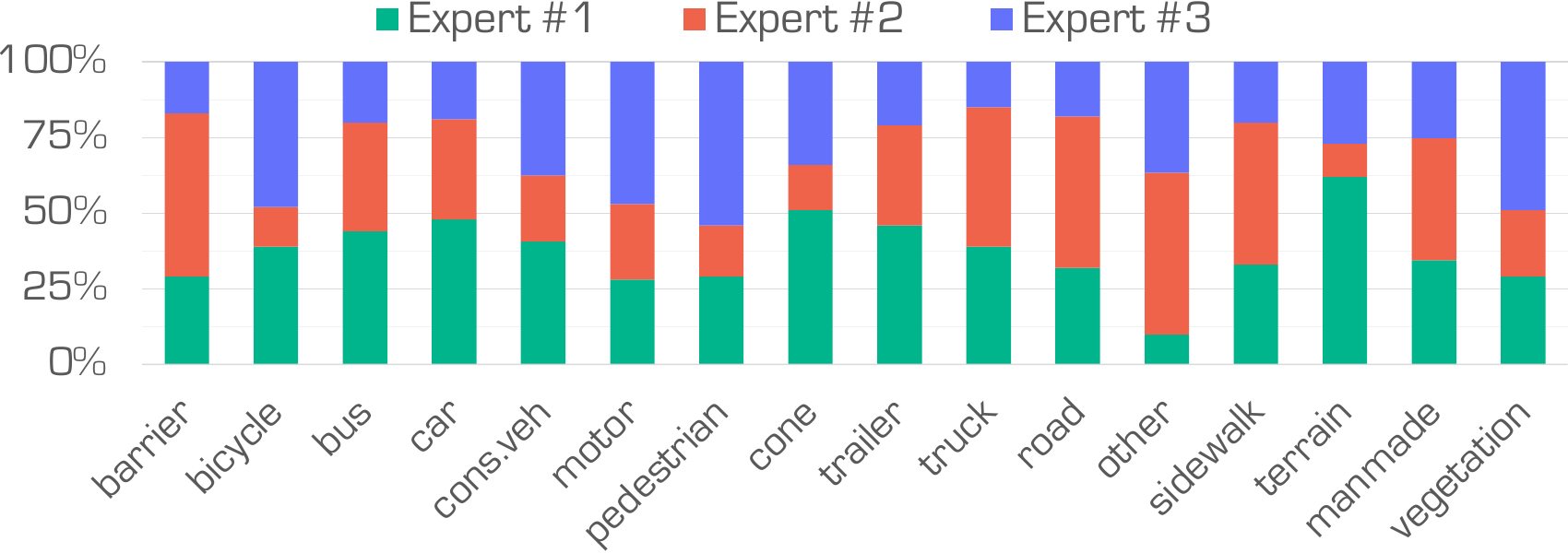}
    \vspace{-0.7cm}
    \caption{\textbf{Class-wise expert loadings} in SMS. The three experts are \textcolor{moe_green}{\texttt{\#1}} range view, \textcolor{moe_red}{\texttt{\#2}} voxel, and \textcolor{moe_blue}{\texttt{\#3}} point, respectively.}
    \label{fig:distribution_classes}
    \vspace{-0.3cm}
\end{figure}

\noindent\textbf{Route Activations.}
In \cref{fig:ablation_distribution}, we present the distributions of beam numbers and distances loaded by each representation during the CML stage. Range images predominantly capture middle beams and distances, sparse voxels focus on upper beams and longer distances, while points concentrate on lower beams and near distances. This distribution highlights the complementary nature of these representations, which provide a more comprehensive representation of the LiDAR data when combined in the MoE layer. In \cref{fig:distribution_classes}, we show the distributions of semantic classes loaded by each representation during the SMS stage. Range images are more sensitive to dynamic objects, sparse voxels highlight background objects, and raw points capture more detailed objects with complex structures. These results demonstrate how each representation contributes to the feature fusion.

\noindent\textbf{Scaling LiDAR Representations.}
In this ablation study, we explore the effect of combining multiple representations in both the CML and SMS stages. First, we test different combinations of representations for pretraining in the CML stage, using a sparse voxel-based network for the downstream task. The results shown in \cref{tab:ablation_pretrain} demonstrate that combining multiple representations consistently outperforms single-representation pretraining. Next, we explore various combinations in the SMS stage, where each representation is pretrained using the CML stage. As presented in \cref{tab:ablation_downstream}, incorporating multiple representations in the SMS stage further boosts performance, validating the advantages of multi-representation fusion in both pretraining and downstream tasks.

\noindent\textbf{Effect of Representation Diversity.} To evaluate the impact of representation diversity, we conduct an ablation study by replacing the three heterogeneous experts in our framework with three identical sparse voxel-based models. Compared to single-representation learning, which achieves $41.64$ and $45.50$ mIoU on nuScenes \cite{fong2022panoptic-nuScenes} and SemanticKITTI \cite{behley2019semanticKITTI} with only $1\%$ annotations, using three identical experts provides a minor improvement, reaching $42.72$ and $46.35$ mIoU, respectively. However, this configuration remains suboptimal compared to \textsf{\textcolor{moe_red}{Li}\textcolor{moe_green}{MoE}}, which achieves $46.89$ and $47.96$ mIoU. These results highlight the advantage of incorporating diverse representations, as the MoE layer can better exploit the unique characteristics and complementary features of each representation, leading to superior performance. 

\noindent\textbf{Downstream Efficiency.}
Finally, we analyze the impact of LiMoE on downstream efficiency. CML integrates multiple LiDAR representations into a unified network for pretraining, with no impact on downstream efficiency. In contrast, SMS combines multiple representations for downstream tasks, which introduces additional computational costs. Specifically, our framework (SMS) has $85.8$M parameters and achieves $8.3$ FPS, remaining high efficiency.

\section{Conclusion}
\label{sec:conclusion}

In this work, we introduced \textsf{\textcolor{moe_red}{Li}\textcolor{moe_green}{MoE}}, a novel framework designed to leverage multiple LiDAR representations for enhanced feature learning in LiDAR scenes. By combining range images, sparse voxels, and raw points through a Mixture of Experts framework, our approach captures complementary information from different representations, enabling more robust scene understanding. We proposed a three-stage feature learning process, consisting of Image-to-LiDAR Pretraining, Contrastive Mixture Learning (CML), and Semantic Mixture Supervision (SMS). Extensive experiments demonstrate that our design outperforms existing approaches across multiple benchmarks. We hope this work paves the way for more scalable and robust LiDAR-based perception systems for real-world applications.

\section*{Acknowledgments}
This work is supported in part by the Natural Science Foundation of China under Grants U24B20155 and U21B2044, and in part by the Jiangsu Province Science and Technology Project under Grants BA2022026 and BK20243051. This work is also supported by the Ministry of Education, Singapore, under its MOE AcRF Tier 2 (MOET2EP20221-0012, MOE-T2EP20223-0002), and under the RIE2020 Industry Alignment Fund – Industry Collaboration Projects (IAF-ICP) Funding Initiative, as well as cash and in-kind contribution from the industry partner(s). Lingdong Kong is supported by the Apple Scholars in AI/ML Ph.D. Fellowship program.

The authors would like to sincerely thank the Program Chairs, Area Chairs, and Reviewers for the time and efforts devoted during the review process.

\section*{Appendix}
\startcontents[appendices]
\printcontents[appendices]{l}{1}{\setcounter{tocdepth}{3}}

\section{Additional Implementation Details}

In this section, we provide additional details to facilitate the implementation and reproducibility of the methods within the proposed \textsf{\textcolor{moe_red}{Li}\textcolor{moe_green}{MoE}} framework.

\subsection{Datasets}

In this work, we conduct extensive experiments across a diverse set of LiDAR semantic segmentation datasets to validate the effectiveness of the proposed \textsf{\textcolor{moe_red}{Li}\textcolor{moe_green}{MoE}} framework.

\begin{itemize}
    \item \textbf{nuScenes} \cite{caesar2020nuScenes,fong2022panoptic-nuScenes} is a large-scale, multimodal dataset designed for autonomous driving, featuring six cameras, five radars, one LiDAR, along with IMU and GPS sensors. The dataset comprises $1,000$ driving scenes collected in Boston and Singapore. For the point cloud semantic segmentation task, it provides $1.4$ billion annotated points across $40,000$ point clouds, with each LiDAR point labeled into one of $32$ semantic categories. The point clouds are captured using a Velodyne HDL-32E LiDAR sensor. In this work, a mini-train split is created from the full training set for model pretraining during the Image-to-LiDAR and CML stages, adhering to the SLidR protocol \cite{sautier2022slidr}. For the SMS stage, the training set is further split to generate subsets containing $1\%$, $5\%$, $10\%$, $25\%$, and $100\%$ of annotated scans for fine-tuning. More details about this dataset can be found at \url{https://nuscenes.org/nuscenes}.

    \item \textbf{SemanticKITTI} \cite{behley2019semanticKITTI} is a large-scale benchmark dataset tailored for semantic scene understanding in autonomous driving. The dataset was collected using a Velodyne HDL-64E LiDAR sensor, capturing diverse real-world scenarios such as urban traffic in city centers, residential neighborhoods, highways, and rural countryside roads around Karlsruhe, Germany. The dataset consists of $22$ densely labeled point cloud sequences derived from the KITTI Odometry benchmark \cite{geiger2012kitti}, with each point annotated into one of $28$ semantic categories. In this work, the training set is uniformly split to create a subset with $1\%$ of the scans for fine-tuning. More details about this dataset can be found at \url{https://semantic-kitti.org}.

    \item \textbf{Waymo Open} \cite{sun2020waymoOpen} is a large-scale, high-quality, and diverse dataset designed to advance perception in autonomous driving. The dataset features multimodal data collected using five high-resolution cameras and five LiDAR sensors. It includes $1,150$ driving scenes recorded across a variety of suburban and urban areas, captured at different times of the day to ensure diversity in lighting, weather, and traffic conditions. For the LiDAR semantic segmentation task, each point in the dataset is annotated into one of $23$ semantic categories. In this work, the training set is uniformly split to create a subset with $1\%$ of the scans for fine-tuning. More details about this dataset can be found at \url{https://waymo.com/open}.

    \item \textbf{ScribbleKITTI} \cite{unal2022scribbleKITTI} is a weakly supervised variant of the SemanticKITTI \cite{behley2019semanticKITTI} dataset, designed to advance research in semantic scene understanding with minimal annotation effort. Unlike SemanticKITTI, which provides dense, point-wise annotations for every LiDAR point, ScribbleKITTI employs sparse line scribble annotations as a cost-effective alternative. This approach drastically reduces annotation requirements, with the dataset containing approximately $189$ million labeled points -- around $8.06\%$ of the fully supervised dataset -- resulting in a $90\%$ reduction in annotation time. In this work, the training set is uniformly split to create subsets with $1\%$ and $10\%$ of the scans for fine-tuning. More details about this dataset can be found at \url{https://github.com/ouenal/scribblekitti}.

    \item \textbf{RELLIS-3D} \cite{jiang2021rellis3D} is a multimodal dataset curated for semantic scene understanding in complex off-road environments. It consists of five traversal sequences collected along three unpaved trails on the RELLIS Campus of Texas A\&M University. For the LiDAR semantic segmentation task, point-wise annotation was generated by projecting image-based semantic labels onto the point cloud using precise camera-LiDAR calibration. Each LiDAR point is categorized into one of $20$ semantic categories. In this work, the training set is uniformly split to create subsets with $1\%$ and $10\%$ of the scans for fine-tuning. More details about this dataset can be found at \url{http://www.unmannedlab.org/research/RELLIS-3D}.

    \item \textbf{SemanticPOSS} \cite{pan2020semanticPOSS} is a small-scale benchmark dataset designed for semantic segmentation, with a particular focus on dynamic instances in real-world off-road environments. The dataset was captured using a Hesai Pandora LiDAR sensor, a forward-facing color camera, and four wide-angle mono cameras. The data was collected on the campus of Peking University. SemanticKITTI comprises $7$ sequences, with each point annotated into one of $14$ semantic categories. In this work, we adopt sequences $00$ and $01$ as half of the annotated training scans and sequences $00$ to $05$ (excluding $02$ for validation) to create the full set of annotated scans for fine-tuning. More details about this dataset can be found at \url{https://www.poss.pku.edu.cn/semanticposs.html}.

    \item \textbf{SemanticSTF} \cite{xiao2023semanticSTF} is a LiDAR point cloud dataset specifically designed to enable robust perception under adverse weather conditions, which is derived from the STF benchmark \cite{bijelic2020stf}. The dataset was collected using a Velodyne HDL-64 S3D LiDAR sensor and includes a diverse set of $2,076$ scans captured across various weather conditions: $694$ snowy, $637$ dense-foggy, $631$ light-foggy, and $114$ rainy scans. Each point in the dataset is labeled with one of $21$ semantic categories. In this work, the training set is uniformly split to create subsets with $50\%$ and $100\%$ of the scans for fine-tuning. More details about this dataset can be found at \url{https://github.com/xiaoaoran/SemanticSTF}.

    \item \textbf{SynLiDAR} \cite{xiao2022synLiDAR} is a synthetic LiDAR dataset generated from various virtual environments. The dataset was created using the Unreal Engine 4 platform, capturing diverse outdoor scenarios such as urban cities, towns, harbors, \etc. It consists of $13$ LiDAR sequences with a total of $198,396$ scans, with each point labeled into one of $32$ semantic categories. In this work, the training set is uniformly split to create subsets with $1\%$ and $10\%$ of the scans for fine-tuning. More details about this dataset can be found at \url{https://github.com/xiaoaoran/SynLiDAR}.

    \item \textbf{DAPS-3D} \cite{klokov2023daps3D} consists of two subsets: DAPS-1 and DAPS-2, both captured by a Ouster OS0 LiDAR sensor. DAPS-1 is semi-synthetic, generated to simulate various real-world cleaning tasks, while DAPS-2 was captured during a real field trip of a cleaning robot operating in the VDNH Park in Moscow. In this work, the training set from the DAPS-1 subset is uniformly split to create subsets with $50\%$ and $100\%$ of the scans for fine-tuning. More details about this dataset can be found at \url{https://github.com/subake/DAPS3D}.

    \item \textbf{Synth4D} \cite{saltori2020synth4D} is a synthetic dataset captured using a simulated HDL LiDAR sensor within the CARLA simulator. The dataset consists of two subsets, collected from a vehicle navigating through four distinct scenarios: town, highway, rural area, and city. In this work, the training set from the Synth4D-nuScenes subset is uniformly split to create subsets with $1\%$ and $10\%$ of the scans for fine-tuning. More details about this dataset can be found at \url{https://github.com/ saltoricristiano/gipso-sfouda}.

    \item \textbf{nuScenes-C} \cite{kong2023robo3d} is a dataset within the Robo3D benchmark, specifically designed to evaluate the robustness of 3D detectors and segmentors under out-of-distribution scenarios and natural corruptions commonly encountered in real-world environments. The dataset incorporates eight types of corruptions: ``fog'', ``wet ground'', ``snow'', ``motion blur'', ``beam missing'', ``crosstalk'', ``incomplete echo'', and ``cross-sensor'' scenarios. Each corruption type is simulated following physical principles or engineering guidelines and includes three severity levels: light, moderate, and heavy. More details about this dataset can be found at \url{https://github.com/ldkong1205/Robo3D}.
\end{itemize}

\subsection{Training Configuration}

In this subsection, we present the implementation details of the \textsf{\textcolor{moe_red}{Li}\textcolor{moe_green}{MoE}} framework, which is organized into three stages.
\begin{itemize}
    \item \textbf{Image-to-LiDAR Pretraining} focuses on transferring knowledge from image representations to LiDAR point clouds. This stage builds on the methodologies of SLidR \cite{sautier2022slidr} and SuperFlow \cite{xu2024superflow}. We employ the ViT \cite{dosovitskiy2021vit} architecture as the image backbone, pretrained using DINOv2 \cite{oquab2023dinov2}, with three variants: Small, Base, and Large. Input images are resized to $224 \times 448$ and augmented with random horizontal flipping. For the LiDAR-based backbone, we select FRNet \cite{xu2025frnet}, MinkUNet-34 \cite{choy2019minkowski}, and SPVCNN \cite{tang2020spvcnn}, corresponding to the \textcolor{moe_green}{range}, \textcolor{moe_red}{voxel}, and \textcolor{moe_blue}{point} representations, respectively. Point cloud augmentations include random flipping along horizontal and vertical axes (with a $50\%$ probability), rotation along the $z$-axis within the range of $-180^{\circ}$ to $180^{\circ}$, and scaling with a factor sampled uniformly from $[0.95, 1.05]$. The LiDAR-based networks are pretrained using eight GPUs for $50$ epochs with a batch size of $4$ per GPU. We initialize the learning rate to $0.01$ and employ the AdamW optimizer \cite{loshchilov2018adamw} with a OneCycle scheduler \cite{onecycle}.

    \item \textbf{Contrastive Mixture Learning (CML)} promotes the integration of diverse LiDAR representations into a unified feature space. In this stage, the pretrained \textcolor{moe_green}{range}, \textcolor{moe_red}{voxel}, and \textcolor{moe_blue}{point} networks are mixed through a Mixture of Experts (MoE) layer, leveraging their complementary strengths to form a cohesive single-representation network. To enhance representation diversity, LiDAR point clouds are augmented with varied parameters, generating multiple respective views for each representation. The network is pretrained on eight GPUs for $50$ epochs, with a batch size of $4$ per GPU. The initial learning rate is set to $0.001$, and training utilizes the AdamW optimizer \cite{loshchilov2018adamw} with a OneCycle scheduler \cite{onecycle}. The pseudo-code for CML is detailed in \cref{alg:cml}.

    \item \textbf{Semantic Mixture Supervision (SMS)} aims to improve downstream segmentation performance by fusing semantic logits from multiple representations under semantic label supervision. For individual representation supervision, the \textcolor{moe_green}{range} network is optimized using Cross-Entropy loss, Lovasz-Softmax loss \cite{berman2018lovasz}, and Boundary loss \cite{razani2021lite} with weights of $1.0$, $2.0$, and $1.0$, respectively. The \textcolor{moe_red}{voxel} network employs Cross-Entropy loss, Lovasz-Softmax loss \cite{berman2018lovasz}, weighted at $1.0$ and $2.0$, while the \textcolor{moe_blue}{point} network relies solely on Cross-Entropy loss. The MoE-fused logits are supervised exclusively with Cross-Entropy loss. The training is conducted on four GPUs for $100$ epochs, with a batch size of $4$ per GPU. The initial learning rate for each representation's backbone is set to $0.001$, and $0.01$ for all other parameters. The AdamW optimizer \cite{loshchilov2018adamw} and a OneCycle scheduler \cite{onecycle} are used for optimization. The pseudo-code for SMS is detailed in \cref{alg:sms}.
\end{itemize}

\subsection{Evaluation Configuration}

To evaluate the semantic segmentation performance across various semantic classes, we employ the widely used Intersection-over-Union (IoU) metric. The IoU score for a specific class is computed as follows:
\begin{equation}
    \label{eq:seg_evel}
    \text{IoU} = \frac{TP}{TP + FP + FN}~,
\end{equation}
where $TP$ (True Positive) denotes the number of points correctly classified as belonging to the class, $FP$ (False Positive) denotes the number of points incorrectly classified as belonging to the class, and $FN$ (False Negative) denotes the number of points belonging to the class but misclassified as another class. To assess overall segmentation performance, we report the mean IoU (mIoU), calculated as the average IoU across all semantic classes.

To evaluate robustness, we adopt the Corruption Error (CE) and Resilience Rate (RR) metrics, following the setup established in Robo3D \cite{kong2023robo3d}. The CE and RR for a specific corruption type are computed as follows:
\begin{equation}
    \label{eq:robo_eval}
    \text{CE} = \frac{\sum_{i=1}^{3}(1 - \text{IoU}_{i})}{\sum_{i=1}^{3}(1 - \text{IoU}_{i}^{\text{base}})}~, \quad \text{RR} = \frac{\sum_{i=1}^{3} \text{IoU}_{i}}{3 \times \text{IoU}_{\text{clean}}}~,
\end{equation}
where $\text{IoU}_{i}^{\text{base}}$ denotes the IoU score of the baseline model for the corresponding corruption severity, and $\text{IoU}_{\text{clean}}$ indices the IoU score on the ``clean'' evaluation set. To measure overall robustness, we report the mean CE (mCE) and mean RR (mRR), which are calculated as the average CE and RR values across all corruption types.

\begin{algorithm}[t]
\caption{CML, PyTorch-stype}
\label{alg:cml}
\definecolor{codeblue}{rgb}{0.25,0.5,0.5}
\definecolor{codekw}{rgb}{0.85,0.18,0.5}
\lstset{
  backgroundcolor=\color{white},
  basicstyle=\fontsize{7.5pt}{7.5pt}\ttfamily\selectfont,
  columns=fullflexible,
  breaklines=true,
  captionpos=b,
  commentstyle=\fontsize{7.5pt}{7.5pt}\color{codeblue},
  keywordstyle=\fontsize{7.5pt}{7.5pt}\color{codekw},
}
\begin{lstlisting}[language=python]
# Point2Range: convert point cloud to range image
# Point2Voxel: convert point cloud to sparse voxel
# Range2Point: project range image to point cloud
# Voxel2Point: project sparse voxel to point cloud
# Group: Group features according to superpoint
# P: point cloud with shape (N, L)
# SP: superpoint
# B_R, B_V, B_P: Range-view, sparse voxel, and point network
# B_S: Student network for distilling
# D: output channel for each representation network
# Cont: contrastive learning function

class MoE(nn.Module):

    def __init__(self, channels):
        super(MoE, self).__init__()
        self.fusion = nn.Linear(channels*3, channels)

        self.w_gate = nn.Parameter(
            torch.zeros(channels, 3),
            requires_grad=True)
        self.w_noise = nn.Parameter(
            torch.zeros(channels, 3),
            requires_grad=True)

        self.softplus = nn.Softplus()
        self.softmax = nn.Softmax(1)

    def forward(self, range_feats, voxel_feats, point_feats):
        # feature alignment
        range_feats = Range2Point(range_feats)
        voxel_feats = Voxel2Point(voxel_feats)
        fusion_feats = torch.cat(
            [range_feats, voxel_feats, point_feats],
            dim=-1)
        fusion_feats = self.fusion(fusion_feats)
        
        clean_logits = feats @ self.w_gate
        raw_noise_stddev = feats @ self.w_noise
        noise_stddev = self.softplus(raw_noise_stddev)
        noise_logits = torch.randn_like(clean_logits) * noise_stddev
        logits = clean_logits + noise_logits
        gates = self.softmax(logits)  # (N, 3)
        alpha, beta, gamma = gates[:, 0:1], gates[:, 1:2], gates[2:3]
        return alpha * range_feats + beta * voxel_feats + gamma * point_feats

moe_layer = MoE(D)
R = Point2Range(P)  # (H, W, L)
V = Point2Voxel(P)  # (M, L)
F_R, F_V, F_P = B_R(R), B_V(V), B_P(P)
moe_feats = moe_layer(F_R, F_V, F_P)
student_feats = B_S(P)
# generate superpoint embedding
K = Group(moe_feats, SP)
Q = Group(student_feats, SP)
# loss function
loss = Cont(K, Q)
\end{lstlisting}
\end{algorithm}

\begin{algorithm}[t]
\caption{SMS, PyTorch-stype}
\label{alg:sms}
\definecolor{codeblue}{rgb}{0.25,0.5,0.5}
\definecolor{codekw}{rgb}{0.85,0.18,0.5}
\lstset{
  backgroundcolor=\color{white},
  basicstyle=\fontsize{7.5pt}{7.5pt}\ttfamily\selectfont,
  columns=fullflexible,
  breaklines=true,
  captionpos=b,
  commentstyle=\fontsize{7.5pt}{7.5pt}\color{codeblue},
  keywordstyle=\fontsize{7.5pt}{7.5pt}\color{codekw},
}
\begin{lstlisting}[language=python]
# Point2Range: convert point cloud to range image
# Point2Voxel: convert point cloud to sparse voxel
# Range2Point: project range image to point cloud
# Voxel2Point: project sparse voxel to point cloud
# P: point cloud with shape (N, L)
# Y: point cloud semantic label with shape (N)
# B_R, B_V, B_P: Range-view, sparse voxel, and point network
# C: number of classes
# CE: loss function between gt and predict logits

class MoE(nn.Module):

    def __init__(self, channels):
        super(MoE, self).__init__()
        self.fusion = nn.Linear(channels*3, channels)

        self.w_gate = nn.Parameter(
            torch.zeros(channels, 3),
            requires_grad=True)
        self.w_noise = nn.Parameter(
            torch.zeros(channels, 3),
            requires_grad=True)

        self.softplus = nn.Softplus()
        self.softmax = nn.Softmax(1)

    def forward(self, range_feats, voxel_feats, point_feats):
        # feature alignment
        range_feats = Range2Point(range_feats)
        voxel_feats = Voxel2Point(voxel_feats)
        fusion_feats = torch.cat(
            [range_feats, voxel_feats, point_feats],
            dim=-1)
        fusion_feats = self.fusion(fusion_feats)

        if self.training:
            clean_logits = feats @ self.w_gate
            raw_noise_stddev = feats @ self.w_noise
            noise_stddev = self.softplus(raw_noise_stddev)
            noise_logits = torch.randn_like(clean_logits) * noise_stddev
            logits = clean_logits + noise_logits
        else:
            logits = clean_logits
        gates = self.softmax(logits)  # (N, 3)
        alpha, beta, gamma = gates[:, 0:1], gates[:, 1:2], gates[2:3]
        return alpha * range_feats + beta * voxel_feats + gamma * point_feats

moe_layer = MoE(C)
R, Y_R = Point2Range(P), Point2Range(Y)  # (H, W, L)
V, Y_V = Point2Voxel(P), Point2Voxel(Y)  # (M, L)
L_R, L_V, L_P = B_R(R), B_V(V), B_P(P)
moe_logits = moe_layer(F_R, F_V, F_P)
# loss function
loss = CE(L_R, Y_R) + CE(L_V, Y_V) + CE(L_P, Y) + CE(moe_logits, Y)
\end{lstlisting}
\end{algorithm}

\begin{figure*}[t]
    \centering
    \begin{subfigure}[h]{0.48\textwidth}
        \centering
        \includegraphics[width=\linewidth]{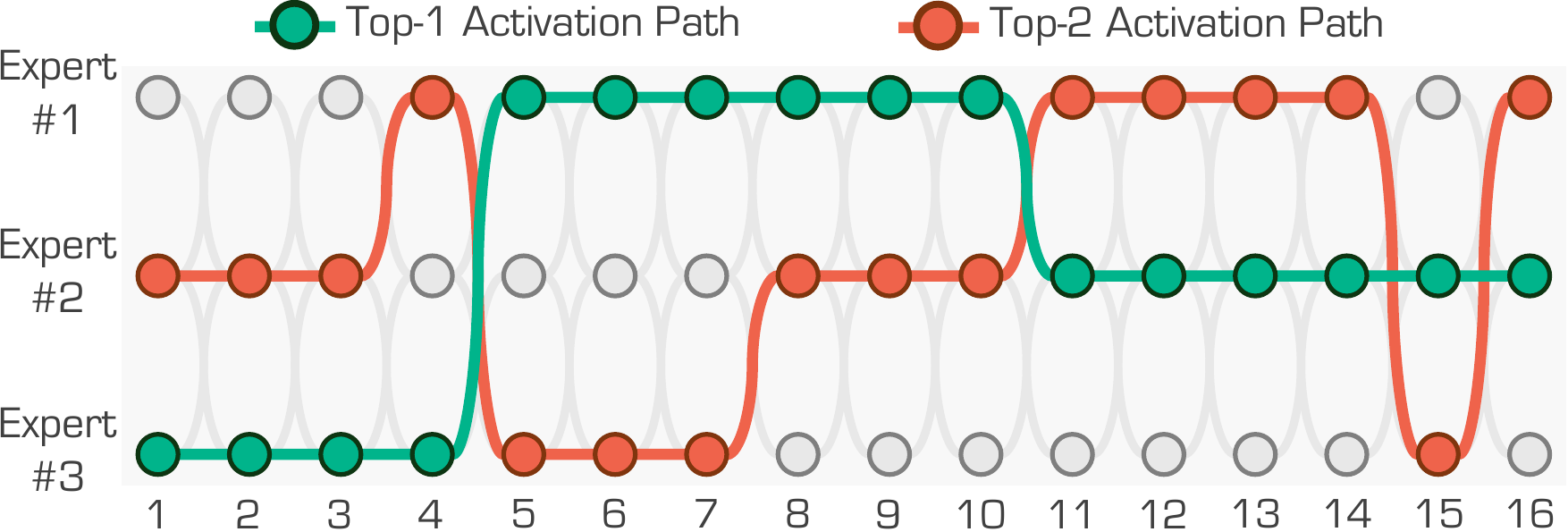}
        \caption{Expert Activation Paths (Beam Number)}
        \label{fig_supp:path_beam}
    \end{subfigure}~~~~~
    \begin{subfigure}[h]{0.48\textwidth}
        \centering
        \includegraphics[width=\linewidth]{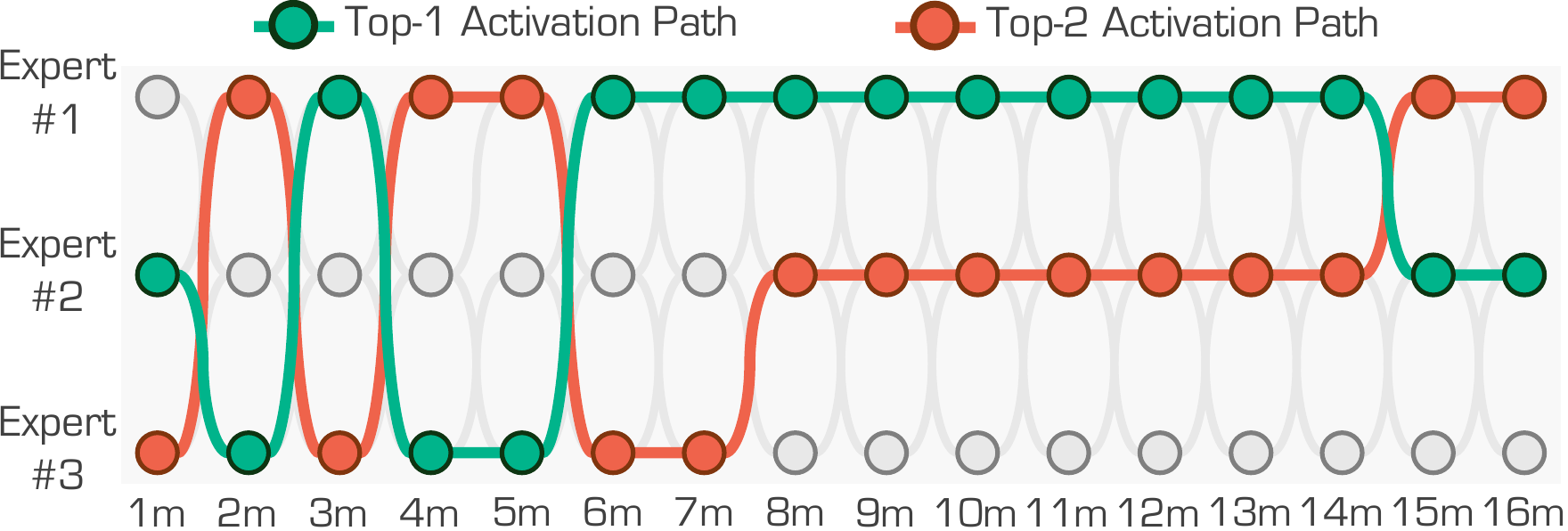}
        \caption{Expert Activation Paths (Distances to Ego)}
        \label{fig_supp:path_distance}
    \end{subfigure}
    \vspace{-0.2cm}
    \caption{Visual interpretations of the expert activation paths in CML. The experts are \textcolor{moe_green}{\texttt{\#1}} range view, \textcolor{moe_red}{\texttt{\#2}} voxel, and \textcolor{moe_blue}{\texttt{\#3}} point, respectively.}
    \label{fig_supp:path_activation}
\end{figure*}

\begin{figure*}[t]
    \centering
    \begin{subfigure}[h]{0.48\textwidth}
        \centering
        \includegraphics[width=\linewidth]{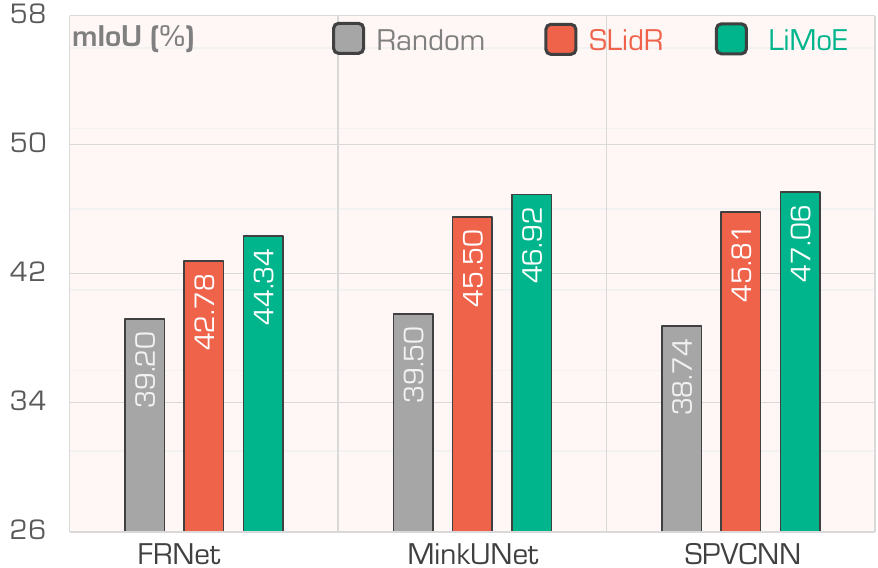}
        \caption{SemanticKITTI}
        \label{fig:backbone_kitti}
    \end{subfigure}~~~~~
    \begin{subfigure}[h]{0.48\textwidth}
        \centering
        \includegraphics[width=\linewidth]{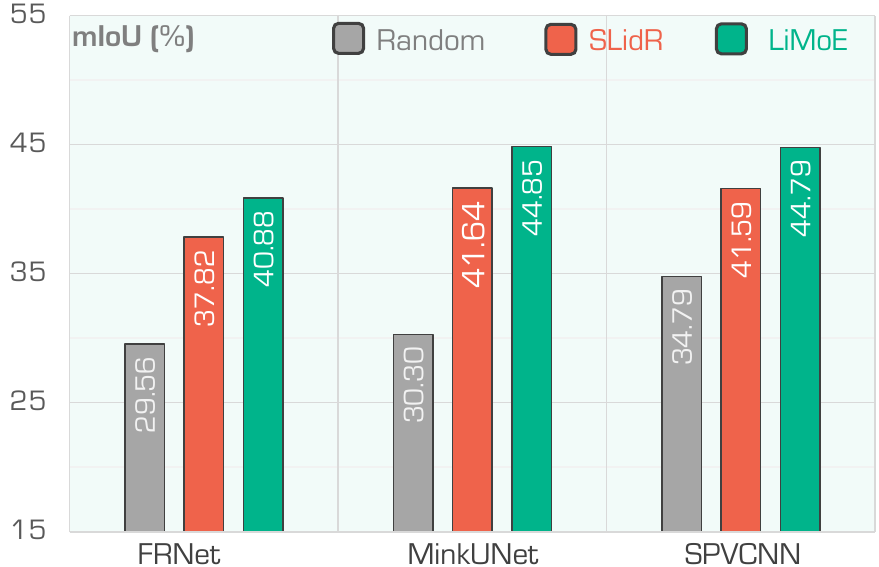}
        \caption{nuScenes}
        \label{fig:backbone_nuscenes}
    \end{subfigure}
    \vspace{-0.2cm}
    \caption{\textbf{Ablation study on different backbones} for downstream tasks. The backbones are initialized with random weights, SLidR~\cite{sautier2022slidr}, and LiMoE, respectively, and fine-tuned on the SemanticKITTI \cite{behley2019semanticKITTI} and nuScenes \cite{fong2022panoptic-nuScenes} datasets using $1\%$ annotations.}
    \label{fig_supp:ablation_backbone}
\end{figure*}

\section{Additional Quantitative Results}

In this section, we present class-wise LiDAR semantic segmentation results to reinforce the findings and conclusions presented in the main body of the paper.

\subsection{Class-Wise Linear Probing Results}

\cref{tab:linear_probing} showcases class-wise LiDAR semantic segmentation results on the \textit{nuScenes} \cite{caesar2020nuScenes,fong2022panoptic-nuScenes} dataset, achieved through pretraining followed by linear probing. The evaluation covers all 16 semantic classes, offering a detailed performance comparison across diverse object categories. \textsf{\textcolor{moe_red}{Li}\textcolor{moe_green}{MoE}} consistently surpasses single-representation baselines for every class, including challenging categories like ``pedestrian'', ``bicycle'', and ``traffic cone''. These results emphasize the advantage of our approach in utilizing complementary features from range images, sparse voxels, and raw points during the CML stage to capture high-level semantic correlations effectively.

\subsection{Class-Wise Fine-Tuning Results}

\cref{tab:1pct} presents class-wise LiDAR semantic segmentation results on the \textit{nuScenes} \cite{caesar2020nuScenes,fong2022panoptic-nuScenes} dataset, derived from pretraining followed by fine-tuning with only $1\%$ of the available annotations. The results highlight that \textsf{\textcolor{moe_red}{Li}\textcolor{moe_green}{MoE}} consistently outperforms single-representation baselines across all classes, with particularly notable gains for dynamic objects such as ``pedestrian'', ``bicycle'', and ``motorcycle'', which often exhibit complex structures. These improvements stem from the SMS stage, where the integration of multiple representations enables the model to capture complementary object attributes, enhancing segmentation performance.

\begin{table}[t]
    \centering
    \caption{\textbf{Detection comparison} of state-of-the-art pretraining methods pretrained and fine-tuned on the \textit{nuScenes} dataset \cite{caesar2020nuScenes}, using specified data proportions. All methods utilize CenterPoint \cite{yin2021centerpoint} as 3D object detection backbones. All scores are given in percentage (\%). The best scores are highlighted in \textbf{bold}.}
    \vspace{-0.2cm}
    \label{tab_supp:detection}
    \resizebox{\linewidth}{!}{
    \begin{tabular}{r|r|p{0.6cm}<{\centering}p{0.6cm}<{\centering}|p{0.6cm}<{\centering}p{0.6cm}<{\centering}|p{0.6cm}<{\centering}p{0.6cm}<{\centering}}
        \toprule
        \multirow{3.7}{*}{\textbf{Method}} & \multirow{3.7}{*}{\textbf{Venue}} & \multicolumn{6}{c}{\textbf{nuScenes}}
        \\\cmidrule{3-8}
        & & \multicolumn{2}{c|}{\textbf{5\%}} & \multicolumn{2}{c|}{\textbf{10\%}} & \multicolumn{2}{c}{\textbf{20\%}}
        \\
        & & \textbf{mAP} & \textbf{NDS} & \textbf{mAP} & \textbf{NDS} & \textbf{mAP} & \textbf{NDS}
        \\\midrule\midrule
        \cellcolor{moe_gray!18}\textcolor{gray}{Random} & - & \cellcolor{moe_gray!18}\textcolor{gray}{$38.0$} & \cellcolor{moe_gray!18}\textcolor{gray}{$44.3$} & \cellcolor{moe_gray!18}\textcolor{gray}{$46.9$} & \cellcolor{moe_gray!18}\textcolor{gray}{$55.5$} & \cellcolor{moe_gray!18}\textcolor{gray}{$50.2$} & \cellcolor{moe_gray!18}\textcolor{gray}{$59.7$}
        \\
        PointContrast \cite{xie2020pointcontrast} & ECCV'20 & $39.8$ & $45.1$ & $47.7$ & $56.0$ & - & -
        \\
        GCC-3D \cite{liang2021exploring} & ICCV'21 & $41.1$ & $46.8$ & $48.4$ & $56.7$ & - & -
        \\
        SLidR \cite{sautier2022slidr} & CVPR'22 & $43.3$ & $52.4$ & $47.5$ & $56.8$ & $50.4$ & $59.9$
        \\
        TriCC \cite{pang2023tricc} & CVPR'23 & $44.6$ & $54.4$ & $48.9$ & $58.1$ & $50.9$ & $60.3$
        \\
        CSC \cite{chen2024csc} & CVPR'24 & $45.3$ & $54.2$ & $49.3$ & $58.3$ & $51.9$ & $61.3$
        \\
        SuperFlow \cite{xu2024superflow} & ECCV'24 & $46.0$ & $54.9$ & $49.7$ & $58.5$ & $52.5$ & $61.5$
        \\
        + \textsf{\textcolor{moe_red}{Li}\textcolor{moe_green}{MoE}} & \textbf{Ours} & \cellcolor{moe_green!12}$\mathbf{47.3}$ & \cellcolor{moe_green!12}$\mathbf{55.3}$ & \cellcolor{moe_green!12}$\mathbf{50.6}$ & \cellcolor{moe_green!12}$\mathbf{59.0}$ & \cellcolor{moe_green!12}$\mathbf{53.2}$ & \cellcolor{moe_green!12}$\mathbf{61.8}$
        \\\bottomrule
    \end{tabular}}
\end{table}

\subsection{3D Object Detection}

To further evaluate the effectiveness of \textsf{\textcolor{moe_red}{Li}\textcolor{moe_green}{MoE}}, we extend our framework to the 3D object detection task. Specifically, we integrate three heterogeneous representation experts and distill their knowledge into VoxelNet \cite{yan2018second} during the CML stage. For downstream fine-tuning, we follow the detection pipeline of CenterPoint \cite{yin2021centerpoint}. As shown in \cref{tab_supp:detection}, our method achieves substantial improvements over single-representation learning, further demonstrating the effectiveness of MoE in unifying multiple representations into a compact and expressive feature space.

\begin{table}
    \centering
    \caption{Ablation study on \textbf{incorporating representation diversity}. All scores are given in percentage (\%).}
    \vspace{-0.2cm}
    \label{tab_supp:representation}
    \resizebox{\linewidth}{!}{
    \begin{tabular}{r|ccc|c|c}
        \toprule
        \multirow{2}{*}{\textbf{Method}} & \multicolumn{3}{c|}{\textbf{nuScenes}} & \textbf{KITTI} & \textbf{Waymo}
        \\
        & \textbf{LP} & \textbf{1\%} & \textbf{5\%} & \textbf{1\%} & \textbf{1\%}
        \\\midrule\midrule
        \cellcolor{moe_gray!18}\textcolor{gray}{Random} & \cellcolor{moe_gray!18}\textcolor{gray}{$8.10$} & \cellcolor{moe_gray!18}\textcolor{gray}{$30.30$} & \cellcolor{moe_gray!18}\textcolor{gray}{$47.84$} & \cellcolor{moe_gray!18}\textcolor{gray}{$39.50$} & \cellcolor{moe_gray!18}\textcolor{gray}{$39.41$}
        \\
        SLidR \cite{sautier2022slidr} & $45.35$ & $41.64$ & $55.83$ & $45.50$ & $48.32$
        \\
        3 $\times$ MinkUNet & $45.51$ & $42.72$ & $56.73$ & $46.35$ & $48.94$
        \\
        \textsf{\textcolor{moe_red}{Li}\textcolor{moe_green}{MoE}} & \cellcolor{moe_green!12}$\mathbf{46.56}$ & \cellcolor{moe_green!12}$\mathbf{46.89}$ & \cellcolor{moe_green!12}$\mathbf{58.09}$ & \cellcolor{moe_green!12}$\mathbf{47.96}$ & \cellcolor{moe_green!12}$\mathbf{49.50}$
        \\\bottomrule
    \end{tabular}}
\end{table}

\subsection{Representation Diversity}

To investigate the role of representation diversity, we conduct an ablation study by replacing the three heterogeneous experts in our framework with three identical sparse voxel-based models, all implemented using MinkUNet \cite{choy2019minkowski}. The results, summarized in \cref{tab_supp:representation}, show that using multiple identical experts provides only a marginal improvement over single-representation learning. However, this setting remains significantly inferior to \textsf{\textcolor{moe_red}{Li}\textcolor{moe_green}{MoE}}, which integrates diverse representations. This performance gap underscores the importance of representation diversity: by leveraging complementary features from range images, sparse voxels, and raw points, the MoE layer effectively captures richer geometric and semantic information, leading to superior segmentation performance.

\begin{table}
    \centering
    \caption{Ablation study on \textbf{mixing strategies} for integrating multiple representations. All scores are given in percentage (\%).}
    \vspace{-0.2cm}
    \label{tab_supp:mixing}
    \resizebox{\linewidth}{!}{
    \begin{tabular}{r|ccc|c|c}
        \toprule
        \multirow{2}{*}{\textbf{Mixing}} & \multicolumn{3}{c|}{\textbf{nuScenes}} & \textbf{KITTI} & \textbf{Waymo}
        \\
        & \textbf{LP} & \textbf{1\%} & \textbf{5\%} & \textbf{1\%} & \textbf{1\%}
        \\\midrule\midrule
        \cellcolor{moe_gray!18}\textcolor{gray}{Random} & \cellcolor{moe_gray!18}\textcolor{gray}{$8.10$} & \cellcolor{moe_gray!18}\textcolor{gray}{$30.30$} & \cellcolor{moe_gray!18}\textcolor{gray}{$47.84$} & \cellcolor{moe_gray!18}\textcolor{gray}{$39.50$} & \cellcolor{moe_gray!18}\textcolor{gray}{$39.41$}
        \\
        SLidR \cite{sautier2022slidr} & $45.35$ & $41.64$ & $55.83$ & $45.50$ & $48.32$
        \\
        Concatenate & $45.82$ & $44.75$ & $56.43$ & $46.76$ & $48.53$
        \\
        Addition & $45.73$ & $44.82$ & $56.83$ & $46.40$ & $48.71$
        \\
        Average & $46.56$ & $46.89$ & $57.12$ & $47.96$ & $49.04$
        \\
        \textsf{\textcolor{moe_red}{Li}\textcolor{moe_green}{MoE}} & \cellcolor{moe_green!12}$\mathbf{46.56}$ & \cellcolor{moe_green!12}$\mathbf{46.89}$ & \cellcolor{moe_green!12}$\mathbf{58.09}$ & \cellcolor{moe_green!12}$\mathbf{47.96}$ & \cellcolor{moe_green!12}$\mathbf{49.50}$
        \\\bottomrule
    \end{tabular}}
\end{table}

\subsection{Effectiveness of MoE-Based Mixing}

In this work, we employ MoE to selectively aggregate complementary features from multiple representations. However, alternative feature mixing strategies, such as concatenation, addition, and averaging, can also be considered. To validate the effectiveness of our MoE-based approach, we conduct an ablation study comparing different mixing strategies, with results summarized in \cref{tab_supp:mixing}. Our method achieves the best performance, as it effectively acts as an attention-based mechanism, dynamically selecting the most relevant features for each LiDAR point. In contrast, the other three mixing strategies treat all representations equally, lacking the ability to adaptively capture complementary features, which limits their effectiveness.

\subsection{Extend to Different Backbones}

We evaluate the downstream performance of FRNet \cite{xu2025frnet}, MinKUNet \cite{choy2019minkowski}, and SPVCNN \cite{tang2020spvcnn} across various pretraining strategies, including random initialization, SLidR \cite{sautier2022slidr}, and LiMoE. Importantly, the downstream fine-tuning does not employ the MoE strategy to combine multiple LiDAR representations, ensuring a fair comparison among the methods. Fine-tuning is conducted on two widely-used datasets, SemanticKITTI \cite{behley2019semanticKITTI} and nuScenes \cite{fong2022panoptic-nuScenes}, with only $1\%$ of the annotations available. As shown in \cref{fig_supp:ablation_backbone}, LiMoE pretraining consistently improves the performance of single-representation methods. This demonstrates the scalability and generalizable feature representations learned during the LiMoE pretraining stage, making it effective for enhancing downstream tasks.

\section{Additional Qualitative Results}

In this section, we provide additional qualitative examples to visually compare different approaches presented in the main body of the paper.

\subsection{Route Activations from CML}

LiDAR sensors inherently operate with a fixed number of beams, resulting in a structured arrangement of data points within the captured point clouds. This beam-based configuration provides a natural attribution for the LiDAR point clouds, with each beam contributing a distinct set of points that collectively form a comprehensive 3D representation of the environment. Furthermore, the distance of points from the ego vehicle often correlates with the orientation and elevation of the laser beams. Upper beams are typically designed to detect objects at longer distances, capturing information about the far-field surroundings. In contrast, middle beams are optimized for medium-range detections, while lower beams primarily focus on capturing close-proximity objects \cite{kong2023lasermix,kong2025lasermix2}.

To understand how each LiDAR representation contributes to the fused features within the MoE layer, we conduct a detailed statistical analysis of export selection patterns during the CML stage. Specifically, we examined the activation frequency of each representation -- \textcolor{moe_green}{range view}, \textcolor{moe_red}{voxel}, and \textcolor{moe_blue}{point} -- across varying laser beams and distances from the ego vehicle, measuring their respective contributions to the fused outputs. As depicted in \cref{fig_supp:path_activation}, distinct focus areas emerge for each LiDAR representation, aligning with their inherent strengths. The \textcolor{moe_green}{range view} representation shows a higher activation frequency in middle-range regions. The \textcolor{moe_red}{voxel} representation demonstrates a significant focus on upper laser beams and far-field regions. The \textcolor{moe_blue}{point} representation dominates in close-range regions. This analysis underscores the complementary nature of the three representations. The MoE layer dynamically selects the most suitable representation based on the spatial and distance characteristics of the input, enabling a more robust and comprehensive understanding of 3D environments.

\subsection{Point-Wise Activation from SMS}

SMS supervises the feature learning process by integrating the semantic logits from multiple LiDAR representations with guidance from semantic labels. To illustrate which object attributes each representation focuses on within the LiDAR point clouds, we analyze the contribution of each representation to the semantic logits during the SMS stage.

\begin{figure*}
    \centering
    \includegraphics[width=\linewidth]{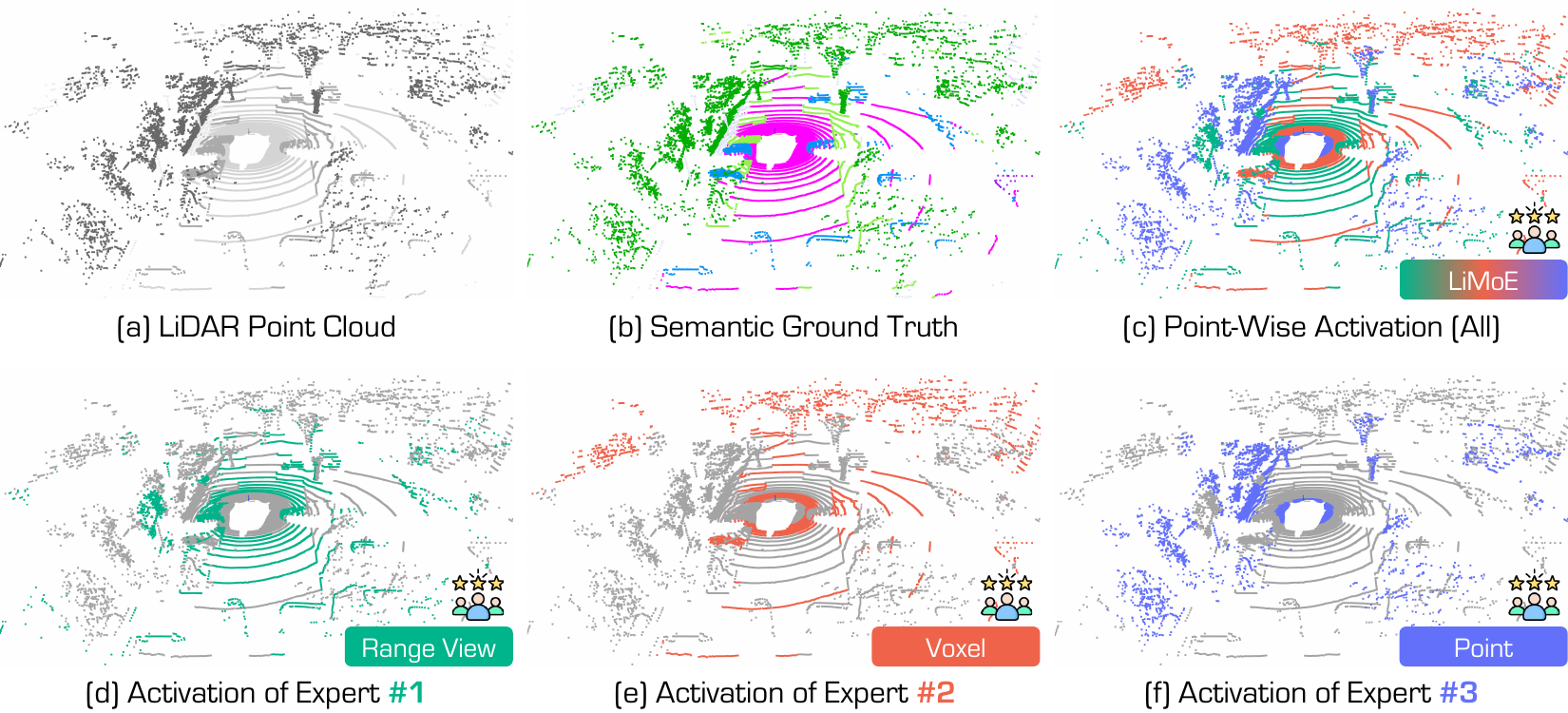}
    \vspace{-0.45cm}
    \caption{Point-wise top-1 activation path in the SMS stage. It highlights the most activated representation for each point during the SMS stage, illustrating how different representations contribute to semantic segmentation based on spatial and object-specific characteristics. Best viewed in colors.}
    \label{fig_supp:point_activation_1}
\end{figure*}

\begin{figure*}
    \centering
    \includegraphics[width=\linewidth]{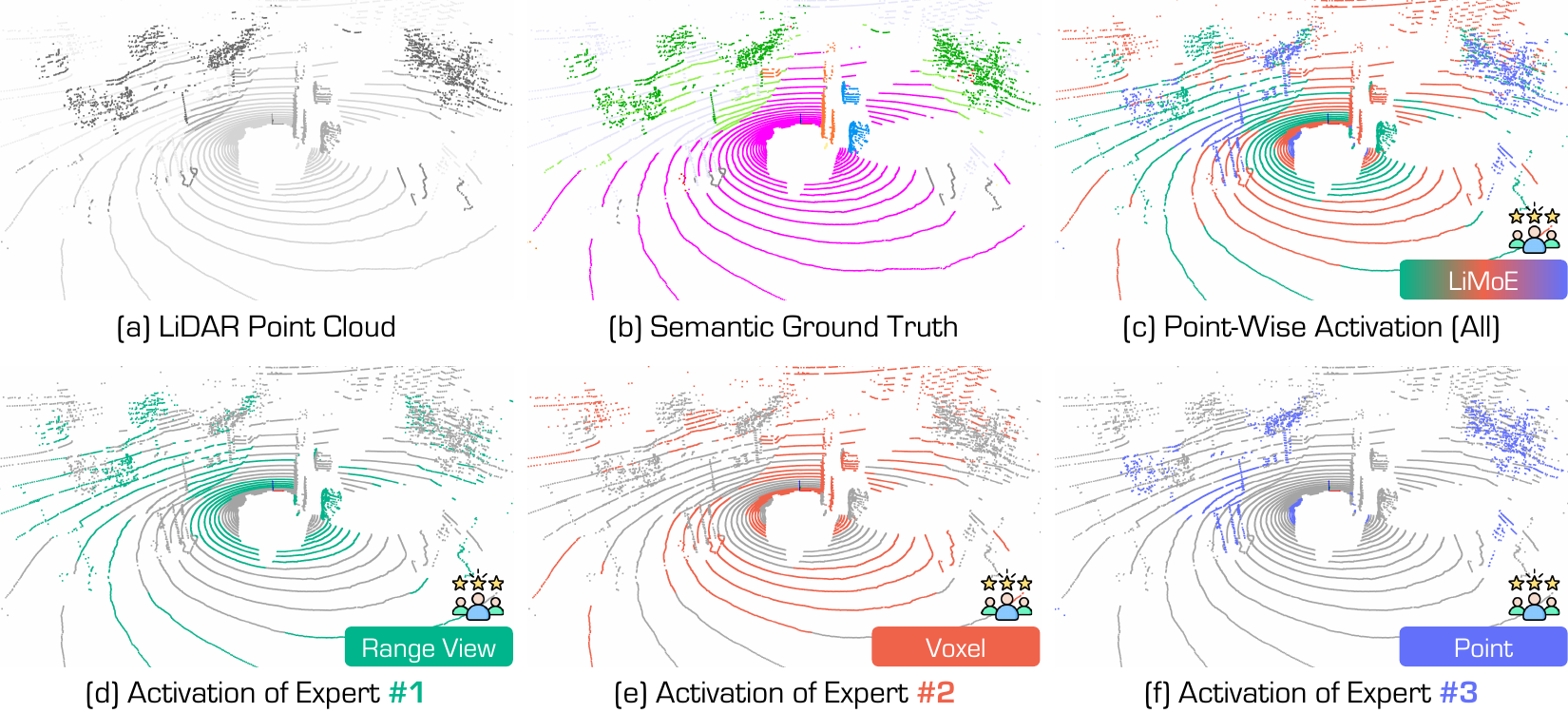}
    \vspace{-0.45cm}
    \caption{Point-wise top-1 activation path in the SMS stage. It highlights the most activated representation for each point during the SMS stage, illustrating how different representations contribute to semantic segmentation based on spatial and object-specific characteristics. Best viewed in colors.}
    \label{fig_supp:point_activation_2}
\end{figure*}

\begin{figure*}
    \centering
    \includegraphics[width=\linewidth]{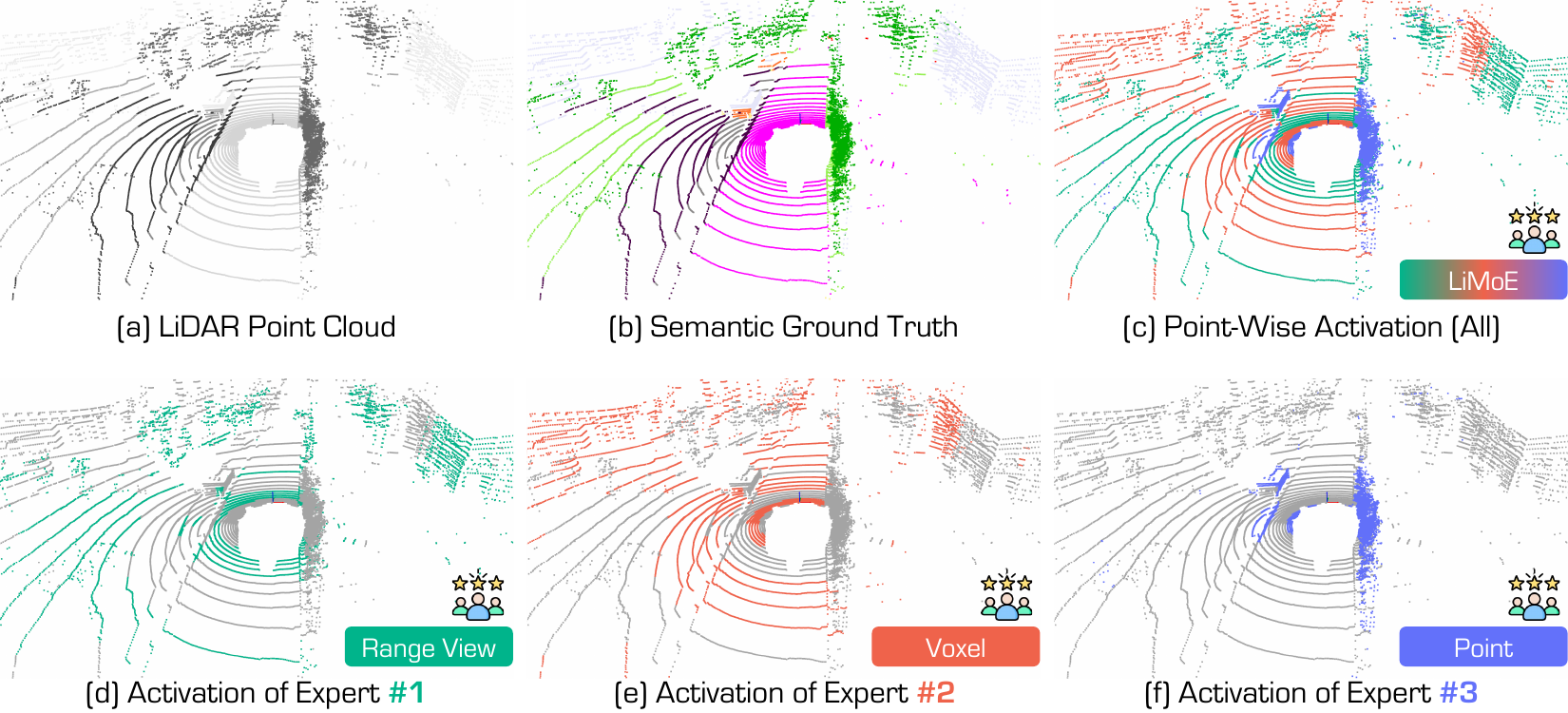}
    \vspace{-0.45cm}
    \caption{Point-wise top-1 activation path in the SMS stage. It highlights the most activated representation for each point during the SMS stage, illustrating how different representations contribute to semantic segmentation based on spatial and object-specific characteristics. Best viewed in colors.}
    \label{fig_supp:point_activation_3}
\end{figure*}

\begin{figure*}
    \centering
    \includegraphics[width=\linewidth]{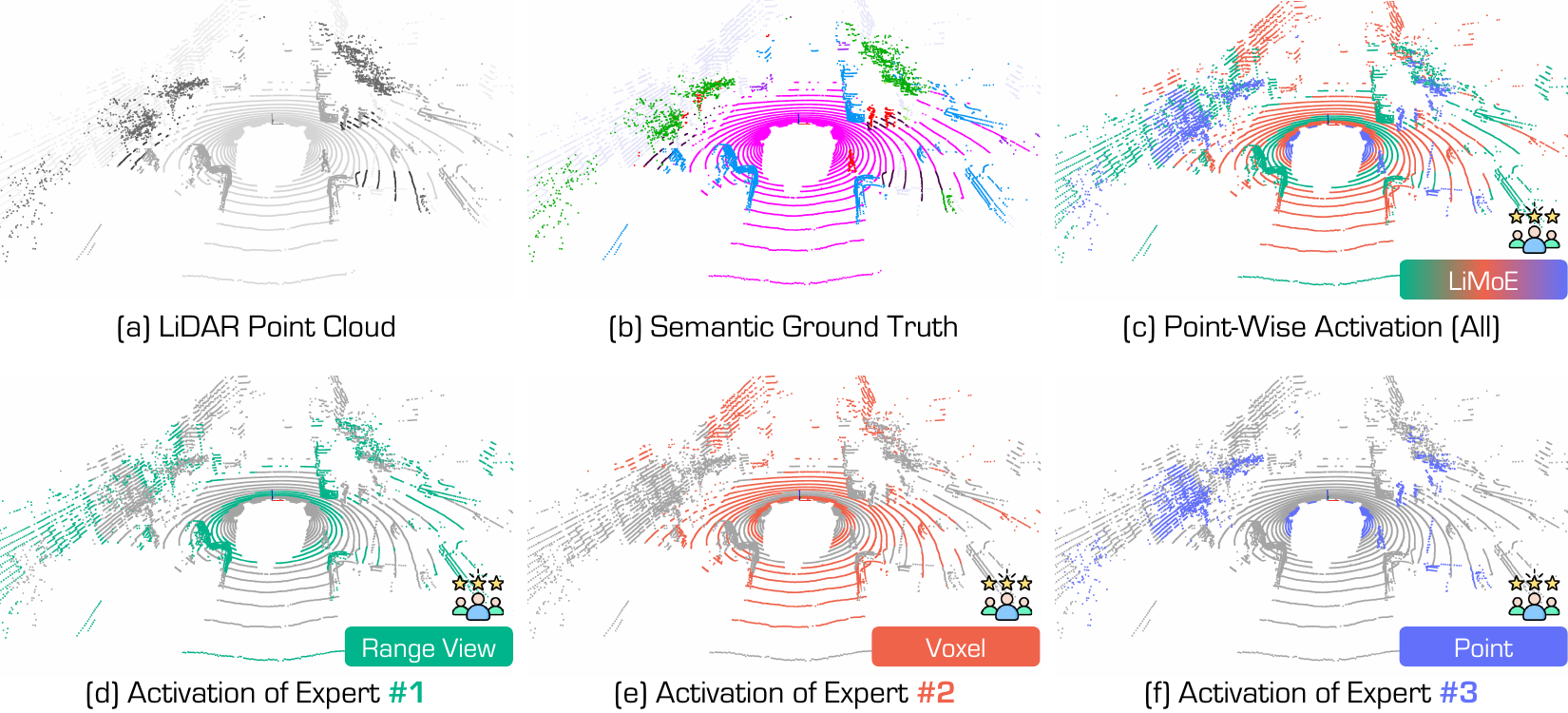}
    \vspace{-0.45cm}
    \caption{Point-wise top-1 activation path in the SMS stage. It highlights the most activated representation for each point during the SMS stage, illustrating how different representations contribute to semantic segmentation based on spatial and object-specific characteristics. Best viewed in colors.}
    \label{fig_supp:point_activation_4}
\end{figure*}

Specifically, the MoE layer computes a gating score that indicates the relative importance of each representation for individual points. We highlight the most relevant attributes contributed by each representation and project them onto the corresponding point clouds. As shown in \cref{fig_supp:point_activation_1}, \cref{fig_supp:point_activation_2}, \cref{fig_supp:point_activation_3}, and \cref{fig_supp:point_activation_4}, the \textcolor{moe_green}{range view} representation predominantly emphasizes dynamic objects, such as ``car'', ``truck'', as well as objects in medium-range regions. The \textcolor{moe_red}{voxel} representation excels in capturing static background elements, such as ``road'' and far-field objects within sparse regions. The \textcolor{moe_blue}{point} representation specializes in intricate details, such as object edges and close-range features, which are crucial for accurate boundary delineation.

This visualization demonstrates the complementary nature of these representations and underscores the effectiveness of SMS in dynamically leveraging their unique strengths. By aligning these diverse features, SMS ensures comprehensive feature learning, leading to improved segmentation performance across varied object types and environmental conditions.

\subsection{LiDAR Segmentation Results}

\begin{figure*}
    \centering
    \includegraphics[width=\linewidth]{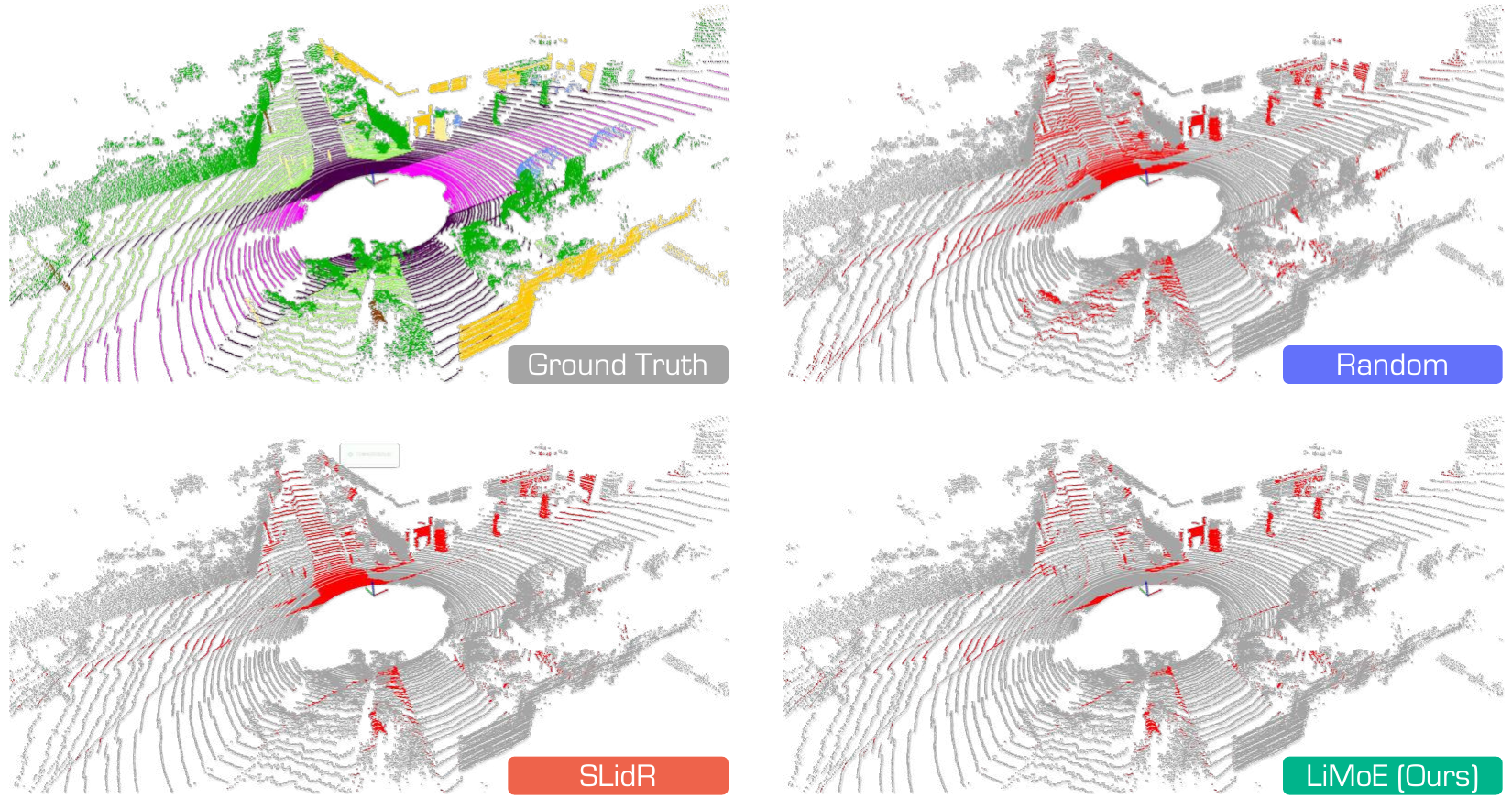}
    \vspace{-0.6cm}
    \caption{\textbf{Qualitative assessments} of state-of-the-art pretraining methods, pretrained on \textit{nuScenes} \cite{caesar2020nuScenes} and fine-tuned on \textit{SemanticKITTI} \cite{behley2019semanticKITTI} with $1\%$ annotations. The error maps depict \textcolor{moe_gray}{correct} and \textcolor{red}{incorrect} predictions in \textcolor{moe_gray}{gray} and \textcolor{red}{red}, respectively. Best viewed in colors.}
    \label{fig_supp:semkitti_1}
\end{figure*}

\begin{figure*}
    \centering
    \includegraphics[width=\linewidth]{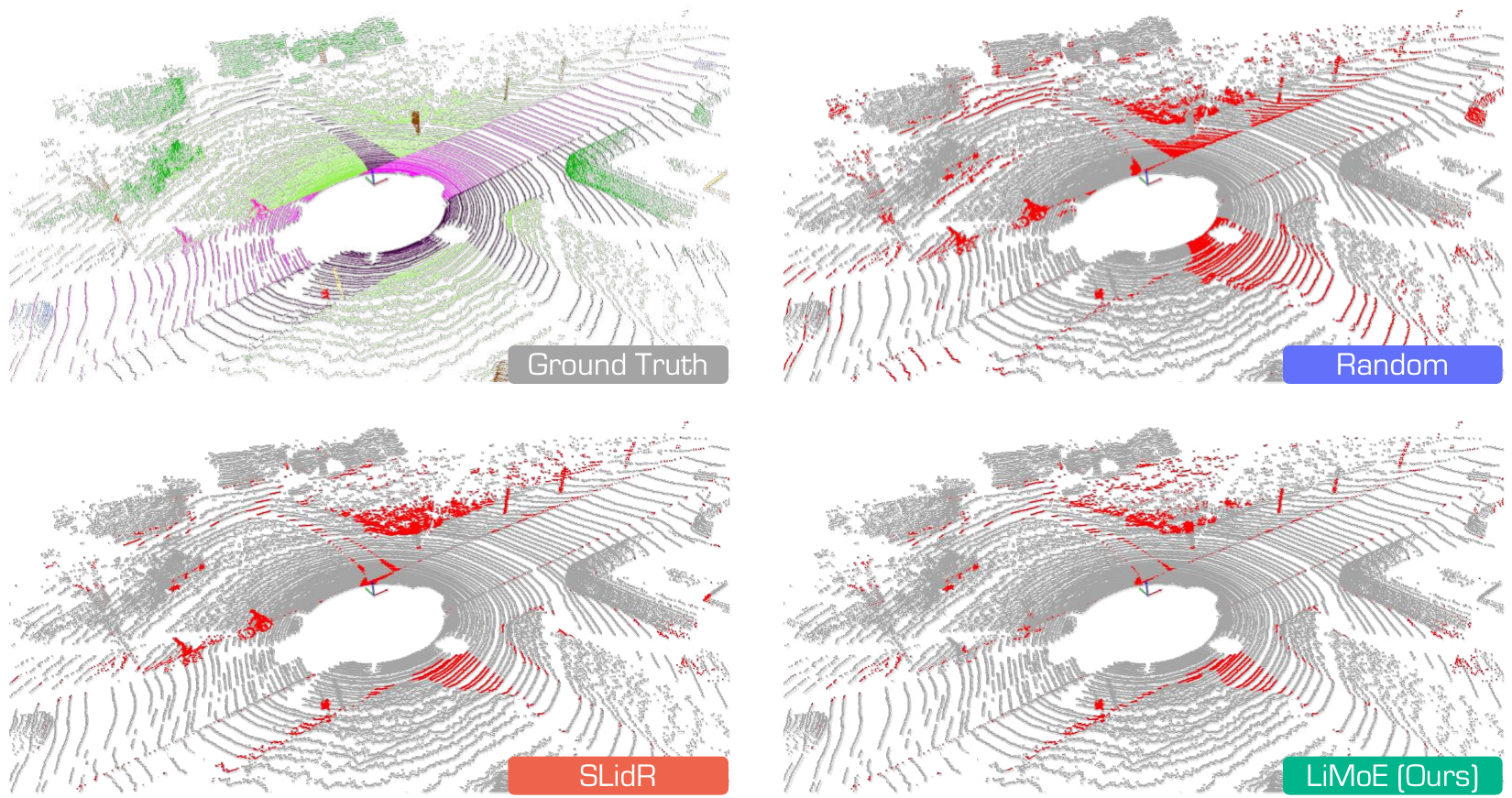}
    \vspace{-0.6cm}
    \caption{\textbf{Qualitative assessments} of state-of-the-art pretraining methods, pretrained on \textit{nuScenes} \cite{caesar2020nuScenes} and fine-tuned on \textit{SemanticKITTI} \cite{behley2019semanticKITTI} with $1\%$ annotations. The error maps depict \textcolor{moe_gray}{correct} and \textcolor{red}{incorrect} predictions in \textcolor{moe_gray}{gray} and \textcolor{red}{red}, respectively. Best viewed in colors.}
    \label{fig_supp:semkitti_2}
\end{figure*}

\begin{figure*}
    \centering
    \includegraphics[width=\linewidth]{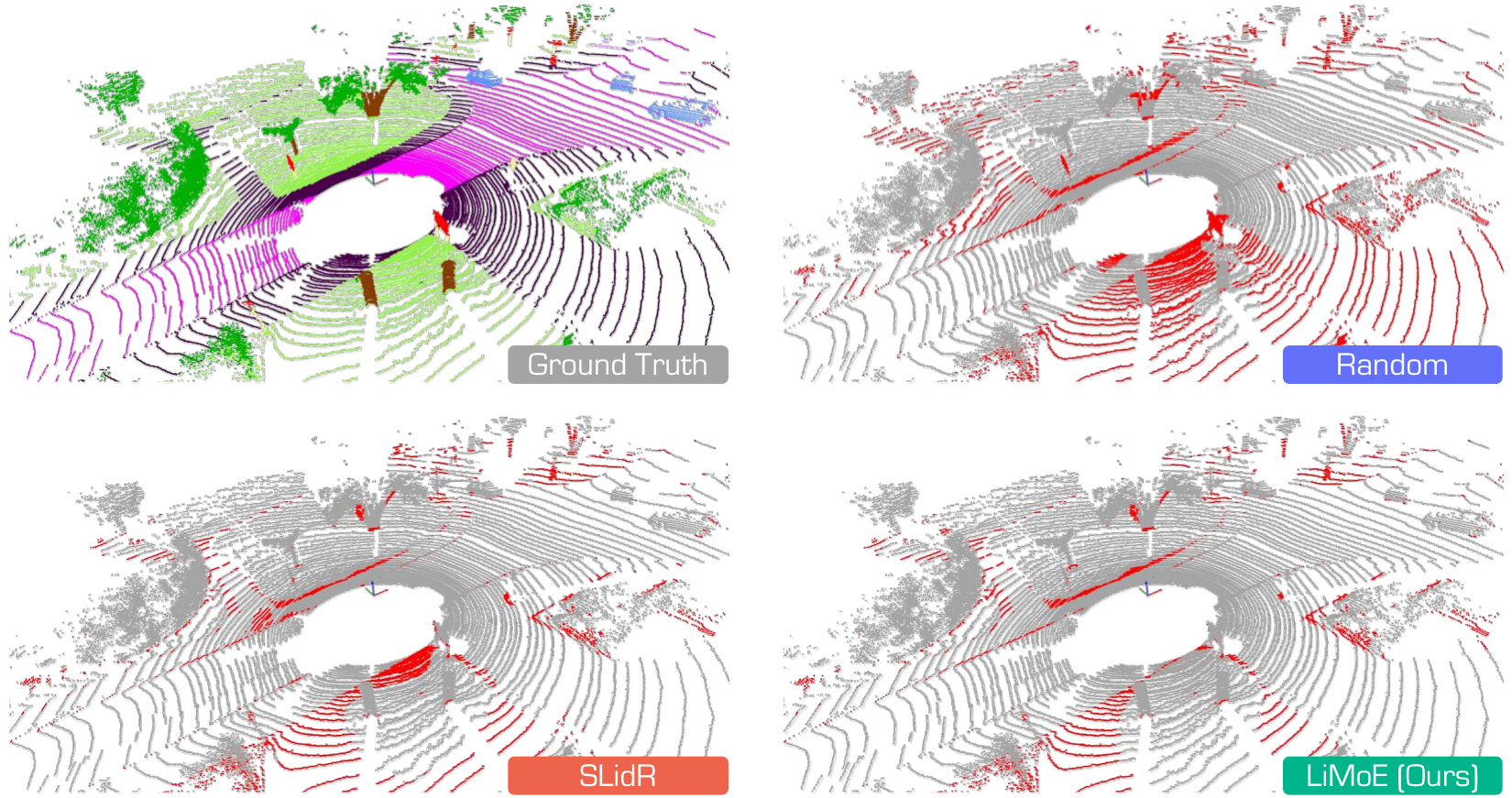}
    \vspace{-0.6cm}
    \caption{\textbf{Qualitative assessments} of state-of-the-art pretraining methods, pretrained on \textit{nuScenes} \cite{caesar2020nuScenes} and fine-tuned on \textit{SemanticKITTI} \cite{behley2019semanticKITTI} with $1\%$ annotations. The error maps depict \textcolor{moe_gray}{correct} and \textcolor{red}{incorrect} predictions in \textcolor{moe_gray}{gray} and \textcolor{red}{red}, respectively. Best viewed in colors.}
    \label{fig_supp:semkitti_3}
\end{figure*}

\begin{figure*}
    \centering
    \includegraphics[width=\linewidth]{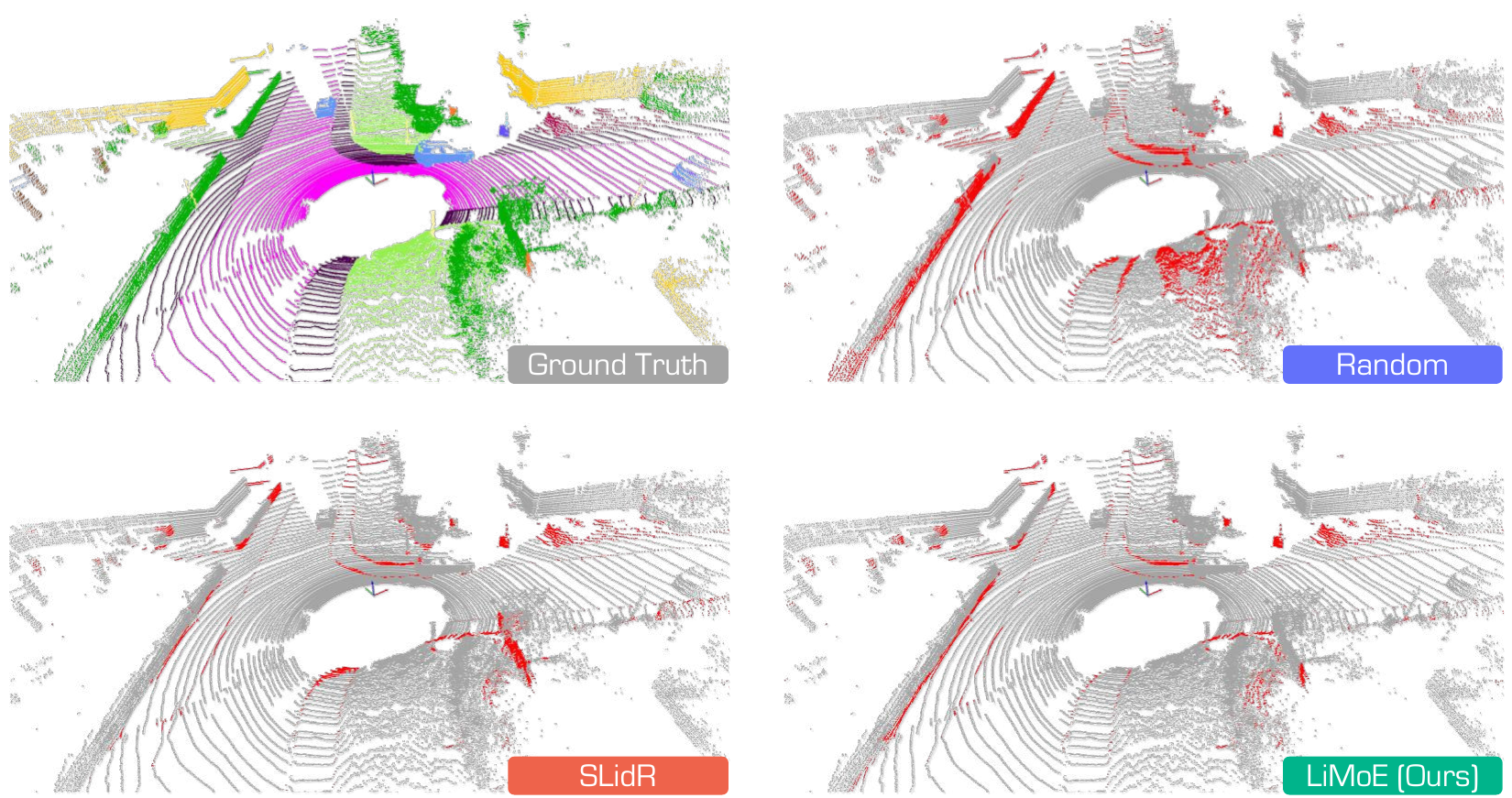}
    \vspace{-0.6cm}
    \caption{\textbf{Qualitative assessments} of state-of-the-art pretraining methods, pretrained on \textit{nuScenes} \cite{caesar2020nuScenes} and fine-tuned on \textit{SemanticKITTI} \cite{behley2019semanticKITTI} with $1\%$ annotations. The error maps depict \textcolor{moe_gray}{correct} and \textcolor{red}{incorrect} predictions in \textcolor{moe_gray}{gray} and \textcolor{red}{red}, respectively. Best viewed in colors.}
    \label{fig_supp:semkitti_4}
\end{figure*}

In \cref{fig_supp:semkitti_1}, \cref{fig_supp:semkitti_2}, \cref{fig_supp:semkitti_3}, and \cref{fig_supp:semkitti_4}, we present qualitative LiDAR segmentation results, highlighting the performance of models pretrained on the \textit{nuScenes} \cite{zhou2021panoptic} dataset using various methods and fine-tuned on the \textit{SemanticKITTI} dataset with $1\%$ of the available annotations. As depicted, \textsf{\textcolor{moe_red}{Li}\textcolor{moe_green}{MoE}} consistently outperforms single-representation approaches by capturing intricate scene details and achieving a significant reduction in segmentation errors across challenging semantic classes. Notably, it excels in handling dynamic objects such as ``pedestrian'', where other methods often struggle. These results highlight the ability of our multi-representation fusion framework to integrate complementary features, leading to more robust and accurate segmentation.

\subsection{Cosine Similarity Results}

\begin{figure*}
    \centering
    \includegraphics[width=\linewidth]{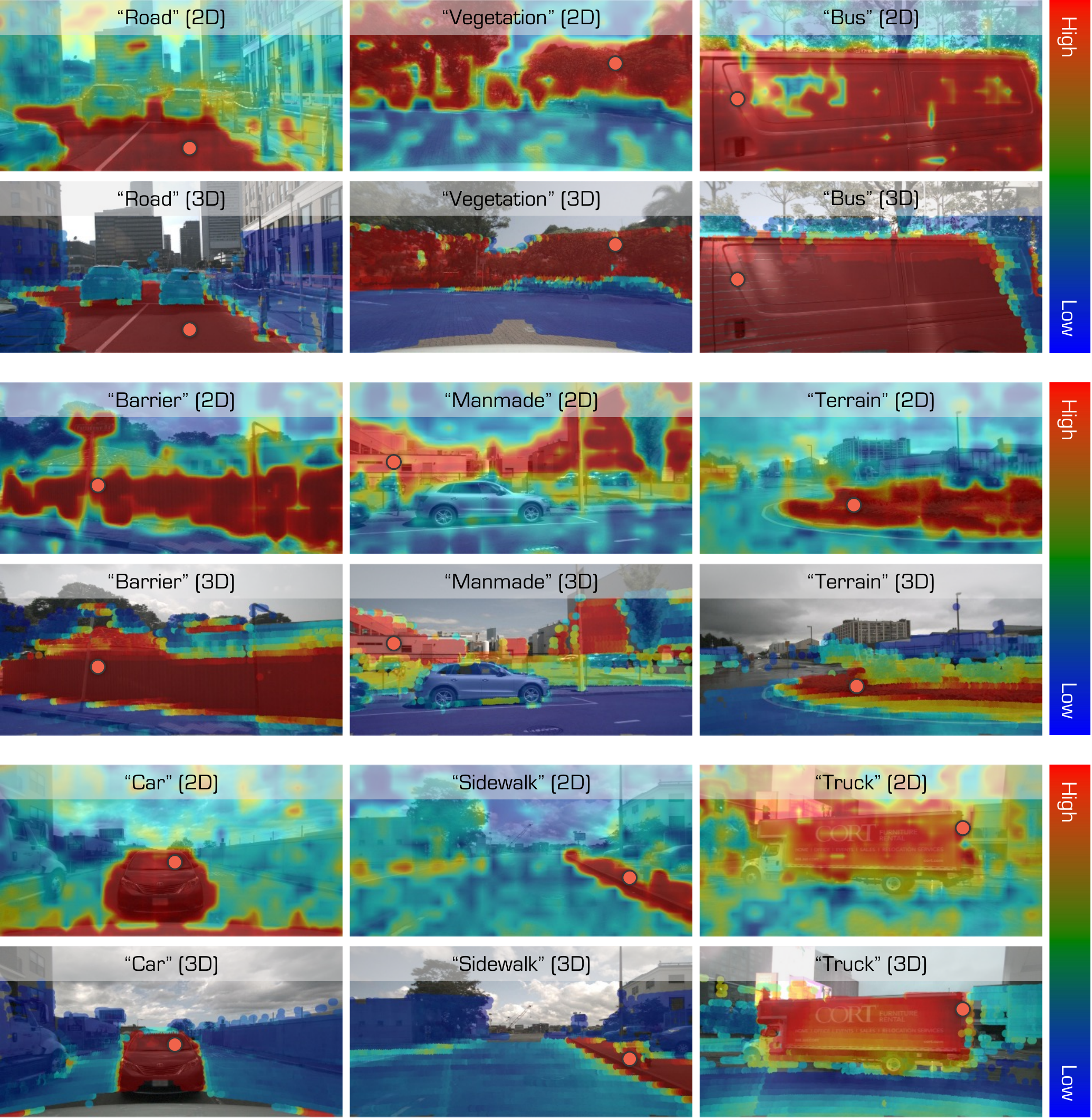}
    \vspace{-0.6cm}
    \caption{\textbf{Cosine similarity} between the learned features of a query point (denoted as the \textcolor{moe_red}{red} dot) and: (1) the features of the image of the same scene (the 1st, 3rd, and 5th rows); and (2) the features of the LiDAR points of the same scene that are projected onto the image (the 2nd, 4th, and 6th rows). Best viewed in colors.}
    \label{fig_supp:cosine}
\end{figure*}

In \cref{fig_supp:cosine}, we provide additional cosine similarity maps generated during the CML stage. These maps demonstrate the ability of \textsf{\textcolor{moe_red}{Li}\textcolor{moe_green}{MoE}} to align features from different LiDAR representations, showcasing high semantic correlations across diverse regions of the scene during pretraining. This alignment reflects the effectiveness of our framework in fusing information from range images, sparse voxels, and raw points to capture complementary semantic cues. By fostering strong inter-representation consistency, our method establishes a solid foundation for downstream tasks, improving the performance and reliability of LiDAR-based segmentation systems in real-world scenarios.

\section{Public Resources Used}

In this section, we acknowledge the use of public resources, during the course of this work.

\subsection{Public Codebase Used}

We acknowledge the use of the following public codebase, during the course of this work.

\begin{itemize}
    \item MMEngine\footnote{\url{https://github.com/open-mmlab/mmengine}.} \dotfill Apache License 2.0
    \item MMCV\footnote{\url{https://github.com/open-mmlab/mmcv}.} \dotfill Apache License 2.0
    \item MMPretrain\footnote{\url{https://github.com/open-mmlab/mmpretrain}.} \dotfill Apache License 2.0
    \item MMDetection\footnote{\url{https://github.com/open-mmlab/mmdetection}.} \dotfill Apache License 2.0
    \item MMDetection3d\footnote{\url{https://github.com/open-mmlab/mmdetection3d}.} \dotfill Apache License 2.0
\end{itemize}

\subsection{Public Datasets Used}

We acknowledge the use of the following public datasets, during the course of this work.

\begin{itemize}
    \item nuScenes\footnote{\url{https://www.nuscenes.org/nuscenes}.} \dotfill CC BY-NC-SA 4.0
    \item SemanticKITTI\footnote{\url{http://semantic-kitti.org}.} \dotfill CC BY-NC-SA 4.0
    \item WaymoOpenDataset\footnote{\url{https://waymo.com/open}.} \dotfill Waymo Dataset License
    \item ScribbleKITTI\footnote{\url{https://github.com/ouenal/scribblekitti}.} \dotfill Unknown
    \item RELLIS-3D\footnote{\url{https://github.com/unmannedlab/RELLIS-3D}.} \dotfill CC BY-NC-SA 3.0
    \item SemanticPOSS\footnote{\url{http://www.poss.pku.edu.cn/semanticposs.html}.} \dotfill CC BY-NC-SA 3.0
    \item SemanticSTF\footnote{\url{https://github.com/xiaoaoran/SemanticSTF}.} \dotfill CC BY-NC-SA 4.0
    \item SynLiDAR\footnote{\url{https://github.com/xiaoaoran/SynLiDAR}.} \dotfill MiT License
    \item DAPS-3D\footnote{\url{https://github.com/subake/DAPS3D}.} \dotfill MiT License
    \item Synth4D\footnote{\url{https://github.com/saltoricristiano/gipso-sfouda}.} \dotfill GPL-3.0 License
    \item Robo3D\footnote{\url{https://github.com/ldkong1205/Robo3D}.} \dotfill CC BY-NC-SA 4.0
\end{itemize}

\subsection{Public Implementations Used}

We acknowledge the use of the following public implementations, during the course of this work.

\begin{itemize}
    \item nuscenes-devkit\footnote{\url{https://github.com/nutonomy/nuscenes-devkit}.} \dotfill Apache License 2.0
    \item semantic-kitti-api\footnote{\url{https://github.com/PRBonn/semantic-kitti-api}.} \dotfill MIT License
    \item waymo-open-dataset\footnote{\url{https://github.com/waymo-research/waymo-open-dataset}.} \dotfill Apache License 2.0
    \item SLidR\footnote{\url{https://github.com/valeoai/SLidR}.} \dotfill Apache License 2.0
    \item SuperFlow\footnote{\url{https://github.com/Xiangxu-0103/SuperFlow}.} \dotfill Apache License 2.0
    \item FRNet\footnote{\url{https://github.com/Xiangxu-0103/FRNet}.} \dotfill Apache License 2.0
    \item DINOv2\footnote{\url{https://github.com/facebookresearch/dinov2}.} \dotfill Apache License 2.0
    \item torchsparse\footnote{\url{https://github.com/mit-han-lab/torchsparse}.} \dotfill MIT License
    \item Conv-LoRA\footnote{\url{https://github.com/autogluon/autogluon}.} \dotfill Apache License 2.0
    \item MoE-LLaVA\footnote{\url{https://github.com/PKU-YuanGroup/MoE-LLaVA}.} \dotfill Apache License 2.0
\end{itemize}

\begin{table*}[t]
\centering
\caption{The \textbf{per-class IoU scores} of state-of-the-art pretraining methods pretrained and linear-probed on the \textit{nuScenes} \cite{caesar2020nuScenes,fong2022panoptic-nuScenes} dataset. All scores are given in percentage (\%). The \textbf{best} and \underline{2nd best} scores under each group are highlighted in \textbf{bold} and \underline{underline}.}
\vspace{-0.2cm}
\label{tab:linear_probing}
\resizebox{\textwidth}{!}{
\begin{tabular}{r|c|cccccccccccccccc}
    \toprule
    \textbf{Method} & \rotatebox{90}{\textbf{mIoU}} & \rotatebox{90}{\textcolor{nu_barrier}{$\blacksquare$}~barrier} & \rotatebox{90}{\textcolor{nu_bicycle}{$\blacksquare$}~bicycle} & \rotatebox{90}{\textcolor{nu_bus}{$\blacksquare$}~bus} & \rotatebox{90}{\textcolor{nu_car}{$\blacksquare$}~car} & \rotatebox{90}{\textcolor{nu_cons}{$\blacksquare$}~construction vehicle~} & \rotatebox{90}{\textcolor{nu_motor}{$\blacksquare$}~motorcycle} & \rotatebox{90}{\textcolor{nu_ped}{$\blacksquare$}~pedestrian} & \rotatebox{90}{\textcolor{nu_cone}{$\blacksquare$}~traffic cone} & \rotatebox{90}{\textcolor{nu_trailer}{$\blacksquare$}~trailer} & \rotatebox{90}{\textcolor{nu_truck}{$\blacksquare$}~truck} & \rotatebox{90}{\textcolor{nu_driv}{$\blacksquare$}~driveable surface} & \rotatebox{90}{\textcolor{nu_flat}{$\blacksquare$}~other flat} & \rotatebox{90}{\textcolor{nu_sidewalk}{$\blacksquare$}~sidewalk} & \rotatebox{90}{\textcolor{nu_terrain}{$\blacksquare$}~terrain} & \rotatebox{90}{\textcolor{nu_manmade}{$\blacksquare$}~manmade} & \rotatebox{90}{\textcolor{nu_veg}{$\blacksquare$}~vegetation}
    \\\midrule
    \cellcolor{moe_gray!18}\textcolor{gray}{Random} & \cellcolor{moe_gray!18}\textcolor{gray}{$8.1$} & \cellcolor{moe_gray!18}\textcolor{gray}{$0.5$} &
    \cellcolor{moe_gray!18}\textcolor{gray}{$0.0$} & \cellcolor{moe_gray!18}\textcolor{gray}{$0.0$} & \cellcolor{moe_gray!18}\textcolor{gray}{$3.9$} & \cellcolor{moe_gray!18}\textcolor{gray}{$0.0$} & \cellcolor{moe_gray!18}\textcolor{gray}{$0.0$} & \cellcolor{moe_gray!18}\textcolor{gray}{$0.0$} & \cellcolor{moe_gray!18}\textcolor{gray}{$6.4$} & \cellcolor{moe_gray!18}\textcolor{gray}{$0.0$} & \cellcolor{moe_gray!18}\textcolor{gray}{$3.9$} & \cellcolor{moe_gray!18}\textcolor{gray}{$59.6$} & \cellcolor{moe_gray!18}\textcolor{gray}{$0.0$} & \cellcolor{moe_gray!18}\textcolor{gray}{$0.1$} & \cellcolor{moe_gray!18}\textcolor{gray}{$16.2$} & \cellcolor{moe_gray!18}\textcolor{gray}{$30.6$} & \cellcolor{moe_gray!18}\textcolor{gray}{$12.0$}
    \\\midrule
    \multicolumn{18}{l}{\textbf{Distill: None}}
    \\
    PointContrast \cite{xie2020pointcontrast} & \underline{$21.9$} & - & - & - & - & - & - & - & - & - & - & - & - & - & - & - & -
    \\
    DepthContrast \cite{zhang2021depthcontrast} & $\mathbf{22.1}$ & - & - & - & - & - & - & - & - & - & - & - & - & - & - & - & -
    \\
    ALSO \cite{boulch2023also} & - & - & - & - & - & - & - & - & - & - & - & - & - & - & - & - & -
    \\
    BEVContrast \cite{sautier2024bevcontrast} & - & - & - & - & - & - & - & - & - & - & - & - & - & - & - & - & -
    \\\midrule
    \multicolumn{18}{l}{\textbf{Distill: ResNet-50}}
    \\
    PPKT \cite{liu2021ppkt} & $35.9$ & - & - & - & - & - & - & - & - & - & - & - & - & - & - & - & -
    \\
    SLidR \cite{sautier2022slidr} & $39.2$ & \underline{$44.2$} & \underline{$0.0$} & $\mathbf{30.8}$ & \underline{$60.2$} & \underline{$15.1$} & \underline{$22.4$} & \underline{$47.2$} & \underline{$27.7$} & \underline{$16.3$} & \underline{$34.3$} & \underline{$80.6$} & \underline{$21.8$} & \underline{$35.2$} & \underline{$48.1$} & \underline{$71.0$} & \underline{$71.9$}
    \\
    ST-SLidR \cite{mahmoud2023st} & $40.5$ & - & - & - & - & - & - & - & - & - & - & - & - & - & - & - & -
    \\
    TriCC \cite{pang2023tricc} & $38.0$ & - & - & - & - & - & - & - & - & - & - & - & - & - & - & - & -
    \\
    Seal \cite{liu2023seal} & \underline{$45.0$} & $\mathbf{54.7}$ & $\mathbf{5.9}$ & \underline{$30.6$} & $\mathbf{61.7}$ & $\mathbf{18.9}$ & $\mathbf{28.8}$ & $\mathbf{48.1}$ & $\mathbf{31.0}$ & $\mathbf{22.1}$ & $\mathbf{39.5}$ & $\mathbf{83.8}$ & $\mathbf{35.4}$ & $\mathbf{46.7}$ & $\mathbf{56.9}$ & $\mathbf{74.7}$ & $\mathbf{74.7}$
    \\
    CSC \cite{chen2024csc} & $\mathbf{46.0}$ & - & - & - & - & - & - & - & - & - & - & - & - & - & - & - & -
    \\
    HVDistill \cite{zhang2024hvdistill} & $39.5$ & - & - & - & - & - & - & - & - & - & - & - & - & - & - & - & -
    \\\midrule
    \multicolumn{18}{l}{\textbf{Distill: ViT-S}}
    \\
    PPKT \cite{liu2021ppkt} & $38.6$ & $43.8$ & $0.0$ & $31.2$ & $53.1$ & $15.2$ & $0.0$ & $42.2$ & $16.5$ & $18.3$ & $33.7$ & $79.1$ & $37.2$ & $45.2$ & $52.7$ & $75.6$ & $74.3$
    \\
    SLidR \cite{sautier2022slidr} & $44.7$ & $45.0$ & $8.2$ & $34.8$ & $58.6$ & \underline{$23.4$} & \underline{$40.2$} & $43.8$ & $19.0$ & $22.9$ & $40.9$ & $82.7$ & $38.3$ & $47.6$ & $53.9$ & \underline{$77.8$} & $77.9$
    \\
    + \textsf{\textcolor{moe_red}{Li}\textcolor{moe_green}{MoE}} & \cellcolor{moe_green!12}$45.8$ & \cellcolor{moe_green!12}$46.2$ & \cellcolor{moe_green!12}$\mathbf{8.5}$ & \cellcolor{moe_green!12}$36.2$ & \cellcolor{moe_green!12}$59.4$ & \cellcolor{moe_green!12}$\mathbf{23.6}$ & \cellcolor{moe_green!12}$\mathbf{41.7}$ & \cellcolor{moe_green!12}$47.2$ & \cellcolor{moe_green!12}$20.7$ & \cellcolor{moe_green!12}$24.1$ & \cellcolor{moe_green!12}$43.2$ & \cellcolor{moe_green!12}$83.9$ & \cellcolor{moe_green!12}\underline{$38.7$} & \cellcolor{moe_green!12}$\mathbf{48.1}$ & \cellcolor{moe_green!12}$55.3$ & \cellcolor{moe_green!12}$\mathbf{78.0}$ & \cellcolor{moe_green!12}\underline{$78.6$}
    \\
    Seal \cite{liu2023seal} & $45.2$ & $48.9$ & \underline{$8.4$} & $30.7$ & $\mathbf{68.1}$ & $17.5$ & $37.7$ & $57.7$ & $17.9$ & $20.9$ & $40.4$ & $83.8$ & $36.6$ & $44.2$ & $54.5$ & $76.2$ & $\mathbf{79.3}$
    \\
    SuperFlow \cite{xu2024superflow} & \underline{$46.4$} & \underline{$49.8$} & $6.8$ & \underline{$45.9$} & $63.4$ & $18.5$ & $31.0$ & \underline{$60.3$} & \underline{$28.1$} & \underline{$25.4$} & \underline{$47.4$} & \underline{$86.2$} & $38.4$ & $47.4$ & \underline{$56.7$} & $74.9$ & $77.8$
    \\
    + \textsf{\textcolor{moe_red}{Li}\textcolor{moe_green}{MoE}} & \cellcolor{moe_green!12}$\mathbf{48.2}$ & \cellcolor{moe_green!12}$\mathbf{50.4}$ & \cellcolor{moe_green!12}$7.9$ & \cellcolor{moe_green!12}$\mathbf{46.7}$ & \cellcolor{moe_green!12}\underline{$65.1$} & \cellcolor{moe_green!12}$19.2$ & \cellcolor{moe_green!12}$32.1$ & \cellcolor{moe_green!12}$\mathbf{61.5}$ & \cellcolor{moe_green!12}$\mathbf{29.5}$ & \cellcolor{moe_green!12}$\mathbf{26.7}$ & \cellcolor{moe_green!12}$\mathbf{48.3}$ & \cellcolor{moe_green!12}$\mathbf{86.5}$ & \cellcolor{moe_green!12}$\mathbf{39.1}$ & \cellcolor{moe_green!12}\underline{$48.0$} & \cellcolor{moe_green!12}$\mathbf{57.4}$ & \cellcolor{moe_green!12}$75.1$ & \cellcolor{moe_green!12}$78.4$
    \\\midrule
    \multicolumn{18}{l}{\textbf{Distill: ViT-B}}
    \\
    PPKT \cite{liu2021ppkt} & $40.0$ & $29.6$ & $0.0$ & $30.7$ & $55.8$ & $6.3$ & $22.4$ & $56.7$ & $18.1$ & $24.3$ & $42.7$ & $82.3$ & $33.2$ & $45.1$ & $53.4$ & $71.3$ & $75.7$
    \\
    SLidR \cite{sautier2022slidr} & $45.4$ & $46.7$ & $7.8$ & $46.5$ & $58.7$ & \underline{$23.9$} & $34.0$ & $47.8$ & $17.1$ & \underline{$23.7$} & $41.7$ & $83.4$ & \underline{$39.4$} & $47.0$ & $54.6$ & $76.6$ & $77.8$
    \\
    + \textsf{\textcolor{moe_red}{Li}\textcolor{moe_green}{MoE}} & \cellcolor{moe_green!12}$46.6$ & \cellcolor{moe_green!12}\underline{$48.2$} & \cellcolor{moe_green!12}$8.6$ & \cellcolor{moe_green!12}$47.1$ & \cellcolor{moe_green!12}$61.1$ & \cellcolor{moe_green!12}$\mathbf{25.0}$ & \cellcolor{moe_green!12}$35.3$ & \cellcolor{moe_green!12}$48.6$ & \cellcolor{moe_green!12}$18.4$ & \cellcolor{moe_green!12}$\mathbf{24.4}$ & \cellcolor{moe_green!12}$43.4$ & \cellcolor{moe_green!12}$84.6$ & \cellcolor{moe_green!12}$\mathbf{39.9}$ & \cellcolor{moe_green!12}$47.4$ & \cellcolor{moe_green!12}\underline{$56.9$} & \cellcolor{moe_green!12}$\mathbf{77.4}$ & \cellcolor{moe_green!12}\underline{$78.9$}
    \\
    Seal \cite{liu2023seal} & $46.6$ & $\mathbf{49.3}$ & $8.2$ & $35.1$ & $\mathbf{70.8}$ & $22.1$ & \underline{$41.7$} & $57.4$ & $15.2$ & $21.6$ & $42.6$ & $84.5$ & $38.1$ & $46.8$ & $55.4$ & \underline{$77.2$} & $\mathbf{79.5}$
    \\
    SuperFlow \cite{xu2024superflow} & \underline{$47.7$} & $45.8$ & \underline{$12.4$} & \underline{$52.6$} & $67.9$ & $17.2$ & $40.8$ & \underline{$59.5$} & \underline{$25.4$} & $21.0$ & \underline{$47.6$} & \underline{$85.8$} & $37.2$ & \underline{$48.4$} & $56.6$ & $76.2$ & $78.2$
    \\
    + \textsf{\textcolor{moe_red}{Li}\textcolor{moe_green}{MoE}} & \cellcolor{moe_green!12}$\mathbf{49.1}$ & \cellcolor{moe_green!12}$46.8$ & \cellcolor{moe_green!12}$\mathbf{13.1}$ & \cellcolor{moe_green!12}$\mathbf{53.9}$ & \cellcolor{moe_green!12}\underline{$68.4$} & \cellcolor{moe_green!12}$19.2$ & \cellcolor{moe_green!12}$\mathbf{42.2}$ & \cellcolor{moe_green!12}$\mathbf{59.9}$ & \cellcolor{moe_green!12}$\mathbf{27.5}$ & \cellcolor{moe_green!12}$21.7$ & \cellcolor{moe_green!12}$\mathbf{48.3}$ & \cellcolor{moe_green!12}$\mathbf{85.9}$ & \cellcolor{moe_green!12}$38.2$ & \cellcolor{moe_green!12}$\mathbf{49.0}$ & \cellcolor{moe_green!12}$\mathbf{57.1}$ & \cellcolor{moe_green!12}$76.3$ & \cellcolor{moe_green!12}$78.8$
    \\\midrule
    \multicolumn{18}{l}{\textbf{Distill: ViT-L}}
    \\
    PPKT \cite{liu2021ppkt} & $41.6$ & $30.5$ & $0.0$ & $32.0$ & $57.3$ & $8.7$ & $24.0$ & \underline{$58.1$} & $19.5$ & $\mathbf{24.9}$ & \underline{$44.1$} & $83.1$ & $34.5$ & $45.9$ & $55.4$ & $72.5$ & $76.4$
    \\
    SLidR \cite{sautier2022slidr} & $45.7$ & $46.9$ & $6.9$ & $44.9$ & $60.8$ & $22.7$ & $40.6$ & $44.7$ & $17.4$ & $23.0$ & $40.4$ & $83.6$ & \underline{$39.9$} & $47.8$ & $55.2$ & $78.1$ & $78.3$
    \\
    + \textsf{\textcolor{moe_red}{Li}\textcolor{moe_green}{MoE}} & \cellcolor{moe_green!12}$47.4$ & \cellcolor{moe_green!12}$48.7$ & \cellcolor{moe_green!12}$9.2$ & \cellcolor{moe_green!12}\underline{$46.7$} & \cellcolor{moe_green!12}$62.7$ & \cellcolor{moe_green!12}$\mathbf{24.2}$ & \cellcolor{moe_green!12}$42.1$ & \cellcolor{moe_green!12}$46.2$ & \cellcolor{moe_green!12}$19.7$ & \cellcolor{moe_green!12}\underline{$24.4$} & \cellcolor{moe_green!12}$43.2$ & \cellcolor{moe_green!12}$\mathbf{85.3}$ & \cellcolor{moe_green!12}$\mathbf{41.6}$ & \cellcolor{moe_green!12}$\mathbf{49.5}$ & \cellcolor{moe_green!12}\underline{$57.4$} & \cellcolor{moe_green!12}$\mathbf{78.7}$ & \cellcolor{moe_green!12}$79.3$
    \\
    Seal \cite{liu2023seal} & $46.8$ & $53.1$ & $6.9$ & $35.0$ & \underline{$65.0$} & $22.0$ & \underline{$46.1$} & $\mathbf{59.2}$ & $16.2$ & $23.0$ & $41.8$ & $84.7$ & $35.8$ & $46.6$ & $55.5$ & \underline{$78.4$} & $\mathbf{79.8}$
    \\
    SuperFlow \cite{xu2024superflow} & \underline{$48.0$} & \underline{$52.3$} & \underline{$12.7$} & $46.5$ & $64.7$ & $21.4$ & $44.9$ & $56.2$ & \underline{$26.7$} & $19.9$ & $43.2$ & $84.2$ & $38.1$ & $47.4$ & $56.9$ & $76.0$ & $79.2$
    \\
    + \textsf{\textcolor{moe_red}{Li}\textcolor{moe_green}{MoE}} & \cellcolor{moe_green!12}$\mathbf{49.4}$ & \cellcolor{moe_green!12}$\mathbf{54.4}$ & \cellcolor{moe_green!12}$\mathbf{14.4}$ & \cellcolor{moe_green!12}$\mathbf{47.9}$ & \cellcolor{moe_green!12}$\mathbf{66.1}$ & \cellcolor{moe_green!12}\underline{$23.9$} & \cellcolor{moe_green!12}$\mathbf{46.7}$ & \cellcolor{moe_green!12}$57.2$ & \cellcolor{moe_green!12}$\mathbf{27.9}$ & \cellcolor{moe_green!12}$20.8$ & \cellcolor{moe_green!12}$\mathbf{44.8}$ & \cellcolor{moe_green!12}\underline{$85.0$} & \cellcolor{moe_green!12}$39.6$ & \cellcolor{moe_green!12}\underline{$48.1$} & \cellcolor{moe_green!12}$\mathbf{58.2}$ & \cellcolor{moe_green!12}$76.5$ & \cellcolor{moe_green!12}\underline{$79.6$}
    \\\bottomrule
\end{tabular}}
\begin{tabular}{c|c}
\end{tabular}
\end{table*}

\begin{table*}[t]
\centering
\caption{The \textbf{per-class IoU scores} of state-of-the-art pretraining methods pretrained and fine-tuned on the \textit{nuScenes} \cite{caesar2020nuScenes,fong2022panoptic-nuScenes} dataset with $1\%$ annotations. All scores are given in percentage (\%). The \textbf{best} and \underline{2nd best} scores under each group are highlighted in \textbf{bold} and \underline{underline}.}
\vspace{-0.2cm}
\label{tab:1pct}
\resizebox{\textwidth}{!}{
\begin{tabular}{r|c|cccccccccccccccc}
    \toprule
    \textbf{Method} & \rotatebox{90}{\textbf{mIoU}} & \rotatebox{90}{\textcolor{nu_barrier}{$\blacksquare$}~barrier} & \rotatebox{90}{\textcolor{nu_bicycle}{$\blacksquare$}~bicycle} & \rotatebox{90}{\textcolor{nu_bus}{$\blacksquare$}~bus} & \rotatebox{90}{\textcolor{nu_car}{$\blacksquare$}~car} & \rotatebox{90}{\textcolor{nu_cons}{$\blacksquare$}~construction vehicle~} & \rotatebox{90}{\textcolor{nu_motor}{$\blacksquare$}~motorcycle} & \rotatebox{90}{\textcolor{nu_ped}{$\blacksquare$}~pedestrian} & \rotatebox{90}{\textcolor{nu_cone}{$\blacksquare$}~traffic cone} & \rotatebox{90}{\textcolor{nu_trailer}{$\blacksquare$}~trailer} & \rotatebox{90}{\textcolor{nu_truck}{$\blacksquare$}~truck} & \rotatebox{90}{\textcolor{nu_driv}{$\blacksquare$}~driveable surface} & \rotatebox{90}{\textcolor{nu_flat}{$\blacksquare$}~other flat} & \rotatebox{90}{\textcolor{nu_sidewalk}{$\blacksquare$}~sidewalk} & \rotatebox{90}{\textcolor{nu_terrain}{$\blacksquare$}~terrain} & \rotatebox{90}{\textcolor{nu_manmade}{$\blacksquare$}~manmade} & \rotatebox{90}{\textcolor{nu_veg}{$\blacksquare$}~vegetation}
    \\\midrule
    \cellcolor{moe_gray!18}\textcolor{gray}{Random} & \cellcolor{moe_gray!18}\textcolor{gray}{$30.3$} & \cellcolor{moe_gray!18}\textcolor{gray}{$0.0$} &
    \cellcolor{moe_gray!18}\textcolor{gray}{$0.0$} & \cellcolor{moe_gray!18}\textcolor{gray}{$8.1$} & \cellcolor{moe_gray!18}\textcolor{gray}{$65.0$} & \cellcolor{moe_gray!18}\textcolor{gray}{$0.1$} & \cellcolor{moe_gray!18}\textcolor{gray}{$6.6$} & \cellcolor{moe_gray!18}\textcolor{gray}{$21.0$} & \cellcolor{moe_gray!18}\textcolor{gray}{$9.0$} & \cellcolor{moe_gray!18}\textcolor{gray}{$9.3$} & \cellcolor{moe_gray!18}\textcolor{gray}{$25.8$} & \cellcolor{moe_gray!18}\textcolor{gray}{$89.5$} & \cellcolor{moe_gray!18}\textcolor{gray}{$14.8$} & \cellcolor{moe_gray!18}\textcolor{gray}{$41.7$} & \cellcolor{moe_gray!18}\textcolor{gray}{$48.7$} & \cellcolor{moe_gray!18}\textcolor{gray}{$72.4$} & \cellcolor{moe_gray!18}\textcolor{gray}{$73.3$}
    \\\midrule
    \multicolumn{18}{l}{\textbf{Distill: None}}
    \\
    PointContrast \cite{xie2020pointcontrast} & $32.5$ & $0.0$ & \underline{$1.0$} & $5.6$ & \underline{$67.4$} & $0.0$ & \underline{$3.3$} & \underline{$31.6$} & $5.6$ & \underline{$12.1$} & \underline{$30.8$} & $\mathbf{91.7}$ & \underline{$21.9$} & $\mathbf{48.4}$ & \underline{$50.8$} & \underline{$75.0$} & \underline{$74.6$}
    \\
    DepthContrast \cite{zhang2021depthcontrast} & $31.7$ & $0.0$ & $0.6$ & \underline{$6.5$} & $64.7$ & \underline{$0.2$} & $\mathbf{5.1}$ & $29.0$ & $\mathbf{9.5}$ & \underline{$12.1$} & $29.9$ & \underline{$90.3$} & $17.8$ & \underline{$44.4$} & $49.5$ & $73.5$ & $74.0$
    \\
    ALSO \cite{boulch2023also} & \underline{$37.7$} & - & - & - & - & - & - & - & - & - & - & - & - & - & - & - & -
    \\
    BEVContrast \cite{sautier2024bevcontrast} & $\mathbf{37.9}$ & $0.0$ & $\mathbf{1.3}$ & $\mathbf{32.6}$ & $\mathbf{74.3}$ & $\mathbf{1.1}$ & $0.9$ & $\mathbf{41.3}$ & \underline{$8.1$} & $\mathbf{24.1}$ & $\mathbf{40.9}$ & $89.8$ & $\mathbf{36.2}$ & $44.0$ & $\mathbf{52.1}$ & $\mathbf{79.9}$ & $\mathbf{79.7}$
    \\\midrule
    \multicolumn{18}{l}{\textbf{Distill: ResNet-50}}
    \\
    PPKT \cite{liu2021ppkt} & $37.8$ & $0.0$ & \underline{$2.2$} & $20.7$ & \underline{$75.4$} & $1.2$ & $13.2$ & $45.6$ & $8.5$ & $17.5$ & $38.4$ & \underline{$92.5$} & $19.2$ & $52.3$ & $56.8$ & $80.1$ & $80.9$
    \\
    SLidR \cite{sautier2022slidr} & $38.8$ & $0.0$ & $1.8$ & $15.4$ & $73.1$ & \underline{$1.9$} & $19.9$ & $47.2$ & $17.1$ & $14.5$ & $34.5$ & $92.0$ & $27.1$ & $53.6$ & \underline{$61.0$} & $79.8$ & $82.3$
    \\
    ST-SLidR \cite{mahmoud2023st} & $40.8$ & - & - & - & - & - & - & - & - & - & - & - & - & - & - & - & -
    \\
    TriCC \cite{pang2023tricc} & $41.2$ & - & - & - & - & - & - & - & - & - & - & - & - & - & - & - & -
    \\
    Seal \cite{liu2023seal} & \underline{$45.8$} & $0.0$ & $\mathbf{9.4}$ & \underline{$32.6$} & $\mathbf{77.5}$ & $\mathbf{10.4}$ & \underline{$28.0$} & \underline{$53.0$} & \underline{$25.0$} & $\mathbf{30.9}$ & $\mathbf{49.7}$ & $\mathbf{94.0}$ & $\mathbf{33.7}$ & $\mathbf{60.1}$ & $59.6$ & $\mathbf{83.9}$ & \underline{$83.4$}
    \\
    CSC \cite{chen2024csc} & $\mathbf{47.0}$ & $0.0$ & $0.0$ & $\mathbf{58.7}$ & $74.0$ & $0.1$ & $\mathbf{40.9}$ & $\mathbf{58.9}$ & $\mathbf{31.8}$ & \underline{$23.7$} & \underline{$45.1$} & \underline{$92.5$} & \underline{$33.0$} & \underline{$56.4$} & $\mathbf{62.4}$ & \underline{$81.6$} & $\mathbf{84.2}$
    \\
    HVDistill \cite{zhang2024hvdistill} & $42.7$ & - & - & - & - & - & - & - & - & - & - & - & - & - & - & - & -
    \\\midrule
    \multicolumn{18}{l}{\textbf{Distill: ViT-S}}
    \\
    PPKT \cite{liu2021ppkt} & $40.6$ & $0.0$ & $0.0$ & $25.2$ & $73.5$ & $9.1$ & $6.9$ & $51.4$ & $8.6$ & $11.3$ & $31.1$ & $93.2$ & $\mathbf{41.7}$ & $58.3$ & $64.0$ & $82.0$ & $82.6$
    \\
    SLidR \cite{sautier2022slidr} & $41.2$ & $0.0$ & $0.0$ & $26.6$ & $72.0$ & $12.4$ & $15.8$ & $51.4$ & $22.9$ & $11.7$ & $35.3$ & $92.9$ & $36.3$ & $58.7$ & $63.6$ & $81.2$ & $82.3$
    \\
    + \textsf{\textcolor{moe_red}{Li}\textcolor{moe_green}{MoE}} & \cellcolor{moe_green!12}$46.8$ & \cellcolor{moe_green!12}$20.6$ & \cellcolor{moe_green!12}\underline{$4.2$} & \cellcolor{moe_green!12}$\mathbf{29.7}$ & \cellcolor{moe_green!12}$74.7$ & \cellcolor{moe_green!12}\underline{$16.9$} & \cellcolor{moe_green!12}$24.6$ & \cellcolor{moe_green!12}$55.7$ & \cellcolor{moe_green!12}\underline{$28.3$} & \cellcolor{moe_green!12}$19.5$ & \cellcolor{moe_green!12}$41.5$ & \cellcolor{moe_green!12}\underline{$93.8$} & \cellcolor{moe_green!12}\underline{$41.0$} & \cellcolor{moe_green!12}$\mathbf{62.4}$ & \cellcolor{moe_green!12}$\mathbf{67.3}$ & \cellcolor{moe_green!12}$82.6$ & \cellcolor{moe_green!12}$\mathbf{85.2}$
    \\
    Seal \cite{liu2023seal} & $44.3$ & $20.0$ & $0.0$ & $19.4$ & $74.7$ & $10.6$ & $\mathbf{45.7}$ & \underline{$60.3$} & $\mathbf{29.2}$ & $17.4$ & $38.1$ & $93.2$ & $26.0$ & $58.8$ & $64.5$ & $81.9$ & $81.9$
    \\
    SuperFlow \cite{xu2024superflow} & \underline{$47.8$} & \underline{$38.2$} & $1.8$ & $25.8$ & \underline{$79.0$} & $15.3$ & $43.6$ & \underline{$60.3$} & $0.0$ & \underline{$28.4$} & \underline{$55.4$} & $93.7$ & $28.8$ & $59.1$ & $59.9$ & \underline{$83.5$} & $83.1$
    \\
    + \textsf{\textcolor{moe_red}{Li}\textcolor{moe_green}{MoE}} & \cellcolor{moe_green!12}$\mathbf{49.6}$ & \cellcolor{moe_green!12}$\mathbf{39.9}$ & \cellcolor{moe_green!12}$\mathbf{4.6}$ & \cellcolor{moe_green!12}\underline{$27.3$} & \cellcolor{moe_green!12}$\mathbf{80.2}$ & \cellcolor{moe_green!12}$\mathbf{17.1}$ & \cellcolor{moe_green!12}\underline{$45.4$} & \cellcolor{moe_green!12}$\mathbf{61.2}$ & \cellcolor{moe_green!12}$6.2$ & \cellcolor{moe_green!12}$\mathbf{29.5}$ & \cellcolor{moe_green!12}$\mathbf{58.4}$ & \cellcolor{moe_green!12}$\mathbf{94.0}$ & \cellcolor{moe_green!12}$34.2$ & \cellcolor{moe_green!12}\underline{$62.3$} & \cellcolor{moe_green!12}\underline{$64.6$} & \cellcolor{moe_green!12}$\mathbf{84.1}$ & \cellcolor{moe_green!12}\underline{$84.5$}
    \\\midrule
    \multicolumn{18}{l}{\textbf{Distill: ViT-B}}
    \\
    PPKT \cite{liu2021ppkt} & $40.9$ & $0.0$ & $0.0$ & $24.5$ & $73.5$ & $12.2$ & $7.0$ & $51.0$ & $13.5$ & $15.4$ & $36.3$ & $93.1$ & \underline{$40.4$} & $59.2$ & $63.5$ & $81.7$ & $82.2$
    \\
    SLidR \cite{sautier2022slidr} & $41.6$ & $0.0$ & $0.0$ & $26.7$ & $73.4$ & $10.3$ & $16.9$ & $51.3$ & \underline{$23.3$} & $12.7$ & $38.1$ & $93.0$ & $37.7$ & $58.8$ & $63.4$ & $81.6$ & $82.7$
    \\
    + \textsf{\textcolor{moe_red}{Li}\textcolor{moe_green}{MoE}} & \cellcolor{moe_green!12}$46.9$ & \cellcolor{moe_green!12}$22.7$ & \cellcolor{moe_green!12}\underline{$2.6$} & \cellcolor{moe_green!12}$28.3$ & \cellcolor{moe_green!12}$75.4$ & \cellcolor{moe_green!12}$\mathbf{13.5}$ & \cellcolor{moe_green!12}$27.8$ & \cellcolor{moe_green!12}$55.0$ & \cellcolor{moe_green!12}$\mathbf{28.5}$ & \cellcolor{moe_green!12}$22.2$ & \cellcolor{moe_green!12}$40.6$ & \cellcolor{moe_green!12}\underline{$93.7$} & \cellcolor{moe_green!12}$\mathbf{42.3}$ & \cellcolor{moe_green!12}$61.9$ & \cellcolor{moe_green!12}$\mathbf{66.8}$ & \cellcolor{moe_green!12}\underline{$83.1$} & \cellcolor{moe_green!12}$\mathbf{85.4}$
    \\
    Seal \cite{liu2023seal} & $46.0$ & $\mathbf{43.0}$ & $0.0$ & $26.7$ & \underline{$81.3$} & $9.9$ & $41.3$ & $56.2$ & $0.0$ & $21.7$ & $51.6$ & $93.6$ & $\mathbf{42.3}$ & \underline{$62.8$} & $64.7$ & $82.6$ & $82.7$
    \\
    SuperFlow \cite{xu2024superflow} & \underline{$48.1$} & $39.1$ & $0.9$ & \underline{$30.0$} & $80.7$ & $10.3$ & \underline{$47.1$} & \underline{$59.5$} & $5.1$ & \underline{$27.6$} & \underline{$55.4$} & \underline{$93.7$} & $29.1$ & $61.1$ & $63.5$ & $82.7$ & $83.6$
    \\
    + \textsf{\textcolor{moe_red}{Li}\textcolor{moe_green}{MoE}} & \cellcolor{moe_green!12}$\mathbf{50.2}$ & \cellcolor{moe_green!12}\underline{$41.5$} & \cellcolor{moe_green!12}$\mathbf{3.8}$ & \cellcolor{moe_green!12}$\mathbf{32.2}$ & \cellcolor{moe_green!12}$\mathbf{81.7}$ & \cellcolor{moe_green!12}\underline{$12.9$} & \cellcolor{moe_green!12}$\mathbf{49.3}$ & \cellcolor{moe_green!12}$\mathbf{61.1}$ & \cellcolor{moe_green!12}$7.3$ & \cellcolor{moe_green!12}$\mathbf{29.3}$ & \cellcolor{moe_green!12}$\mathbf{57.8}$ & \cellcolor{moe_green!12}$\mathbf{94.2}$ & \cellcolor{moe_green!12}$35.1$ & \cellcolor{moe_green!12}$\mathbf{62.9}$ & \cellcolor{moe_green!12}\underline{$65.4$} & \cellcolor{moe_green!12}$\mathbf{84.0}$ & \cellcolor{moe_green!12}\underline{$84.8$}
    \\\midrule
    \multicolumn{18}{l}{\textbf{Distill: ViT-L}}
    \\
    PPKT \cite{liu2021ppkt} & $42.1$ & $0.0$ & $0.0$ & $24.4$ & $78.8$ & $15.1$ & $9.2$ & $54.2$ & $14.3$ & $12.9$ & $39.1$ & $92.9$ & $37.8$ & $59.8$ & $64.9$ & $82.3$ & $83.6$
    \\
    SLidR \cite{sautier2022slidr} & $42.8$ & $0.0$ & $0.0$ & $23.9$ & $78.8$ & $15.2$ & $20.9$ & $55.0$ & \underline{$28.0$} & $17.4$ & $41.4$ & $92.2$ & $41.2$ & $58.0$ & $64.0$ & $81.8$ & $82.7$
    \\
    + \textsf{\textcolor{moe_red}{Li}\textcolor{moe_green}{MoE}} & \cellcolor{moe_green!12}$46.9$ & \cellcolor{moe_green!12}$21.6$ & \cellcolor{moe_green!12}\underline{$1.6$} & \cellcolor{moe_green!12}\underline{$25.2$} & \cellcolor{moe_green!12}$80.1$ & \cellcolor{moe_green!12}$17.3$ & \cellcolor{moe_green!12}$28.0$ & \cellcolor{moe_green!12}$56.4$ & \cellcolor{moe_green!12}$\mathbf{28.3}$ & \cellcolor{moe_green!12}$18.6$ & \cellcolor{moe_green!12}$43.1$ & \cellcolor{moe_green!12}$92.7$ & \cellcolor{moe_green!12}$\mathbf{41.7}$ & \cellcolor{moe_green!12}$60.9$ & \cellcolor{moe_green!12}$\mathbf{65.5}$ & \cellcolor{moe_green!12}$83.8$ & \cellcolor{moe_green!12}$\mathbf{85.6}$
    \\
    Seal \cite{liu2023seal} & $46.3$ & $41.8$ & $0.0$ & $23.8$ & \underline{$81.4$} & \underline{$17.7$} & $46.3$ & $58.6$ & $0.0$ & $23.4$ & $54.7$ & $93.8$ & \underline{$41.4$} & \underline{$62.5$} & \underline{$65.0$} & $\underline{83.9}$ & $83.8$
    \\
    SuperFlow \cite{xu2024superflow} & \underline{$50.0$} & \underline{$44.5$} & $0.9$ & $22.4$ & $80.8$ & $17.1$ & \underline{$50.2$} & \underline{$60.9$} & $21.0$ & \underline{$25.1$} & \underline{$55.1$} & \underline{$93.9$} & $35.8$ & $61.5$ & $62.6$ & $83.7$ & $83.7$
    \\
    + \textsf{\textcolor{moe_red}{Li}\textcolor{moe_green}{MoE}} & \cellcolor{moe_green!12}$\mathbf{51.4}$ & \cellcolor{moe_green!12}$\mathbf{45.3}$ & \cellcolor{moe_green!12}$\mathbf{4.1}$ & \cellcolor{moe_green!12}$\mathbf{25.3}$ & \cellcolor{moe_green!12}$\mathbf{82.2}$ & \cellcolor{moe_green!12}$\mathbf{18.4}$ & \cellcolor{moe_green!12}$\mathbf{52.5}$ & \cellcolor{moe_green!12}$\mathbf{61.8}$ & \cellcolor{moe_green!12}$22.3$ & \cellcolor{moe_green!12}$\mathbf{26.4}$ & \cellcolor{moe_green!12}$\mathbf{56.2}$ & \cellcolor{moe_green!12}$\mathbf{94.3}$ & \cellcolor{moe_green!12}$37.6$ & \cellcolor{moe_green!12}$\mathbf{63.3}$ & \cellcolor{moe_green!12}$63.9$ & \cellcolor{moe_green!12}$\mathbf{84.4}$ & \cellcolor{moe_green!12}\underline{$85.0$}
    \\\bottomrule
\end{tabular}}
\begin{tabular}{c|c}
\end{tabular}
\end{table*}

\clearpage

\clearpage

\clearpage\clearpage

{
    \small
    \bibliographystyle{ieeenat_fullname}
    \bibliography{main}
}

\end{document}